\documentclass[10pt,twocolumn,letterpaper]{article}

\usepackage{iccv}
\usepackage{times}
\usepackage{epsfig}
\usepackage{graphicx}
\usepackage{amsmath}
\usepackage{amssymb}
\usepackage{subfigure}
\usepackage{color}

 \makeatletter
 \DeclareRobustCommand\onedot{\futurelet\@let@token\@onedot}
 \def\@onedot{\ifx\@let@token.\else.\null\fi\xspace}
 \def\eg{e.g\onedot}

 \makeatother

\DeclareRobustCommand{\Figref}[1]{Figure~\ref{#1}}

\DeclareRobustCommand{\Figureref}[1]{Figure~\ref{#1}}

\graphicspath{{./fig/}{./fig/plots/}}

\newcommand{\md}[1]{\textcolor{black}{#1}}

\usepackage[breaklinks=true,bookmarks=false]{hyperref}

\iccvfinalcopy %

\def\iccvPaperID{1704} %

\ificcvfinal\fi
\begin{document}

\title{Spatial Semantic Regularisation for Large Scale Object Detection}

\author{Damian Mrowca\textsuperscript{$1,2$}, 
Marcus Rohrbach\textsuperscript{$1,3$}, 
Judy Hoffman\textsuperscript{$1$}, 
Ronghang Hu\textsuperscript{$1$}, 
Kate Saenko\textsuperscript{$4$},
Trevor Darrell\textsuperscript{$1$}\\
\textsuperscript{$1$}UC Berkeley, USA;
\textsuperscript{2}TU Muenchen, Germany;
\textsuperscript{3}ICSI, Berkeley, USA;
\textsuperscript{4}UMass Lowell, USA\\
}

\maketitle
\begin{abstract}
Large scale object detection with thousands of classes introduces the problem of many contradicting false positive detections, which have to be suppressed. Class-independent non-maximum suppression has traditionally been used for this step, but it does not scale well as the number of classes grows. Traditional non-maximum suppression does not consider label- and instance-level relationships nor does it allow an exploitation of the spatial layout of detection proposals.
We propose a new multi-class spatial semantic regularisation method based on affinity propagation clustering \md{\cite{frey2007clustering,eth_biwi_01126}}, which simultaneously optimises across all categories and all proposed locations in the image, to improve both the localisation and categorisation of selected detection proposals. 
Constraints are shared across the labels through the semantic WordNet hierarchy.
Our approach proves to be especially useful in large scale settings with thousands of classes, where spatial and semantic interactions are very frequent and only weakly supervised detectors can be built due to a lack of bounding box annotations.
Detection experiments are conducted on the ImageNet and COCO dataset, and in settings with thousands of detected categories. Our method provides a significant precision improvement by reducing false positives, while simultaneously improving the recall.
\end{abstract}

\begin{figure}[t]
\includegraphics[width=\linewidth,keepaspectratio]{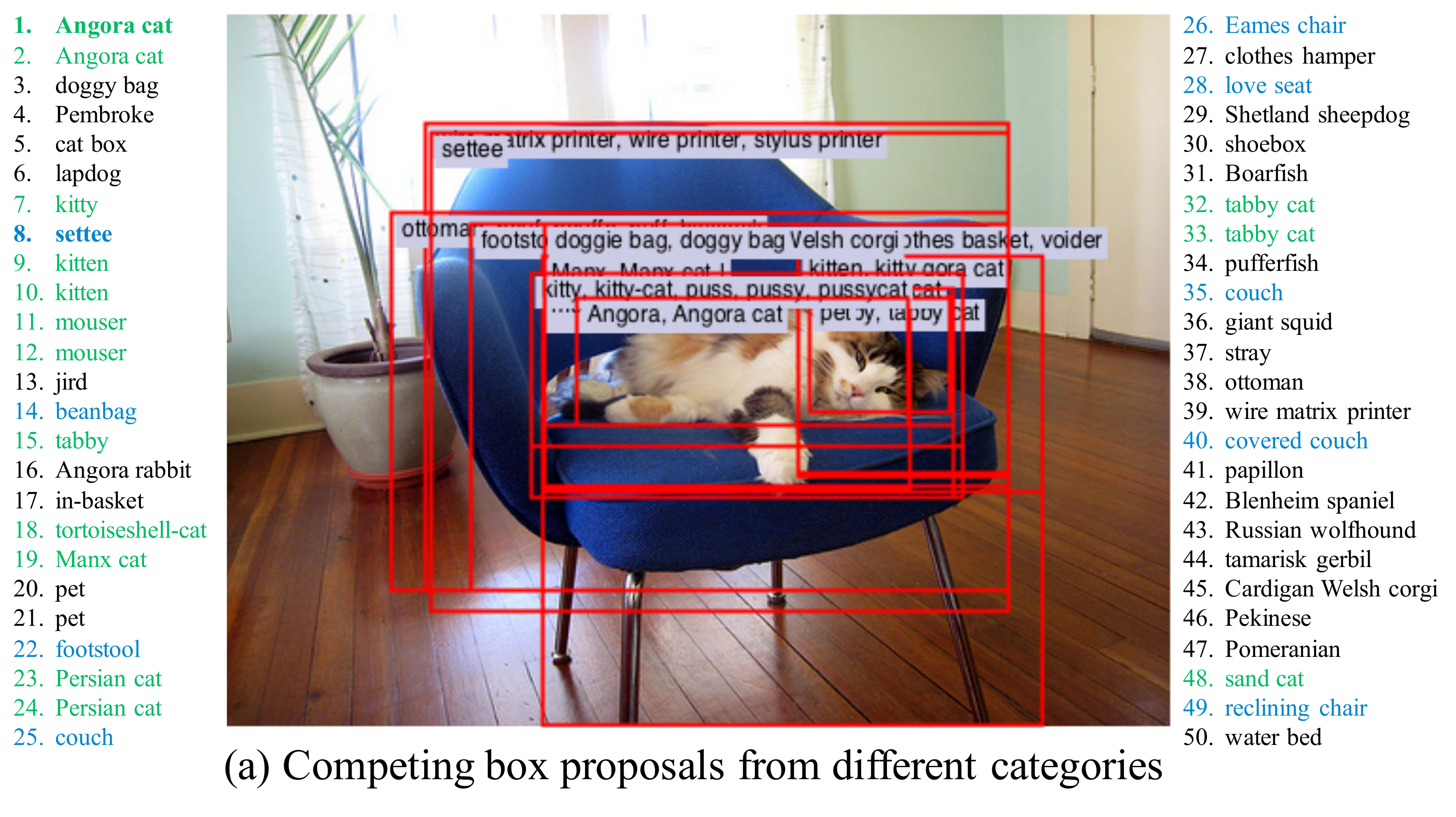}
\includegraphics[width=\linewidth,keepaspectratio]{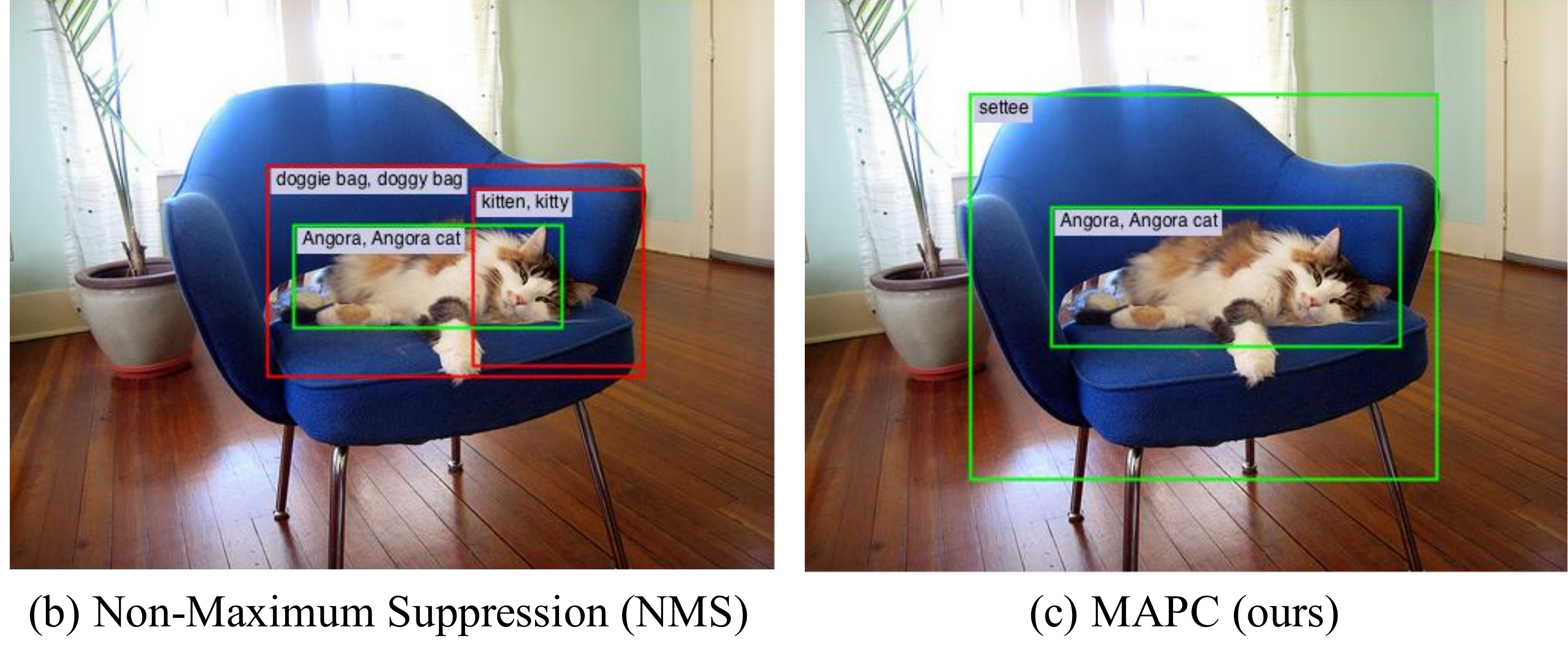}
   \caption{Raw and spatially regularised detection of 7,404 classes using the LSDA extension \cite{Hoffman14Lsda} of the R-CNN method \cite{girshick2014rich}.
   (a) Top 50 scoring candidate detections and associated categories are listed: all proposals which support the depicted \emph{cat} are green, for \emph{chair} blue. Black entries do not describe any object in this picture. (b) NMS clusters boxes only according to their overlap not according to their class leading to multiple detections of different fine-grained classes for the same object. (c) Our approach (MAPC) exploits category relationships, clustering overlapping boxes with similar classes together which results in less false positives on the same object and enables to detect classes which are suppressed by NMS because of their overlap.}
\label{fig:teaser}
\end{figure}

\section{Introduction}
Human assistance technologies or question answering require a precise and detailed object recognition of a visual scene.  
Recently, large scale detection approaches have been proposed which aim to distinguish hundreds or thousands of object categories \cite{Chen2013,dean2013fast,Guillaumin2012,Hoffman14Lsda,Russakovsky2013}. While impressive progress has been shown, they suffer from competing object category candidate detections as can be seen in \Figref{fig:teaser} (a). Commonly, non-maximum suppression (NMS) is used to select the bounding boxes with the highest detection score for each category. This method is not globally optimal as only locally overlapping boxes are suppressed by the highest scoring box. Further, in the multi-class case, it does not take semantic relations between objects into account, \eg the \emph{couch}, \emph{floorstool} and \emph{beanbag} proposals should support the \emph{settee} candidate detection box in \Figref{fig:teaser}, such that it is not suppressed by  \emph{doggy bag} as in \Figref{fig:teaser}(b).

With thousands of different object categories, semantic relationships become a valuable source of information. Using semantics, consistency can be ensured across different detections. %
Hence, this work examines the benefit of a semantic hierarchy %
to object detection of thousands of object categories. We show that in such a large scale setting %
semantic %
constraints significantly improve detection. 

The key contribution of this work is a large scale spatial semantic regulariser for the correct selection of candidate object detection proposals.
Under the framework of Affinity Propagation Clustering (APC) \cite{frey2007clustering}, our developed method is characterised by two new ideas.

First, we present an approach which unifies within and across class selection of candidate object detections. Our new multi-class affinity propagation clustering (MAPC) allows 
for global reasoning over the whole image simultaneously, rather than reasoning locally over image parts or single classes separately, to determine the correct setup of an image. Unlike NMS or [20], which perform the selection separately for each class, our algorithm uses the relationships of highly related fine-grained categories in a large scale detection setting. Based on WordNet relationships, our algorithm knows that \emph{golden retrievers}, \emph{dalmatians} and \emph{dachshunds} are all different dog breeds and should support each other, rather than suppress, if the corresponding boxes cover almost identical regions of interest in the image.

Second, we propose a large scale detection evaluation including over a thousand categories, which requires discriminating among competing classes, in contrast to standard detection challenges, which focus on a per category mean Average Precision (mAP) evaluation. We demonstrate that our algorithm improves performance in two challenging scenarios. First, for a large number of objects per image, we show results on COCO. Second, for a large number of categories, we evaluate on a subset of ImageNet, which is labeled with bounding boxes of 1,825 categories, a large scale detection scenario, which has not been previously evaluated.
\section{Related Work}

Our work is most related to spatial regularisation over detection proposals. In most detection methods, detection proposals (raw bounding box outputs with scores from detectors) need to be regularised over space to remove double detections on the same object, prune false positives, and improve localisation. Although greedy non-maximum suppression (NMS) is the most often used spatial regularisation approach, other approaches, such as merging nearby detection boxes, are sometimes shown to be more robust \cite{Sermanet2013}. In \cite{Viola2004}, overlapping detections are averaged and a threshold is set based on overlapping box numbers. In \cite{Sermanet2013}, a greedy merge strategy is proposed to group detection proposals together and reward bounding box coherence. Spatial and cooccurrence priors are introduced in \cite{Choi2010,Torralba2003} to prune detection results. In \cite{Desai2009}, labels of detection proposals are obtained via approximate inference over several types of spatial relationships instead of greedy NMS. Recently, Affinity Propagation Clustering (APC) \cite{frey2007clustering}, an unsupervised clustering method based on message passing, has been used to cluster proposed bounding boxes of the same class based on their overlap~\cite{eth_biwi_01126}. In \cite{eth_biwi_01126}, background and repellence terms are introduced to APC to allow the suppression of false positives and to avoid selecting object proposals lying too close to each other. Our work builds on \cite{eth_biwi_01126}, but is different in that: (1) our algorithm clusters object proposals of the same and different classes simultaneously, whereas \cite{eth_biwi_01126} is applied only within each class,  (2) we introduce new constraints to ensure that one label per detection proposal is selected, and (3) we design our similarity measure such that semantically close objects get clustered together.

Another line of related work is exploiting semantic category hierarchies in visual recognition and detection \cite{deng2014large,Deng2012,Guillaumin2012,Kuettel2012,Li2010,marszalek2007semantic,rohrbach13nips,Salakhutdinov2011,zweig2007exploiting}. Real world object categories often form a hierarchical structure, which can provide useful information for large scale detection. Such hierarchical relationships can be obtained from predefined semantic structures such as WordNet, or learned by data-driven methods. In \cite{deng2014large}, a conditional random field based hierarchy-exclusion Graph is proposed to represent subsumption and exclusion relationships between classes. In \cite{Guillaumin2012,Kuettel2012}, the ImageNet hierarchy, which is based on WordNet, is used to transfer bounding box annotations and segmentations to semantically close categories. In \cite{Deng2012}, an accuracy-specificity trade-off based on the ImageNet hierarchy is optimised through the DARTS algorithm. PST \cite{rohrbach13nips} uses the WordNet hierarchy to transfer knowledge from known to novel categories and propagates information between instances of the novel categories. In \cite{Salakhutdinov2011} a visual hierarchy is discovered based on the Chinese Restaurant Prior and used to share detector parameters between classes. \cite{Li2010} learn a semantic hierarchy based on visual and semantical constraints. Our work is complementary to previous methods in this area, as we integrate a semantic hierarchy into Multi-class Affinity Propagation Clustering (MAPC) for spatial regularisation, while hierarchies have been only used to train classifiers or share features in previous methods.

Our work is also related to large scale detection. In \cite{dean2013fast}, large scale detectors on over 100,000 classes are trained based on hashing. In \cite{Chen2013}, NEIL, a semi-supervised learning system, is proposed to train detectors from Internet images. One major obstacle for large scale detection is the lack of bounding box annotations, which has been recently partially resolved by weakly supervised methods such as knowledge transfer \cite{Guillaumin2012}, Multiple Instance Learning \cite{song2014learning, Pathak2014}, domain adaptation \cite{Hoffman14Lsda} or combined approaches \cite{Hoffman2014}. Among these methods, LSDA \cite{Hoffman14Lsda} is a framework for classifier-to-detector adaptation, and was shown to effectively train large scale detectors based on image-level labels. Thus, in this paper we use LSDA to train a baseline detector on 7,404 leaf classes of the ImageNet hierarchy. However, we note that our spatial regularisation method does not depend on how detectors are trained, and can be applied to arbitrary sets of detectors.

To our knowledge, this is the first time that hierarchical semantic relationships are used together with spatial information to determine the correct scene configuration from contradicting candidate object detections. Furthermore, it is even more challenging to apply this algorithm on a large scale setting, as it requires inference over thousands of fine-grained and diverse categories. %
Our detection system is unique in its amount of categories, both in terms of the degree of fine-grained detail, for instance incorporating different dog breeds, and the variety of categories, including various animals, plants, fruits, and man-made structures.

\section{Spatial semantic regulariser}

In this section we describe our spatial semantic regulariser. Our method is based on Affinity Propagation Clustering (APC), which has shown to outperform other clustering techniques such as k-means clustering \cite{frey2007clustering}. \md{\cite{eth_biwi_01126} successfully adapted APC to the task of selecting candidate object detections of the same class. This method is denoted as Single-class APC (SAPC) in the following.} 

\begin{figure}[t]
\begin{center}
\includegraphics[width=\linewidth,keepaspectratio]{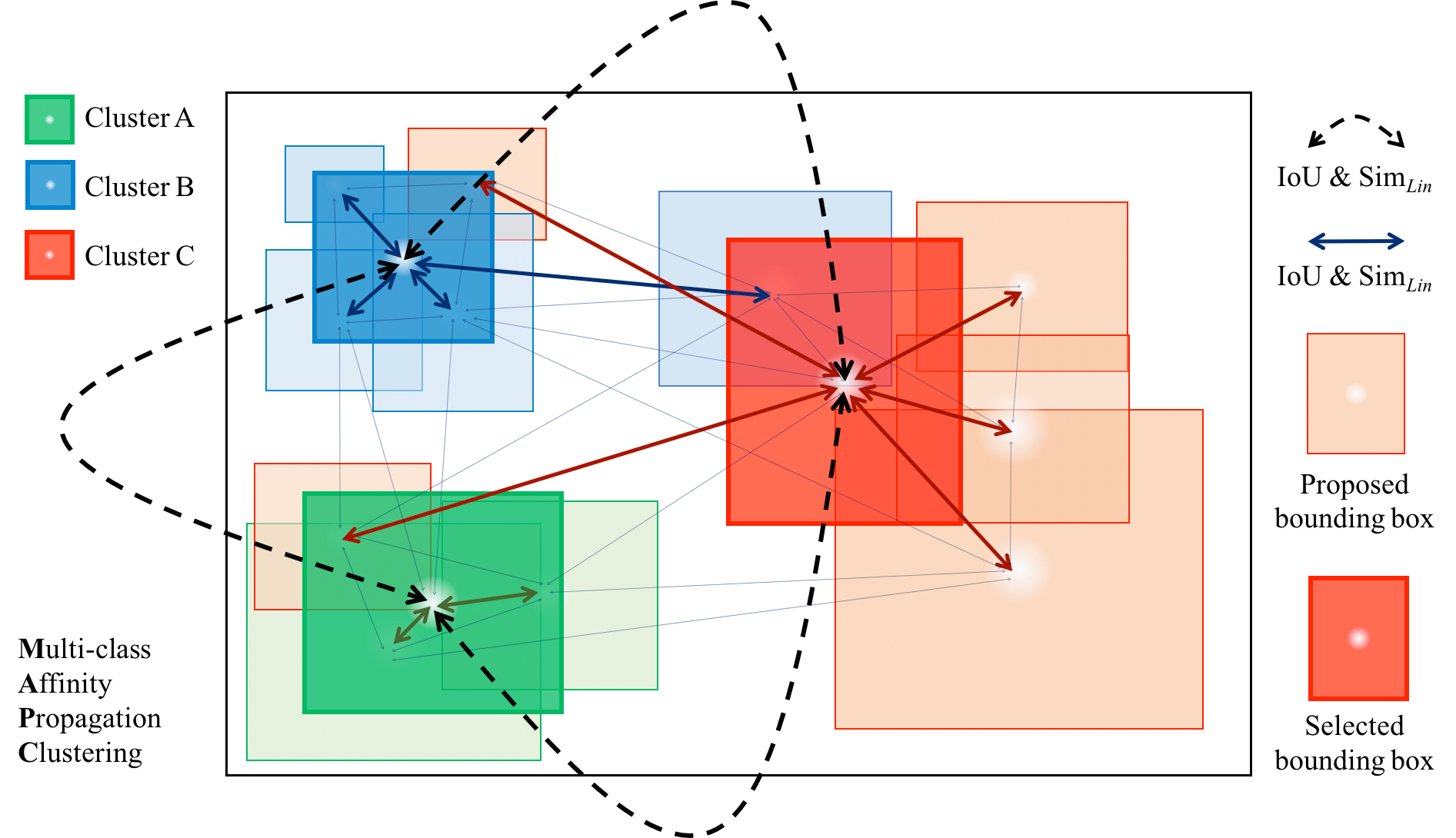}
\end{center}
   \caption{MAPC message passing. Messages are passed between all candidate \md{detections} until a subset of \md{detections} gets selected as \md{exemplars}. IoU stands for Intersection over Union, %
   and $\text{sim}_{Lin}$ is the $Lin$ measure. For simplicity not all messages are depicted.}
\label{fig:msgpassing}
\end{figure}

Our main contributions are to \md{extend the previous work on APC \md{\cite{frey2007clustering, givoni2009binary, eth_biwi_01126}} to multi-class detection and} a large scale setting with thousands of fine-grained classes. Therefore, we incorporate a new constraint ensuring that each bounding box exemplar gets assigned only one label. Similar to \cite{eth_biwi_01126}, we use an intercluster repellence term and a background category to remove false positives. Additionally, in order to leverage the visual similarity of semantically related fine-grained classes, we introduce hierarchical label relations into APC to cluster semantically similar objects. 
The resulting Multi-class APC (MAPC) algorithm is presented in \Figref{fig:msgpassing} after introducing standard APC.

\subsection{Standard affinity propagation clustering}

APC is a message passing based clustering method. It uses data similarities to identify exemplars such that the sum of similarities between \md{cluster} exemplars and cluster members is maximised. Let $s(i, j)$ denote the similarity between data points $i$ and $j \in \{1,... , N\}$ with $N$ being the number of data points. $s(i, j) \leq 0$ indicates how well $j$ would serve as an exemplar for $i$ \cite{frey2007clustering}. The self-similarity $s(i,i)$ indicates how likely a certain point will be chosen as an exemplar.
Using the binary formulation of \cite{givoni2009binary}, we encode the exemplar assignment with a set of $N^{2}$ binary variables $c_{ij}$: $c_{ij} = 1$ if $i$ is represented by $j$ and $c_{ij} = 0$ otherwise. A valid clustering must hold two constraints: (i) each point is represented by exactly one \md{exemplar} and (ii) when $j$ represents any other point $i$, then $j$ must be \md{an exemplar} representing itself. In the following objective function, $I$ represents constraint (i) and $E$ represents constraint (ii):

\begin{equation} \label{eq:objectivefunction}
\begin{aligned} 
E_{APC}(\{c_{ij}\}) &= \sum_{i,j}S_{ij}(c_{ij}) + \sum_{i}I_{i}(c_{i1},...,c_{iN}) \\ &+ \sum_{j}E_{j}(c_{1j},...,c_{Nj})
\end{aligned}
\end{equation}
\begin{equation} \label{eq:Sij}
  S_{ij}(c_{ij})=\begin{cases}
               s(i,j) &\text{if} \ c_{ij} = 1\\
               0 &\text{otherwise}
            \end{cases}
\end{equation}
\begin{equation} \label{eq:Ii}
  I_{i}(c_{i1},...,c_{iN})=\begin{cases}
               - \infty &\text{if} \ \sum_{j}c_{ij} \ne 1\\
               0 &\text{otherwise}
            \end{cases}
\end{equation}
\begin{equation} \label{eq:Ej}
  E_{j}(c_{1j},...,c_{Nj})=\begin{cases}
               - \infty &\text{if} \ c_{jj}=0 \ \text{and} \  \exists i \ne j \\ &\text{s.t.} \ c_{ij}=1\\
               0 &\text{otherwise}
            \end{cases}
\end{equation}

Max-sum message passing is applied to maximise equation (\ref{eq:objectivefunction}) \cite{frey2007clustering, givoni2009binary} consisting of two messages: The responsibility $\rho_{ij}$ (sent from $i$ to $j$) describes how suited $j$ would be as an exemplar for $i$. The availability $\alpha_{ij}$ (sent from $j$ to $i$) reflects the accumulated evidence for point $i$ to choose point $j$ as its exemplar:

\begin{equation} \label{eq:alpha}
 \alpha_{ij}=\begin{cases}
               \sum_{k \ne j} \max(\rho_{kj},0) &\text{for} \  i = j\\
               \min(0,\rho_{jj} + \sum_{k\notin\{k,j\}}\max(\rho_{kj},0)) &\text{for} \ i \ne j
            \end{cases}
\end{equation}
\begin{equation} \label{eq:rho}
\rho_{ij}=s(i,j) - \max_{q \ne j} (s(i,j) + \alpha_{iq})
\end{equation}

\subsection{Affinity propagation clustering for multi-class object detection}

We introduce our novel Multi-class Affinity Propagation Clustering (MAPC) algorithm, \md{which extends SAPC \cite{eth_biwi_01126} from single-class to multi-class detection.}
\md{In multi-class detection} most object detectors propose multiple category labels with a certain confidence score for each bounding box. However, the label with the highest confidence is not always the correct one. Hence, \md{not only the correct location} but also the correct class for each box has to be inferred. \md{Therefore, we redefine each data point $i$ or $j$ as a combined box-class detection, e.g. \emph{$\text{box}_{1}$-dog}, \emph{$\text{box}_{1}$-cat}, or \emph{$\text{box}_{2}$-cat}. This allows us to define a similarity measure $s(i,j)$ between detections which includes both the spatial relation between bounding boxes and the relation between their labels} (\ref{eq:similarity1}):
\begin{equation} \label{eq:similarity1}
s(i,j) = \lambda \ \text{IoU}(\md{i},\md{j}) + (1 - \lambda) \ \text{sim}_{Lin}(\md{i},\md{j})
\end{equation}

\begin{figure}[t]
\begin{center}
\includegraphics[width=\linewidth,keepaspectratio]{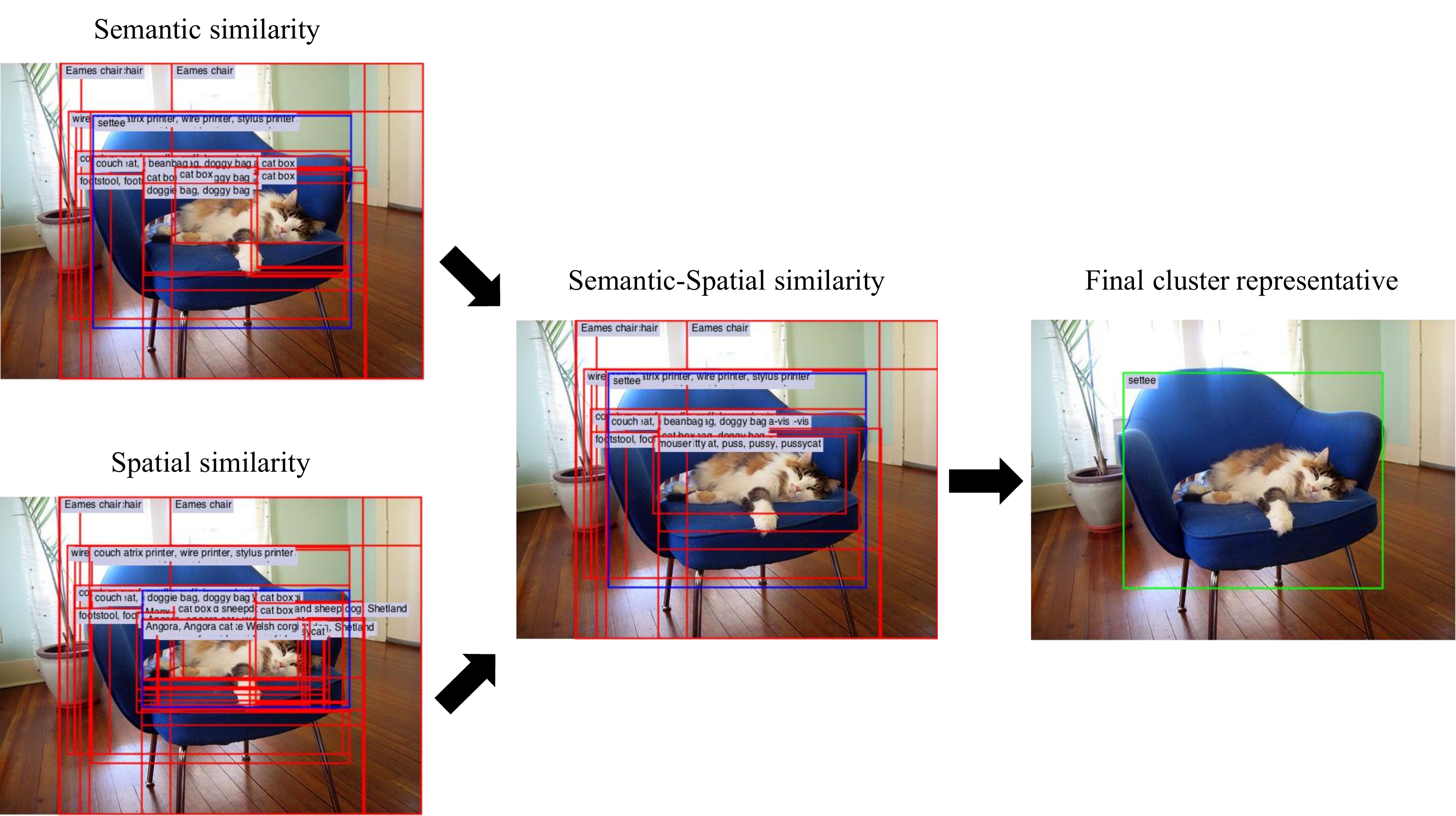}
\end{center}
   \caption{Combining spatial and semantic similarity in MAPC. All red boxes form one cluster \md{in which the blue box emerged as their exemplar. With a semantic-spatial similarity, semantically similar and spatially localised detections get clustered} which finally results in a well localised true positive detection.}
\label{fig:similaritymix}
\end{figure}

\md{Whereas SAPC bases its similarities solely on the IoU between bounding boxes \cite{eth_biwi_01126}, our} similarity measure clusters overlapping \md{detections}, represented by the $\text{IoU}(\md{i},\md{j})$ term, as well as semantically similar \md{detections}, represented by the $\text{sim}_{Lin}(\md{i},\md{j})$ term. An example can be seen in \Figureref{fig:similaritymix}. $\lambda$ is a weighting factor trading off spatial and semantic similarity. 
The Intersection over Union is defined as $ \text{IoU}(\md{i},\md{j}) = \frac{| \md{A_i} \cap \md{A_j} |}{| \md{A_i} \cup \md{A_j} |}$\md{, where $A_i$ is the area of the image covered by the bounding box of $i$.} It is used to describe the overlap and hence the visual similarity of two \md{detections}. 
The $Lin$ measure $\text{sim}_{Lin}(\md{i},\md{j}) = \frac{2∗IC(\md{lcs(C_i,C_j)})}{IC(\md{C_i})IC(\md{C_j})}$ denotes how semantically similar the labels of two \md{detections} are. \md{$lcs(C_i,C_j)$} denotes the lowest common subsumer of the \md{classes $C_i$ of $i$ and $C_j$ of $j$} in the WordNet hierarchy and $IC(\md{C})=\log p(\md{C})$ equals to the information content of a \md{class,} where $p(\md{C})$ is the probability of encountering an instance of the class $\md{C}$ in a corpus. The relative corpus frequency of $\md{C}$ and the probabilities of all \md{child classes} that $\md{C}$ subsumes are used to estimate the probability $p(\md{C})$ \cite{resnik1995using, rohrbach2011evaluating}. 

\md{The self-similarity is defined as $s(i,i) = -\frac{1}{p-\theta_{bg}}$, where $p$ is the detection score generated by the object detector and $\theta_{bg}$ is a background threshold used to discard detections scoring lower than $\theta_{bg}$ before APC inference.}

To further \md{avoid that contradicting detections are chosen as exemplars, we introduce} a new constraint: 
If \md{class $C_i$ is an exemplar for a specific box $k$ (i.e. $c_{ii}=1$), no other class can be an exemplar for box $k$:}
\md{
\begin{equation} \label{eq:noDouble}
    \widetilde{E}_{k}(c_{11},...,c_{NN})=\begin{cases}
               - \infty &\text{if} \ \sum_{j \text{ with box } k}c_{jj} > 1  \\
               0 &\text{otherwise}
            \end{cases}
\end{equation}
}

The remaining algorithm exactly follows \cite{eth_biwi_01126}, which uses a repellence term $R = \sum_{i \ne j} R_{ij}(c_{ii},c_{jj})$, but with $r(i,j) = -(s(\md{i},\md{j})+1)$ to avoid selecting semantic-spatially close exemplars, and a background category to allow for false positives to be suppressed, denoted by the $\widetilde{I}_{i}(c_{i1},...,c_{iN})$ term in equation (\ref{eq:newobjectivefunction}).
Linearly combining all of the terms presented yields in the following objective function to be maximised:
\begin{equation} \label{eq:newobjectivefunction}
\begin{aligned} 
\widetilde{E}_{APC} &= w_a\sum_{i}S_{ii} + w_b\sum_{i \ne j}S_{ij} + w_c\sum_{i}\widetilde{I}_{i} \\ &+ w_d\sum_{i<j}R_{ij} + w_e\sum_{j}E_{j} + w_f\md{\sum_{k}\widetilde{E}_{k}}
\end{aligned}
\end{equation}
\md{All function arguments in equation (\ref{eq:newobjectivefunction}) were left out for the sake of clarity.}
To solve this optimisation problem the message passing paradigm of \cite{eth_biwi_01126} is used.
All messages are initialised with zero and iteratively updated until convergence.

\section{Experiments}
\md{In this section we evaluate the performance of MAPC in a large scale setting. At this time, there is no standardised large scale dataset} with both a large amount of object instances within one image as well as a large amount of different object categories. \md{Hence, we evaluate MAPC on two different datasets. We use the Microsoft COCO dataset \cite{lin2014microsoft} for the evaluation on a large amount of object instances within one image. To evaluate on a large amount of fine-grained categories, we create a new dataset built of images with bounding box annotations from ImageNet \cite{deng2009imagenet}. This dataset covers 1,825 categories, but contains only a few object instances per image due to incomplete annotations.}

However, we believe that in a setting with both, thousands of fine-grained categories as well as dozens of object instances per image, our method would perform best. Hence, we also present qualitative results in the supplemental material, where we show the performance of our MAPC algorithm on all 7,404 LSDA categories \cite{Hoffman14Lsda}.

We mainly use precision and recall as well as the F1-score, which is the harmonic mean of  precision and recall, to evaluate MAPC on these datasets. The mAP metric, which is usually used to evaluate the performance on detection tasks, is not an appropriate performance measure for our multi-class detection setup.
\md{mAP is a metric for retrieval tasks. Traditionally, single-class detection has been seen as a retrieval task: all window detections that contain an object of the given class are to be retrieved. As most object detectors were designed as window-scoring methods it was obvious to rank all window detections according to their scores. With the clustering view, there is no absolute score which could be used for a global ranking and mAP can not be used correctly. The multi-class setting makes it even less suited.}
mAP favors multiple detections for each class and overall punishes across class selection of object proposals. In contrast, our method actually tries to provide a better way of selecting detections across classes. Hence, we can not use mAP to evaluate this task. For a true understanding of a depicted scene we have to focus especially on a high precision and F1-score for selecting object proposals across classes, while trying to maintain the recall. It is obvious that a high recall could also be achieved by selecting many object proposals without doing across class suppression. As can be seen in Figure 1(a) within class suppression alone---which would be desirable for the mAP measure---still leaves the question unanswered which objects \md{are actually depicted in an image. For a more detail investigation of wrong detections, we examine whether a false positive occurred due to a wrong localisation or classification. \emph{Wrong label} is the amount of all false positives with wrong labels of all false positives. \emph{Wrong overlap} is the amount of all false positives with a wrong location of all false positives.}

\md{To setup MAPC and determine all of its parameters, we use grid search on a training set obtained from ImageNet \cite{deng2009imagenet} as follows: First, we search for all ImageNet categories with available bounding box annotations. Next, we determine which of these categories overlap with the 7,404 categories of the LSDA detector \cite{Hoffman14Lsda}. This results in 1,825 categories with annotated images. Next, we discard all images used during the LSDA training and in the ImageNet test set described in section \ref{ImageNet2000}. We obtain our final training set by randomly selecting two annotated images per category from the remaining images. After performing grid search on this training set the MAPC parameters are set such that recall and precision are maximised.}

In all our experiments common non-maximum suppression (NMS) is used as the baseline. \md{More specifically,} detections of the same category overlapping more than a defined IoU threshold are suppressed \md{in a first step.} Then, all the remaining detections are suppressed across all classes with a different IoU threshold. Both NMS thresholds were determined using grid search as previously described. The best configuration resulted in a higher IoU threshold for within class suppression than for across class suppression. The intuition for this is that \md{detections} of the same class instance \md{are typically located at similar positions in the image}. Thus, in order to suppress \md{within} classes a higher threshold is necessary. \md{This baseline will be denoted as Within Class and Across Class NMS (WC+AC-NMS)}.
\md{MAPC is also compared against SAPC \cite{eth_biwi_01126}. However, SAPC was designed for single-class detection. As we evaluate in a multi-class detection scenario, we simply accumulate the per class output of SAPC across all classes for a first SAPC version. However, accumulating all detections without suppressing across classes is more suitable for an object retrieval task than for multi-class object detection. Thus, in a second version,} we use across class NMS \md{(AC-NMS)} on top of the accumulated SAPC output to \md{select object detections also across classes.} This makes \md{SAPC} \cite{eth_biwi_01126} better comparable to our method. The IoU threshold for this across class NMS was also determined using grid search.

\subsection{Multiple instance detection on COCO}
The Microsoft COCO dataset \cite{lin2014microsoft} consists of images that depict objects in their real world context rather than centered objects. Because of this, the detection on COCO is much more challenging than on the mostly centered ImageNet pictures. Hence, this dataset is chosen to evaluate the performance of our semantic spatial regulariser in a contextual setup with numerous object instances per image.
\subsubsection{Experimental setup}
COCO consists of 80 different categories with on average 7 object instances per image. \md{In a first experiment,} we use the latest LSDA setup with 7,404 fine-grained categories \cite{Hoffman14Lsda}. 15 COCO categories neither overlap with the leaf node categories of LSDA nor with either of their parents in the WordNet hierarchy\footnote{\emph{traffic light}, \emph{fire hydrant}, \emph{stop sign}, \emph{snowboard}, \emph{person}, \emph{kite}, \emph{fork}, \emph{sandwich}, \emph{hot dog}, \emph{pizza}, \emph{donut}, \emph{cake}, \emph{potted plant}, \emph{book}, \emph{teddy bear}}. \md{For those of the remaining 65 categories which overlap with a parent category, we use all of their children as an input to our method and the baselines.} For example we detect \emph{beagle} and \emph{dachshunds} instead of their parent category \emph{dog}. This results in 1,133 fine-grained child categories, which all methods have to infer on. We simply relabel the children output after inference to their parent categories to compare it with the COCO ground truth. We neither train LSDA nor adapt the MAPC paramters to COCO. 

\md{In a second experiment, we fine-tune our detection network on the COCO training set using all 80 COCO categories as input to our method and the baselines.}
Both experiments are evaluated on the COCO validation set.
\subsubsection{Experimental results}
Table \ref{table:COCOresults} shows the detection results \md{of our first experiment without finetuning our detector on COCO} on the COCO validation set. As can be seen MAPC outperforms WC+AC-NMS by 3.16\% in terms of precision when maintaining the recall. This performance gain can be explained by \md{less wrongly labeled (65.13\%) and wrongly localised (74.31\%) detections.} The F1 score for the chosen setup is 13.46\% for WC+AC-NMS versus 15.09\% for MAPC. \md{In \Figureref{fig:plot4}(c) \& (d) we vary the IoU evaluation threshold above which a detection is counted as a true positive. As can be seen MAPC is always better than WC+AC-NMS.} In general, almost all operating points of MAPC lie above WC+AC-NMS as can be seen in the precision-recall curve depicted in \Figureref{fig:plot4}(a). These results clearly show that MAPC is superior to WC+AC-NMS in scenarios with a lot of object instances per image.
Also when compared to SAPC \cite{eth_biwi_01126} our MAPC method shows an improvement over all numbers, except for the recall of 20.72\% since no across class suppression is applied. Hence, many detections are selected resulting in a cluttered outcome, which manifests in the low precision value of 5.25\% and decreases the F1-score to 8.38\%. As \cite{eth_biwi_01126} was designed for within class suppression and does not suppress across classes, these results are not surprising. However, when across class NMS (AC-NMS) is applied on the accumulated outcome of \cite{eth_biwi_01126} the precision increases to 14.66\% at the cost of a recall decrease. \md{Overall the F1-score increases to 13.12\%.} However, MAPC performs best on the COCO validation set amongst all tested methods.

\begin{figure}
\begin{center}
\begin{tabular}{c@{}c}
MAPC (optimal precision) \ & MAPC (optimal F1) \\
\hline
\hline
\includegraphics[trim = 45mm 10mm 45mm 0mm , clip=true,width=0.22\textwidth,height=\textheight,keepaspectratio]{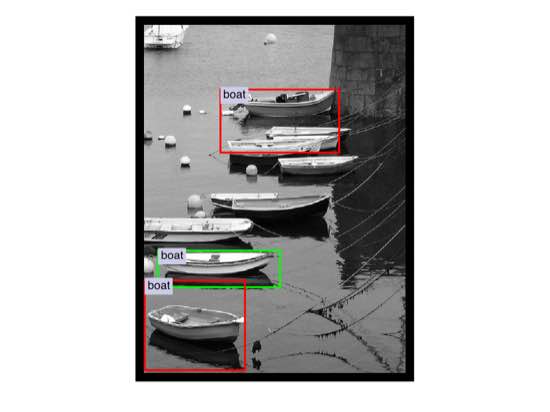} & 
\includegraphics[trim = 45mm 10mm 45mm 0mm , clip=true,width=0.22\textwidth,height=\textheight,keepaspectratio]{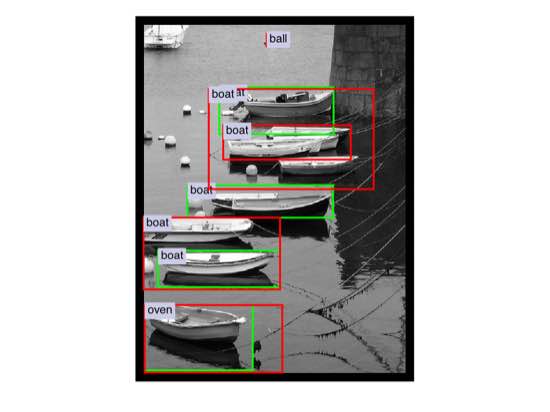} \\

\end{tabular}
\end{center}
\caption{Different optimisation criteria. When optimised for F1 score instead of precision, MAPC selects more detections, resulting in more true and false positives.}
\label{table:COCOpictures4}
\end{figure}

The greater precision of MAPC can be especially seen when we look at example images. The pictures in \Figureref{table:COCOpictures2} show the output of WC+AC-NMS and MAPC after optimising both algorithms for the highest precision with comparable recall. \md{The detector not fine-tuned on COCO was used.} Green boxes are true positive detections. Red boxes are false positive detections. WC+AC-NMS reaches its precision limit after suppressing all overlapping boxes, while MAPC can also suppress non-overlapping boxes. At the same time, MAPC still enables the selection of overlapping object proposals as can be clearly seen in the example pictures. Allowing a greater overlap for WC+AC-NMS would increase true positives at the cost of lower precision and a cluttered detection output. In general, MAPC outputs less false positives and better localised true positives. 

If required, MAPC can also be optimised towards a higher recall. Figure \ref{table:COCOpictures4} examplarily compares a F1 score optimised MAPC to a precision optimised MAPC. Clearly more \emph{boats} get detected when we optimise towards F1, but also more false positives are selected. All in all, MAPC can be optimised towards a high recall and a high precision, while WC+AC-NMS reaches its precision limit when trying to suppress non overlapping boxes. Thus, MAPC can be preciser in selecting the correct bounding box proposals.

\md{In our second experiment, we fine-tune our object detector on COCO. The results can be seen in table \ref{table:COCOresults2}. As expected all of our metrics highly improve. Most striking the MAPC precision rises to 37.64\%, while the recall remains comparable, which increases the F1 score difference between MAPC and WC+AC-NMS to 5.40\%. Also the F1 score of SAPC strongly improves to 21.17\%. All methods obviously greatly profit from better detections. Thus, a detector which provides good candidate detections in the first place is crucial for all of the examined methods.}

\begin{table} 
\begin{center}
\footnotesize
\begin{tabular}{|l|c|c|c|c|c|c|}
\hline
Method & Pre- & Re- & Wrong & Wrong & F1 \\
 & cision & call & Label & Overlap & Score \\
\hline\hline
WC+AC-NMS & 13.44 & 13.47 & 79.39 & 88.97 & 13.46 \\
SAPC \cite{eth_biwi_01126} & 5.25 & 20.72 & 74.79 & 72.73 & 8.38 \\
SAPC + AC-NMS & 14.66 & 11.86 & 81.36 & 87.15 & 13.12 \\
MAPC (ours) & 16.60 & 13.84 & 65.13 & 74.31 & 15.09 \\
\hline
\end{tabular}
\end{center}
\caption{Detection results on COCO without finetuning, in \%.}
\label{table:COCOresults}

\begin{center}
\footnotesize
\color{black}
\begin{tabular}{|l|c|c|c|c|c|c|}
\hline
Method & Pre- & Re- & Wrong & Wrong & F1 \\
 & cision & call & Label & Overlap & Score \\
\hline\hline
WC+AC-NMS & 23.50 & 24.80 & 62.99 & 94.97 & 24.10 \\
SAPC \cite{eth_biwi_01126} & 15.66 & 32.61 & 69.01 & 72.43 & 21.17 \\
SAPC + AC-NMS & 30.01 & 21.97 & 74.95 & 92.90 & 25.39 \\
MAPC (ours) & 37.64 & 24.23 & 55.47 & 71.79 & 29.50 \\
\hline
\end{tabular}
\end{center}
\caption{\md{Detection results on COCO, fine-tuned on COCO, in \%.}}
\label{table:COCOresults2}

\begin{center}
\footnotesize
\begin{tabular}{|l|c|c|c|c|c|c|}
\hline
Method & Pre- & Re- & Wrong & Wrong & F1 \\
 & cision & call & Label & Overlap & Score \\
\hline\hline
WC+AC-NMS & 8.34 & 11.29 & 91.90 & 85.53 & 9.59 \\	
SAPC \cite{eth_biwi_01126} & 3.46 & 22.57 & 93.69 & 68.05 & 6.00 \\
SAPC + AC-NMS & 9.76 & 10.34 & 91.02 & 81.54 & 10.04 \\
MAPC (ours) & 10.94 & 16.22 & 86.41 & 68.57 & 13.07 \\
\hline
\end{tabular}
\end{center}
\caption{Detection results on ImageNet without finetuning, in \%.}
\label{table:ImageNetresultsTop1}

\end{table} 

\begin{figure*}[!t]
\centering

\subfigure[]{
\includegraphics[width=150px]{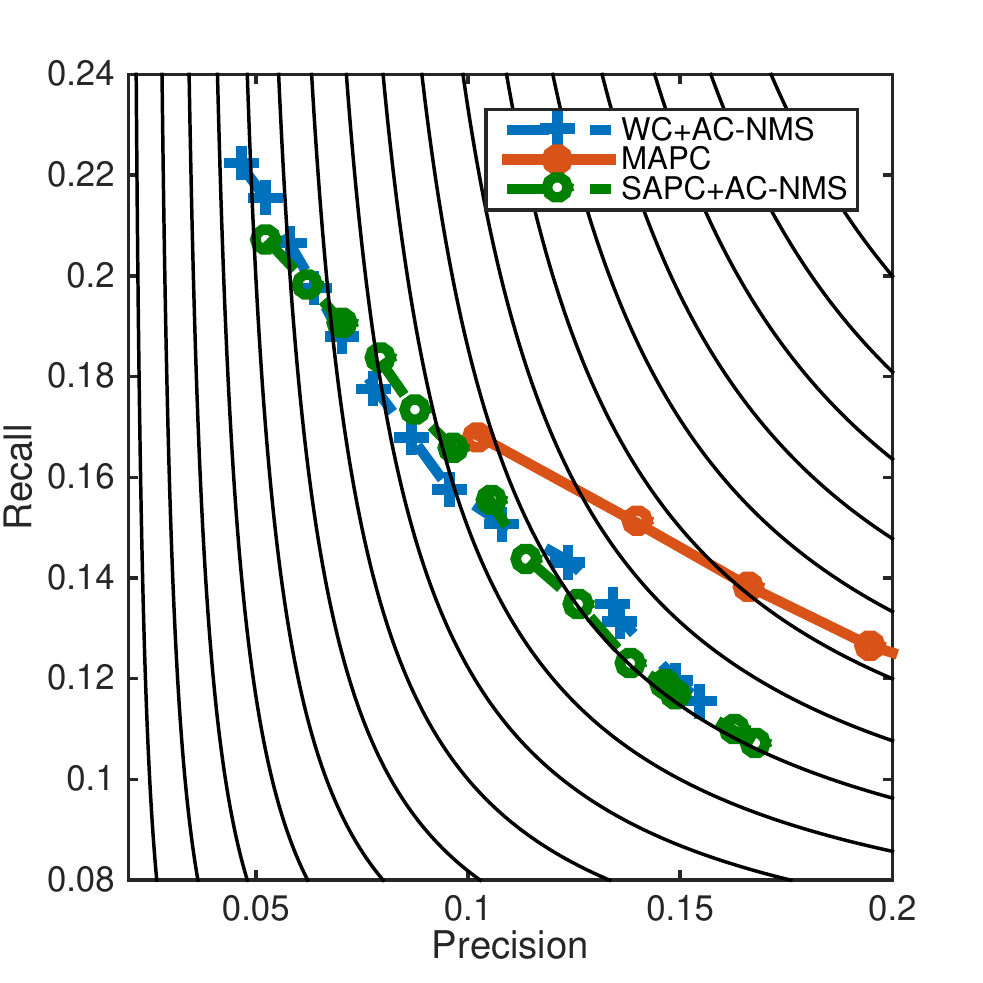}
}
\subfigure[]{
\includegraphics[width=150px]{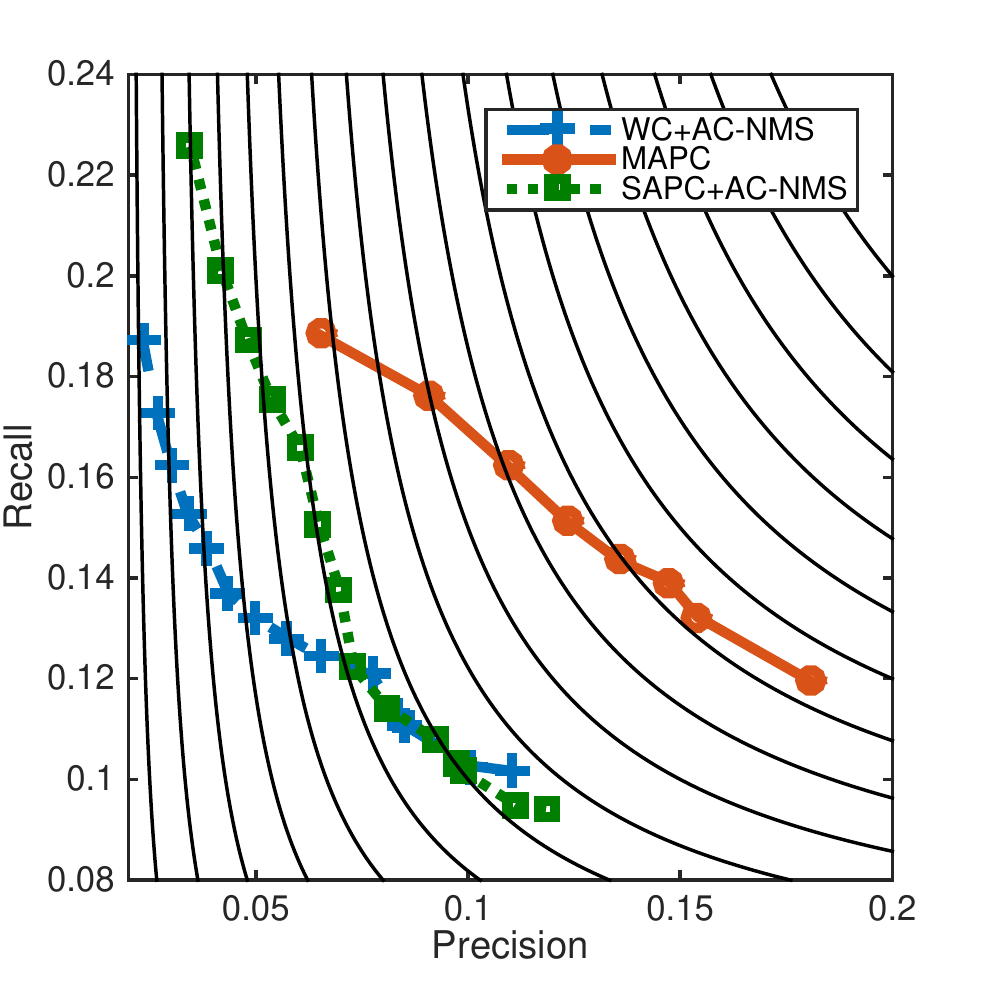}
}
\begin{minipage}[b]{0.25\textwidth}
\subfigure[]{
\includegraphics[width=115px]{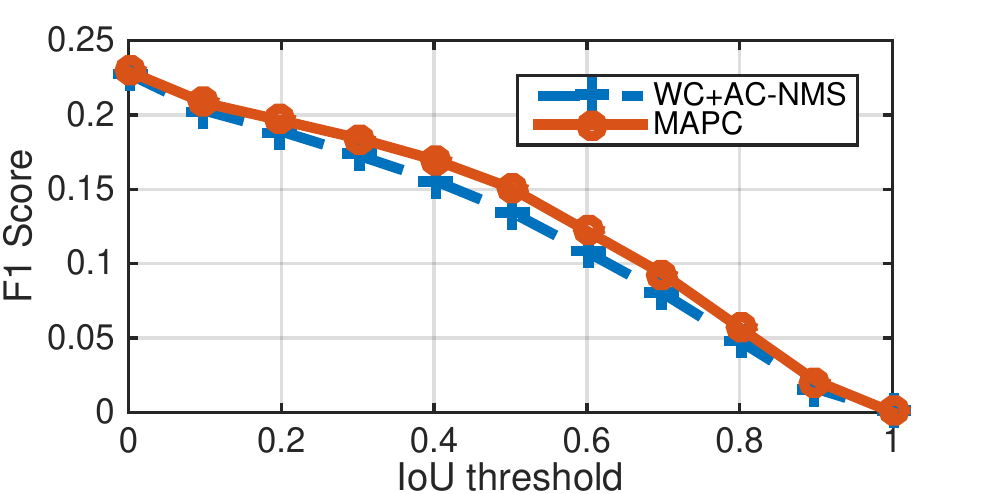}
} \\
\subfigure[]{
\includegraphics[width=115px]{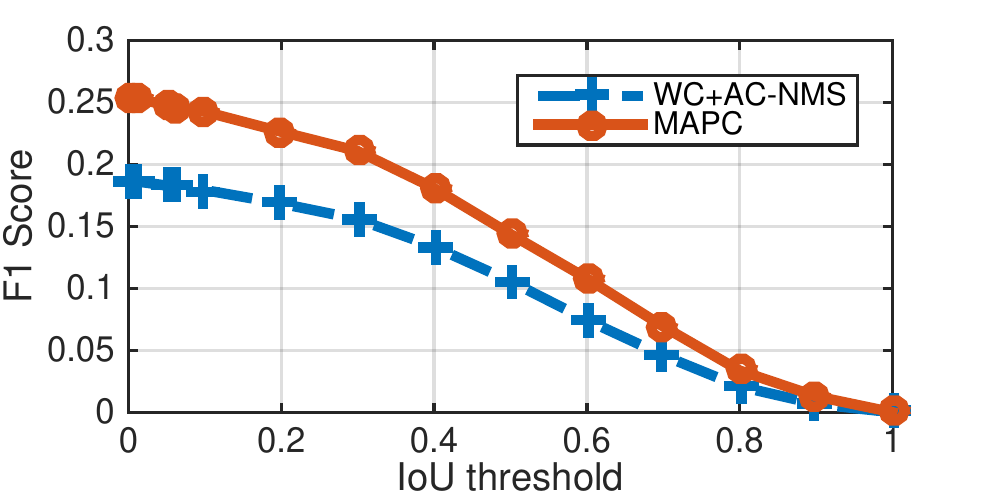}
}
\end{minipage}

\caption{\% \& (b) Precision-recall curve for WC+AC-NMS, SAPC + AC-NMS and MAPC on COCO (a) and on the set of 1,825 ImageNet categories (b) without finetuning. The curved lines mark points of equal F1 score. The F1 score increases from lower left to upper right. \md{Multiple operating points are obtained by varying the across class IoU threshold in AC-NMS (for WC+AC-NMS and SAPC + AC-NMS) and $w_{a}$ for MAPC.} MAPC consistently outperforms WC+AC-NMS and SAPC + AC-NMS on COCO and ImageNet. SAPC + AC-NMS is superior to WC+AC-NMS in the lower precision range. 
(c) \& (d) F1-score plotted against IoU evaluation threshold for COCO (c) and on the set of 1,825 ImageNet categories (d). MAPC consistently outperforms WC+AC-NMS on both datasets.}

\label{fig:plot4}
\end{figure*}

\subsection{Fine-grained multi-class detection on ImageNet}
\label{ImageNet2000}
In this section we evaluate MAPC on a large scale multi-class detection setup constructed from ImageNet data \cite{deng2009imagenet}. \md{Since there is no standardised dataset with thousands of categories, we construct our own dataset to evaluate MAPC on a large amount of fine-grained categories. The final dataset covers 1,825 categories, but only a few object instances per image due to incomplete annotations of ImageNet.}
\subsubsection{Experimental setup}
\md{In order to construct a dataset with numerous fine-grained categories, we search for all ImageNet categories with available detection annotations. As we use the LSDA detector \cite{Hoffman14Lsda}, we determine which of these categories overlap with its 7,404 categories. This results in 1,825 categories with annotated images. Next, all images used during the training of the LSDA detector are discarded.} As most of the remaining images have only one object annotated, we further restrict our test set to images with at least two annotated objects. This way, we ensure that we evaluate on a true detection setup rather than a localisation setup. \md{After this step, we obtain our final fine-grained ImageNet test set.}

\subsubsection{Experimental results}

\begin{figure*}
\begin{center}
\begin{tabular}{c@{}c|c@{}c}
WC+AC-NMS & MAPC & WC+AC-NMS & MAPC \\
\hline\hline
\includegraphics[trim = 14mm 20mm 14mm 10mm , clip=true,width=0.25\textwidth,height=\textheight,keepaspectratio]{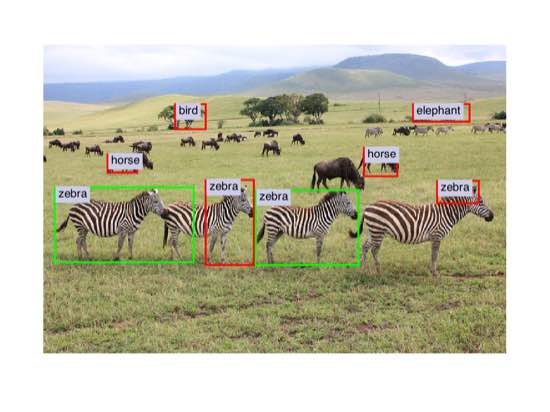} & 
\includegraphics[trim = 14mm 20mm 14mm 10mm , clip=true,width=0.25\textwidth,height=\textheight,keepaspectratio]{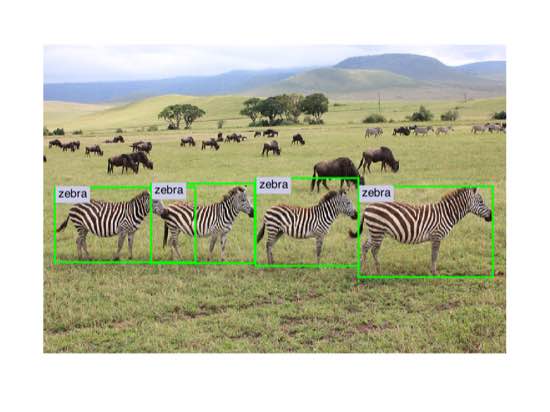} &

\includegraphics[trim = 14mm 20mm 14mm 10mm , clip=true,width=0.25\textwidth,height=\textheight,keepaspectratio]{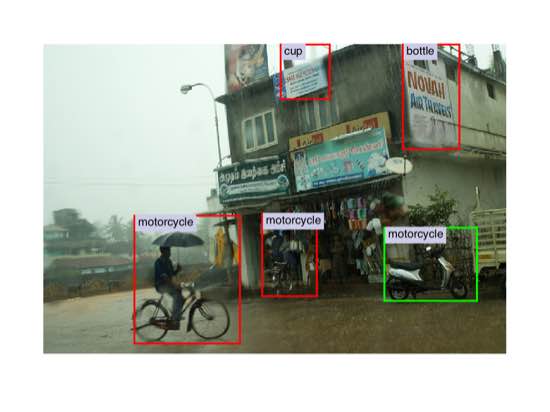} & 
\includegraphics[trim = 14mm 20mm 14mm 10mm , clip=true,width=0.25\textwidth,height=\textheight,keepaspectratio]{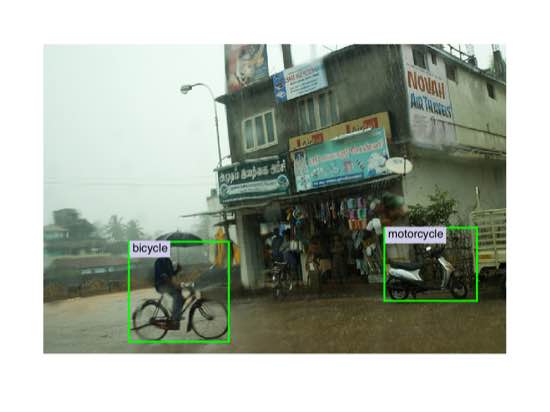} \\

\includegraphics[trim = 18mm 20mm 18mm 15mm , clip=true,width=0.25\textwidth,height=\textheight,keepaspectratio]{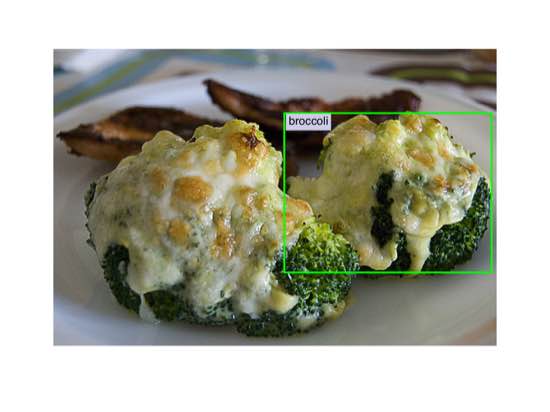} & 
\includegraphics[trim = 18mm 20mm 18mm 15mm , clip=true,width=0.25\textwidth,height=\textheight,keepaspectratio]{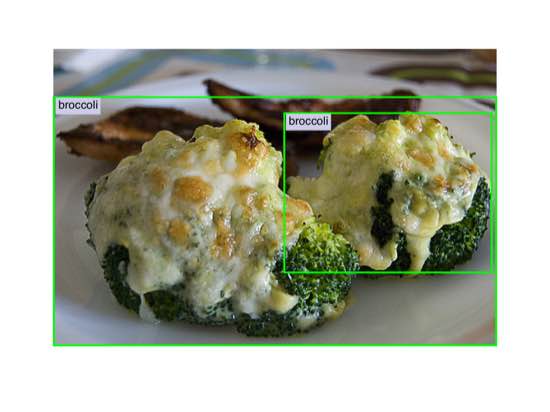} &

\includegraphics[trim = 14mm 20mm 14mm 15mm , clip=true,width=0.25\textwidth,height=\textheight,keepaspectratio]{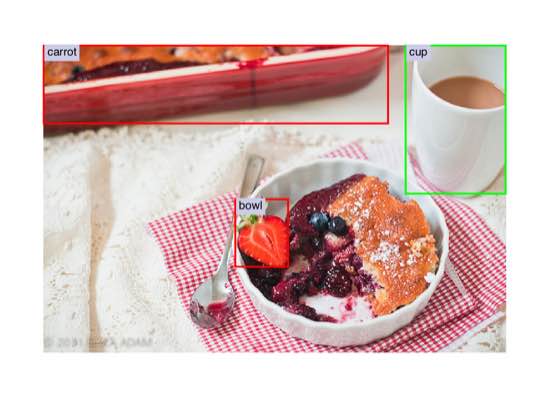} & 
\includegraphics[trim = 14mm 20mm 14mm 15mm , clip=true,width=0.25\textwidth,height=\textheight,keepaspectratio]{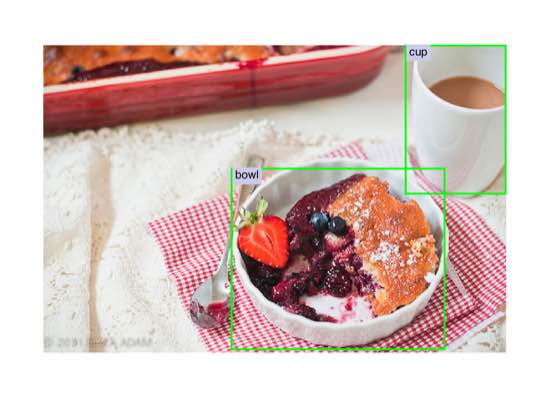} \\

\includegraphics[trim = 18mm 20mm 18mm 10mm , clip=true,width=0.25\textwidth,height=\textheight,keepaspectratio]{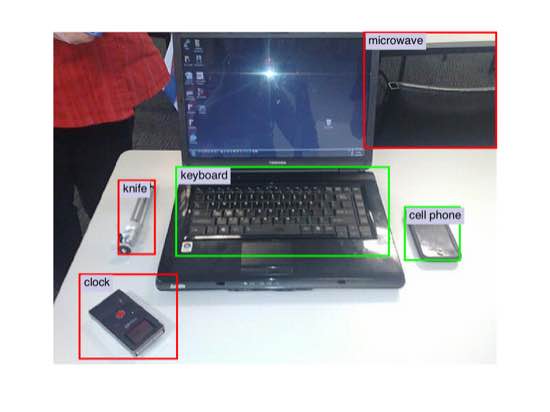} & 
\includegraphics[trim = 18mm 20mm 18mm 10mm , clip=true,width=0.25\textwidth,height=\textheight,keepaspectratio]{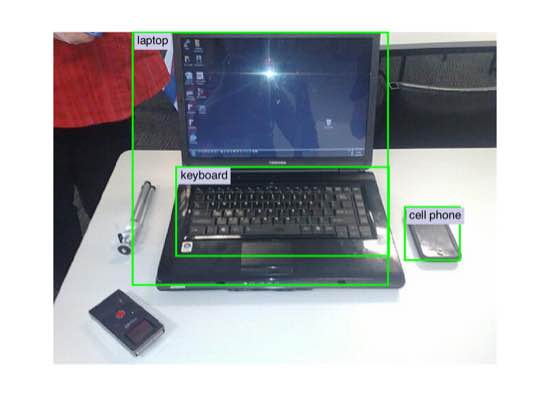} &

\includegraphics[trim = 14mm 20mm 14mm 10mm , clip=true,width=0.25\textwidth,height=\textheight,keepaspectratio]{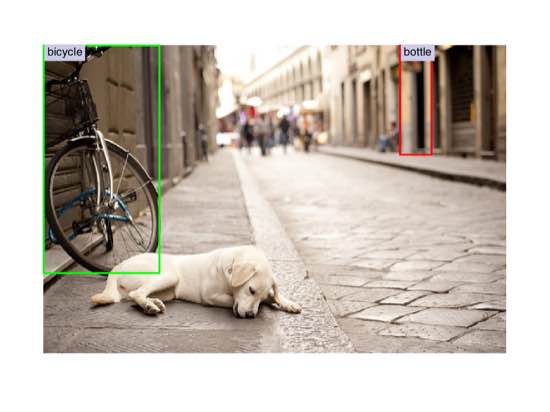} & 
\includegraphics[trim = 14mm 20mm 14mm 10mm , clip=true,width=0.25\textwidth,height=\textheight,keepaspectratio]{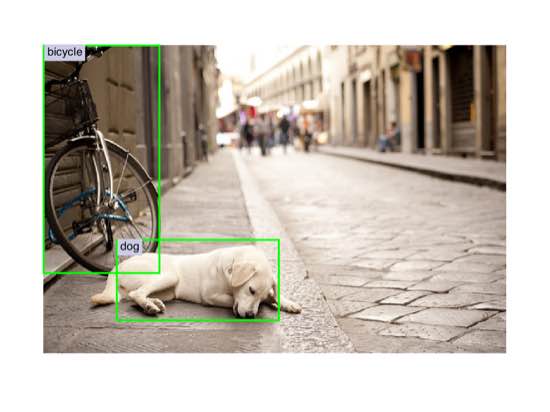} \\

\includegraphics[trim = 14mm 20mm 14mm 10mm , clip=true,width=0.25\textwidth,height=\textheight,keepaspectratio]{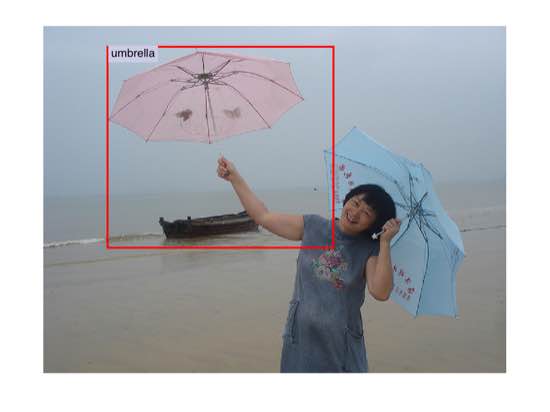} & 
\includegraphics[trim = 14mm 20mm 14mm 10mm , clip=true,width=0.25\textwidth,height=\textheight,keepaspectratio]{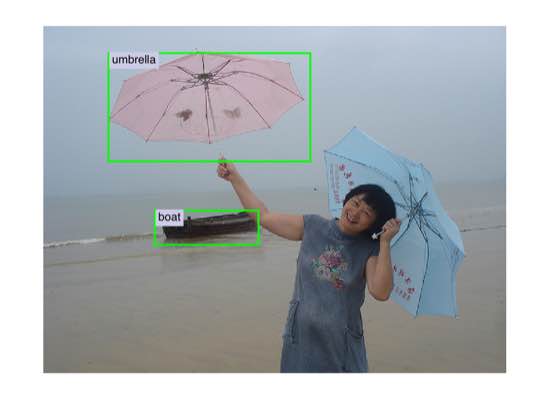} &

\includegraphics[trim = 14mm 20mm 14mm 10mm , clip=true,width=0.25\textwidth,height=\textheight,keepaspectratio]{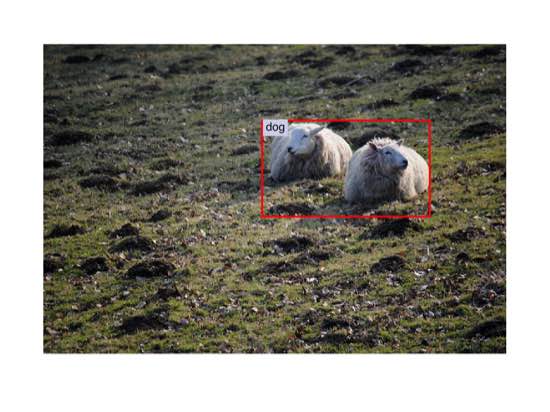} & 
\includegraphics[trim = 14mm 20mm 14mm 10mm , clip=true,width=0.25\textwidth,height=\textheight,keepaspectratio]{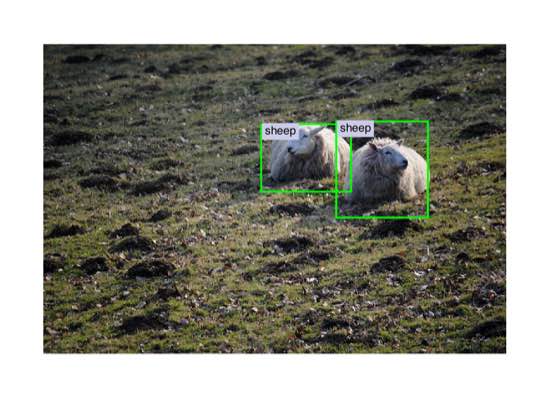} \\

\end{tabular}
\end{center}
\caption{Examples, where MAPC outperforms WC+AC-NMS on Microsoft COCO. True positives: green, false positives: red.}
\label{table:COCOpictures2}
\vspace{0.5cm}
\begin{center}
\begin{tabular}{c@{}c@{}c}
Ground Truth & WC+AC-NMS & MAPC \\
\hline\hline

\includegraphics[trim = 14mm 20mm 14mm 20mm , clip=true,width=0.25\textwidth,height=\textheight,keepaspectratio]{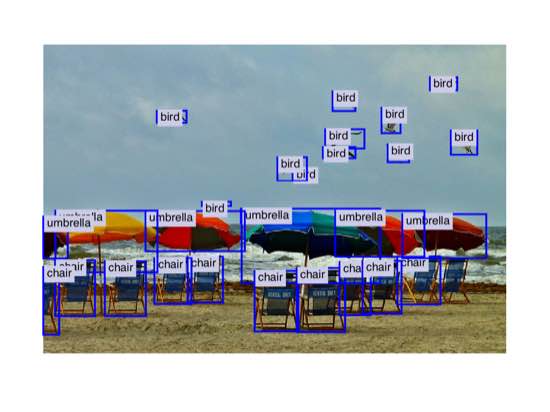} & 
\includegraphics[trim = 14mm 20mm 14mm 20mm , clip=true,width=0.25\textwidth,height=\textheight,keepaspectratio]{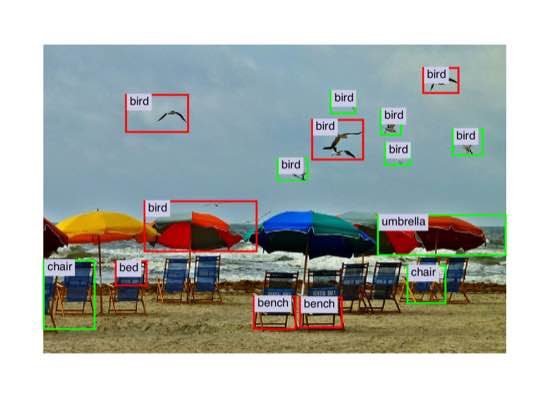} & 
\includegraphics[trim = 14mm 20mm 14mm 20mm , clip=true,width=0.25\textwidth,height=\textheight,keepaspectratio]{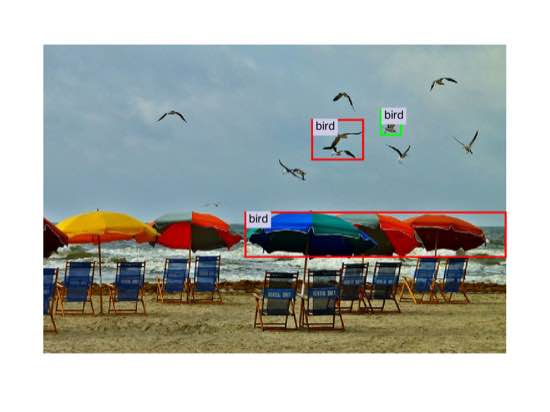} \\

\includegraphics[trim = 14mm 20mm 14mm 20mm , clip=true,width=0.25\textwidth,height=\textheight,keepaspectratio]{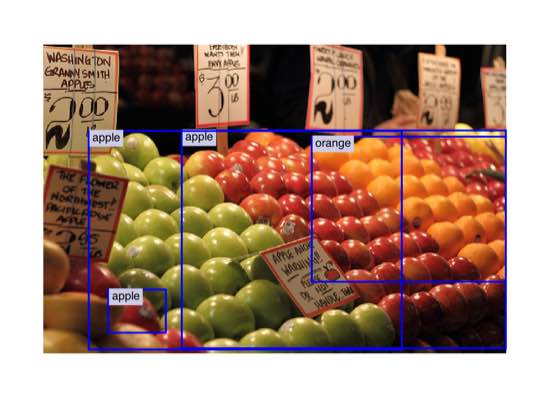} & 
\includegraphics[trim = 14mm 20mm 14mm 20mm , clip=true,width=0.25\textwidth,height=\textheight,keepaspectratio]{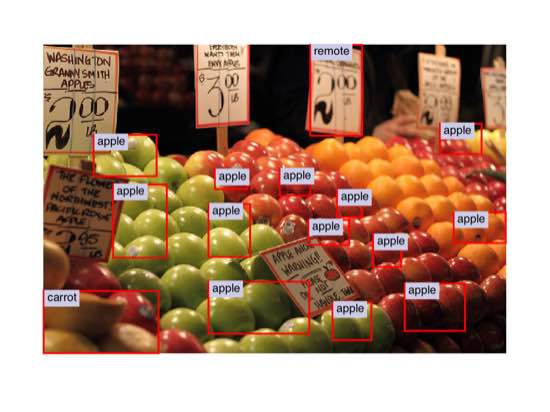} & 
\includegraphics[trim = 14mm 20mm 14mm 20mm , clip=true,width=0.25\textwidth,height=\textheight,keepaspectratio]{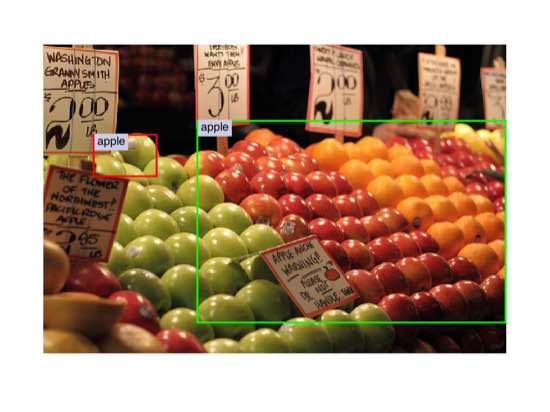} \\

\end{tabular}
\end{center}
\caption{Examples where WC+AC-NMS outperforms MAPC on Microsoft COCO. Ground truth: blue, true positives: green, false positives: red.}
\label{table:COCOpictures3}
\end{figure*}

Table \ref{table:ImageNetresultsTop1} shows the detection results on our large scale ImageNet detection dataset. The same tendencies as on COCO can also be observed here. The precision and recall for MAPC are 2.60\% and 4.93\% higher than for WC+AC-NMS. The F1-score increases from 9.59\% to 13.07\%. False positives due to wrong labels drop by 5.49\% and localisation errors drop from 85.53\% to 68.57\%. Again SAPC performs better after applying across class NMS. However, MAPC still performs best, which confirms our results on COCO. What is striking however is that the improvement to WC+AC-NMS and SAPC + AC-NMS is bigger in the fine-grained setting. Also the improvement of MAPC over WC+AC-NMS, when the IoU evaluation threshold is varied, is bigger than on COCO, which can be seen in \Figureref{fig:plot4}. It seems that the more fine-grained the categories, the more visually similar are semantically similar categories, and thus, the more useful the label relations from the WordNet are. This indicates that our approach is especially useful in a large scale setting when a lot of visually similar fine-grained object categories are competing against each other.

\section{Conclusions and Future Work}
We presented MAPC, a large scale multi-class regulariser which globally maximises both the semantic and spatial similarity, and thus, visual similarity of clusters of detection proposals. MAPC reduces false positives significantly in multi-class detection, resulting in an improved classification and localisation.
Our results show that the selection of detection proposals can be significantly improved over baseline class-independent non-maximum suppression by formulating a clustering problem across class labels and spatial dimensions, which can be solved by affinity propagation.
Overall, we consistently improve precision and recall for different operating points and evaluation thresholds.

As future work, it would be interesting to compare the fine-grained category detection on COCO with detectors trained on all parent categories to see whether training more fine-grained classes to detect the actual parent class helps the detection of objects. MAPC could also be extended to the temporal domain, in order to cluster over consecutive video frames for activity recognition and video description.

\vspace{1cm}

\emph{\textbf{Acknowledgements.} This work was supported by DARPA, AFRL, DoD MURI award N000141110688, NSF awards IIS-1427425, IIS-1212798, and IIS-113629, and the Berkeley Vision and Learning Center. Marcus Rohrbach was supported by a fellowship within the FITweltweit-Program of the German Academic Exchange Service (DAAD).}

{\small
\bibliographystyle{ieee}
\bibliography{egbib}
}

\vspace{0.5cm}
{\Large \centering \centerline{\textbf{Supplementary Material}}}
\vspace{1cm}

\maketitle
\thispagestyle{empty}

\begin{abstract}
This document accompanies the paper: "Spatial Semantic Regularisation for Large Scale Object Detection". It contains an additional detection experiment on COCO, where we use the VGG net \cite{simonyan2014very} in RCNN \cite{girshick2014rich}, to examine the influence of the neural network architecture on our approach. Further, it comprises additional exemplary detection results to show the key differences of multi-class affinity propagation clustering (MAPC) to single-class affinity propagation clustering (SAPC) \cite{eth_biwi_01126} and non-maximum suppression (NMS). Results are shown using the COCO classes, the 1,825 ImageNet classes as mentioned in the main paper and all 7,404 fine-grained classes of the LSDA large-scale detector \cite{Hoffman14Lsda}. The example images show that MAPC both localises and classifies objects in images better than SAPC + AC-NMS and WC+AC-NMS. We also show some failure cases of MAPC. We further break down the iterations of our algorithm on one example for a detailed step-by-step explanation of MAPC. Code will be published on our website.
\end{abstract}

\section{Multiple Instance Detection on COCO based on VGG net} 

In order to examine the influence of the network architecture on the selection process, we conduct a third experiment on the COCO dataset, additionally to the two experiments mentioned in the main paper. As in the second experiment, we fine-tune our detection network on the COCO training set using all 80 COCO categories as input to our method and the baselines. However, we replace the original RCNN detection network architecture \cite{girshick2014rich} by VGG net, a deeper neural network architecture, which has shown to significantly improve classification and detection performance \cite{simonyan2014very}. We evaluate our method against the baselines using the output of fine-tuned VGG net as the input to all regularisation methods and
obtain the results depicted in Table \ref{table:COCOresults3}.

\begin{table*}
\begin{center}
\begin{tabular}{|l|c|c|c|c|c|c|}
\hline
Method & Precision & Recall & Wrong Label & Wrong Overlap & F1 Score \\
\hline\hline
WC+AC-NMS & 11.26 & 29.41 & 51.28 & 98.42 & 16.29 \\
SAPC (\cite{eth_biwi_01126}) & 21.11 & 32.39 & 80.71 & 76.95 & 25.56 \\
SAPC + AC-NMS & 32.96 & 23.84 & 82.93 & 93.19 & 27.67 \\
MAPC (ours) & 37.23 & 31.21 & 55.18 & 75.71 & 33.96 \\
\hline
\end{tabular}
\end{center}
\caption{Detection results on COCO, fine-tuned on COCO using VGG net, in
percent. Within Class and Across Class NMS (WC+AC-NMS), Single-class APC
(SAPC), Single-class APC and Across Class NMS (SAPC + AC-NMS), are compared
against Multi-class APC (MAPC).}
\label{table:COCOresults3}
\end{table*}
\begin{figure*}
\begin{center}
\begin{tabular}{c@{}c@{}c@{}c}
Ground Truth & WC+AC-NMS & SAPC + AC-NMS & MAPC (ours) \\
\hline\hline

\includegraphics[trim = 14mm 20mm 14mm 10mm ,
clip=true,width=0.25\textwidth,height=\textheight,keepaspectratio]{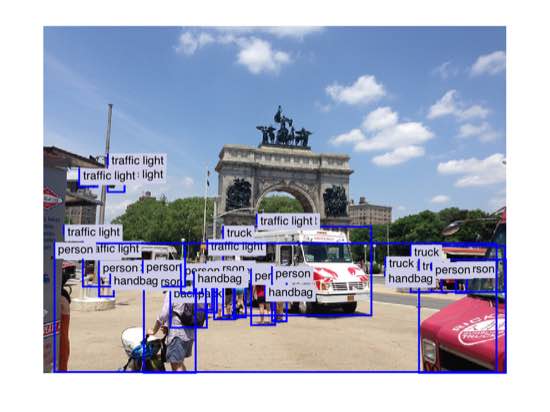}
& \includegraphics[trim = 14mm 20mm 14mm 10mm ,
clip=true,width=0.25\textwidth,height=\textheight,keepaspectratio]{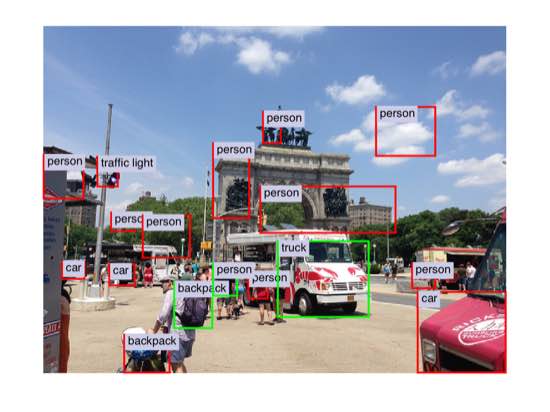}
& \includegraphics[trim = 14mm 20mm 14mm 10mm ,
clip=true,width=0.25\textwidth,height=\textheight,keepaspectratio]{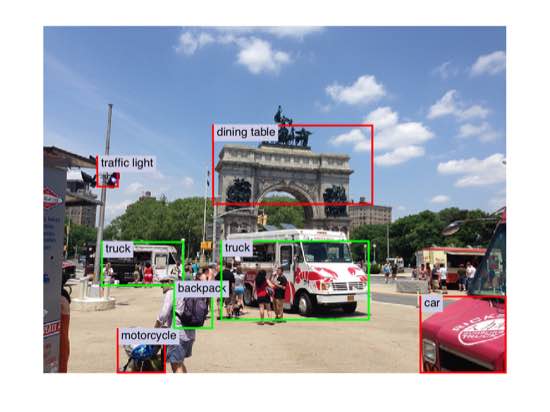}
& \includegraphics[trim = 14mm 20mm 14mm 10mm ,
clip=true,width=0.25\textwidth,height=\textheight,keepaspectratio]{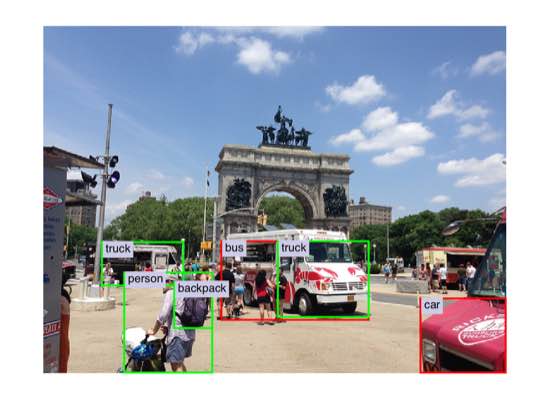}
\\

\includegraphics[trim = 14mm 30mm 14mm 10mm ,
clip=true,width=0.25\textwidth,height=\textheight,keepaspectratio]{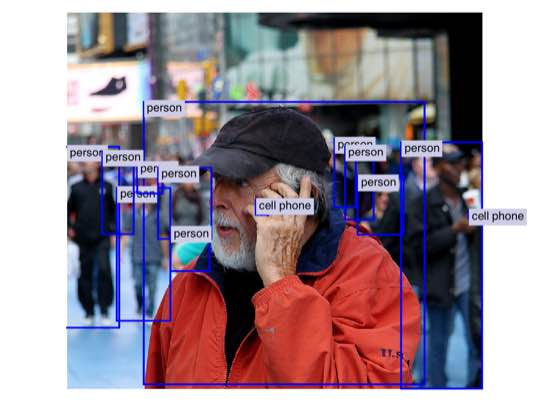}
& \includegraphics[trim = 14mm 30mm 14mm 10mm ,
clip=true,width=0.25\textwidth,height=\textheight,keepaspectratio]{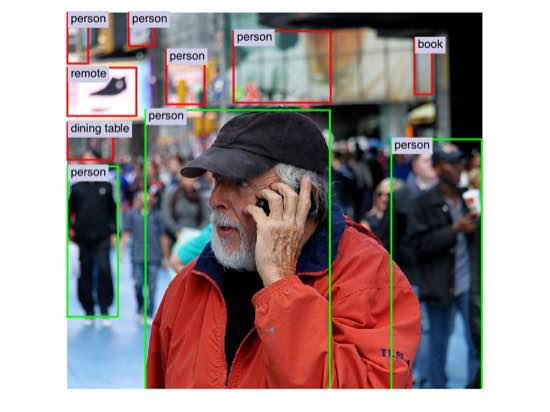}
& \includegraphics[trim = 14mm 30mm 14mm 10mm ,
clip=true,width=0.25\textwidth,height=\textheight,keepaspectratio]{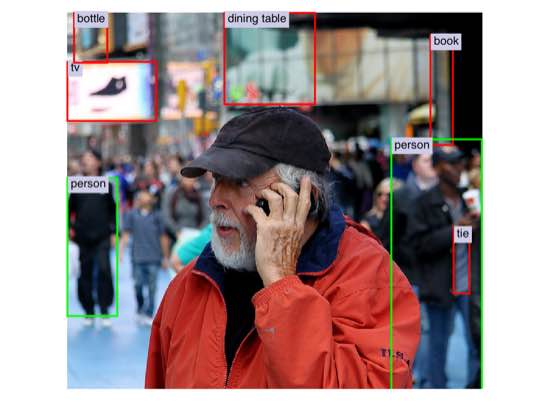}
& \includegraphics[trim = 14mm 30mm 14mm 10mm ,
clip=true,width=0.25\textwidth,height=\textheight,keepaspectratio]{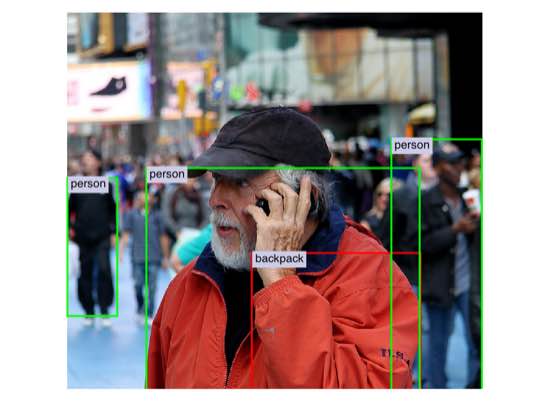}
\\

\includegraphics[trim = 14mm 25mm 14mm 20mm ,
clip=true,width=0.25\textwidth,height=\textheight,keepaspectratio]{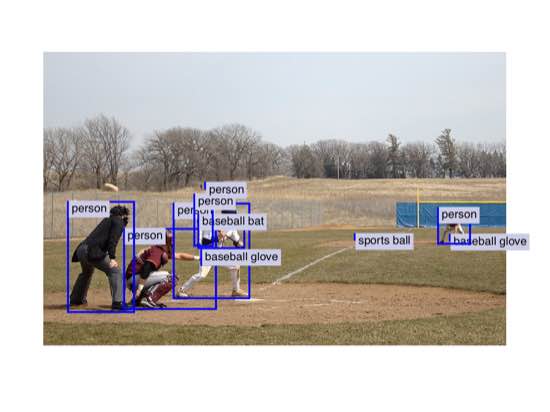}
& \includegraphics[trim = 14mm 25mm 14mm 20mm ,
clip=true,width=0.25\textwidth,height=\textheight,keepaspectratio]{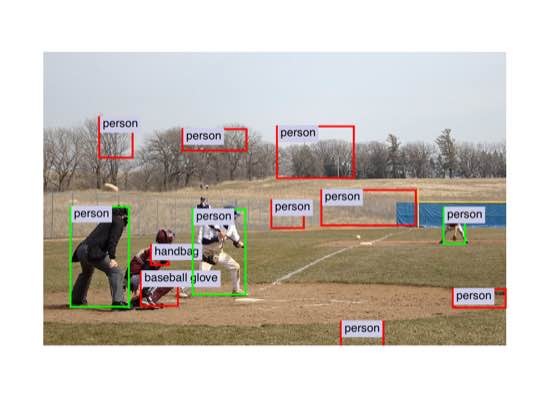}
& \includegraphics[trim = 14mm 25mm 14mm 20mm ,
clip=true,width=0.25\textwidth,height=\textheight,keepaspectratio]{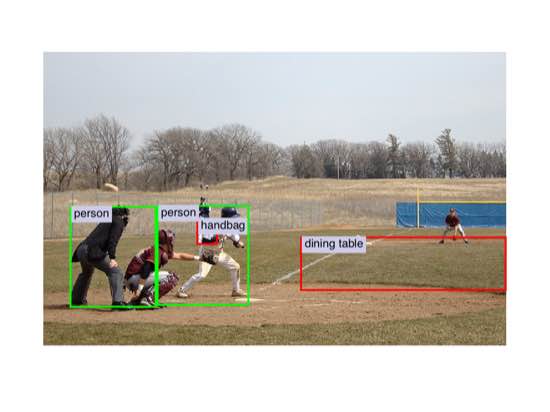}
& \includegraphics[trim = 14mm 25mm 14mm 20mm ,
clip=true,width=0.25\textwidth,height=\textheight,keepaspectratio]{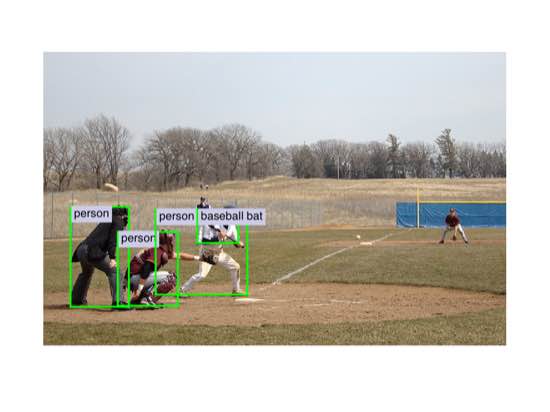}
\\
\end{tabular}
\end{center}
\caption{Detection examples when using VGG net with WC+AC-NMS, SAPC + AC-NMS and MAPC on Microsoft COCO. Ground truth: blue, true positives: green, false positives:
red.}
\label{fig:COCOresults3}
\end{figure*}

When we compare Table \ref{table:COCOresults3} with the COCO tables from the main paper, we can clearly see that all methods benefit from the deeper VGG net except WC+AC-NMS. Compared to our second COCO experiment the F1 score of SAPC improves from 21.17\% to 25.56\%, SAPC + AC-NMS improves from 25.39\% to 27.67 \% and MAPC improves from 29.50\% to 33.96\%. Only WC+AC-NMS drops from 24.10\% to 16.29
\%. Whereas the recall and precision of all other methods increases or at least stays the same, the precision of WC+AC-NMS decreases significantly. The obtained numbers by itself are not able to explain this behavior. 
However, a qualitative analysis of several obtained detection results gives an explanation for the drop in precision of WC+AC-NMS. When we look at the detection results of WC+AC-NMS in the images of Figure \ref{fig:COCOresults3}, we can see that a lot of small non-overlapping false positive detections are obtained. VGG learns better than the original RCNN network, that the COCO dataset contains a lot of small objects, and thus, scores object that are usually small in the dataset, such as groups of \emph{persons}, higher. This creates a lot of small non-overlapping detection proposals scattered all over the image. In these cases, NMS simply cannot suppress the false positive detections, since they do not overlap with any higher scoring detection. SAPC and MAPC however do not need false positive detections to overlap to be able to suppress them. Thus, both APC methods are able to suppress more false positive detections than NMS, which suppression ability is limited by the necessity of overlapping detection proposals. This results in an upper boundary for the precision of NMS. Hence, especially when using the VGG net, the better precision of our MAPC method compared to NMS gets obvious. But also in comparison to SAPC and SAPC + AC-NMS, MAPC has a significantly higher F1 score. When using the VGG net, the gap between MAPC and the two SAPC approaches gets even bigger. In the end, MAPC improves the detection performance in terms of the F1 score by 6.29\%, when compared to SAPC + WC-NMS, and even by 8.40\%, when compared to the original SAPC method. Summarising, our results clearly show that MAPC is able to improve the detection performance over state-of-the-art methods in terms of precision, while maintaining the recall.

\section{Comparison of MAPC, SAPC and NMS on COCO and ImageNet}
In this section we show and discuss example images, where our MAPC method
outperforms the SAPC and NMS baselines. We use the original LSDA network not fine-tuned to the COCO dataset. All regularisation methods were optimized to maximize the F1-score. The images were not selected randomly, but chosen based on the best performance of each method.

We evaluate on images from COCO and ImageNet on all categories of the 7,404 categories of the LSDA large scale detector \cite{Hoffman14Lsda}, which overlap with the respective dataset. Following the experimental setup in the main paper, we evaluate on 65 categories, i.e. on their 1,133 child categories, for the COCO validation set and on 1,825 categories for the ImageNet 1,825 categories set. Since a dataset with ground truth annotations for all 7,404 LSDA categories is missing, we also present qualitative MAPC results on all 7,404 categories of the LSDA detector. We evaluate against the WC+AC-NMS and the SAPC + AC-NMS baseline as defined in the main paper.

\begin{figure*}
\begin{center}
\begin{tabular}{c@{}c@{}c@{}c}
Ground Truth & WC+AC-NMS & SAPC + AC-NMS & MAPC (ours) \\
\hline\hline

\includegraphics[trim = 14mm 20mm 14mm 10mm , clip=true,width=0.25\textwidth,height=\textheight,keepaspectratio]{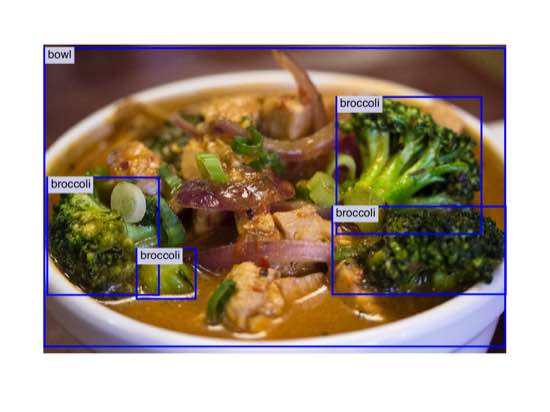} & 
\includegraphics[trim = 14mm 20mm 14mm 10mm , clip=true,width=0.25\textwidth,height=\textheight,keepaspectratio]{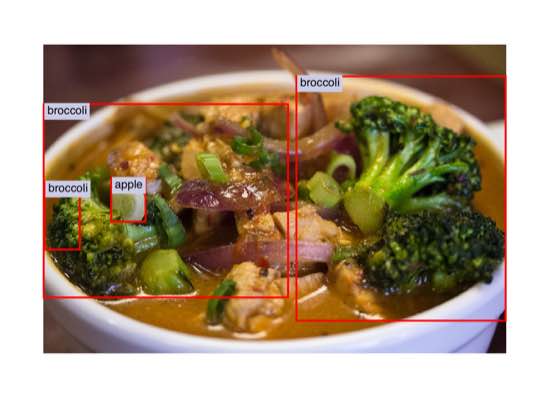} &
\includegraphics[trim = 14mm 20mm 14mm 10mm , clip=true,width=0.25\textwidth,height=\textheight,keepaspectratio]{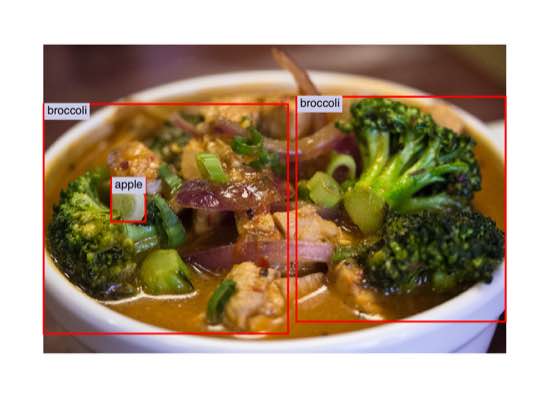} & 
\includegraphics[trim = 14mm 20mm 14mm 10mm , clip=true,width=0.25\textwidth,height=\textheight,keepaspectratio]{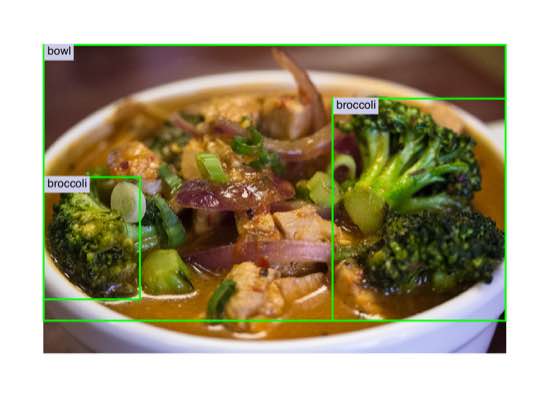} \\

\includegraphics[trim = 14mm 20mm 14mm 10mm , clip=true,width=0.25\textwidth,height=\textheight,keepaspectratio]{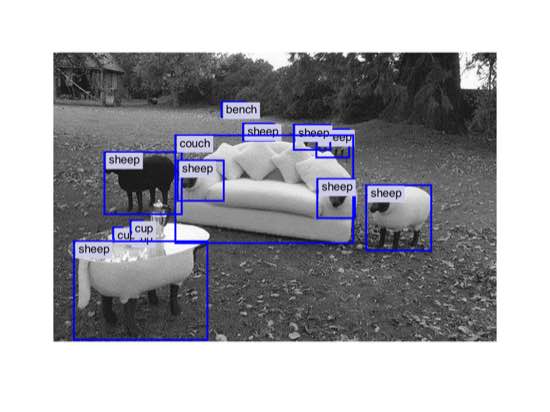} & 
\includegraphics[trim = 14mm 20mm 14mm 10mm , clip=true,width=0.25\textwidth,height=\textheight,keepaspectratio]{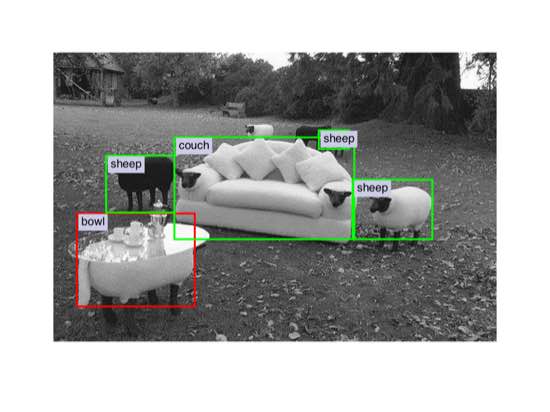} &
\includegraphics[trim = 14mm 20mm 14mm 10mm , clip=true,width=0.25\textwidth,height=\textheight,keepaspectratio]{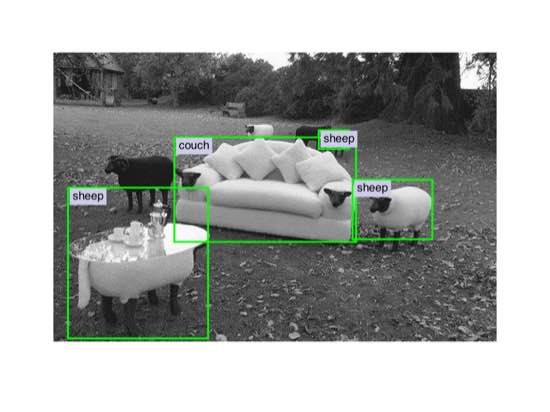} & 
\includegraphics[trim = 14mm 20mm 14mm 10mm , clip=true,width=0.25\textwidth,height=\textheight,keepaspectratio]{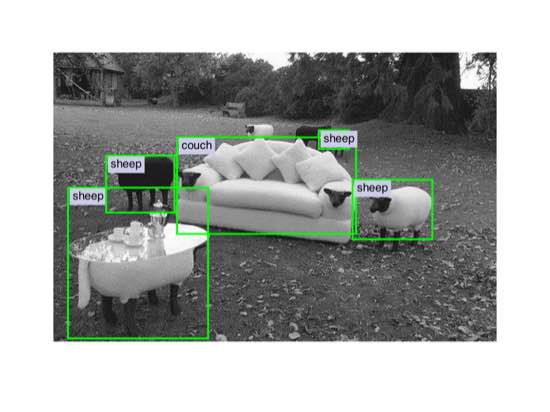} \\

\includegraphics[trim = 14mm 20mm 14mm 5mm , clip=true,width=0.25\textwidth,height=\textheight,keepaspectratio]{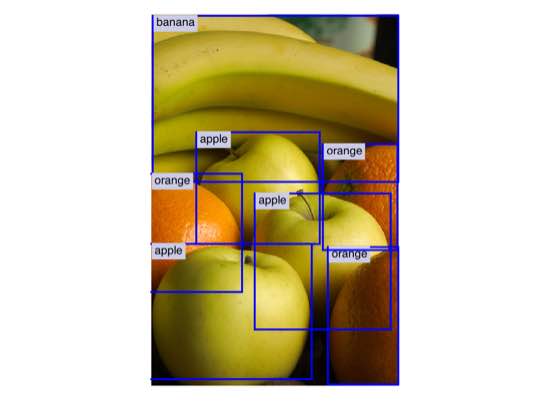} & 
\includegraphics[trim = 14mm 20mm 14mm 5mm , clip=true,width=0.25\textwidth,height=\textheight,keepaspectratio]{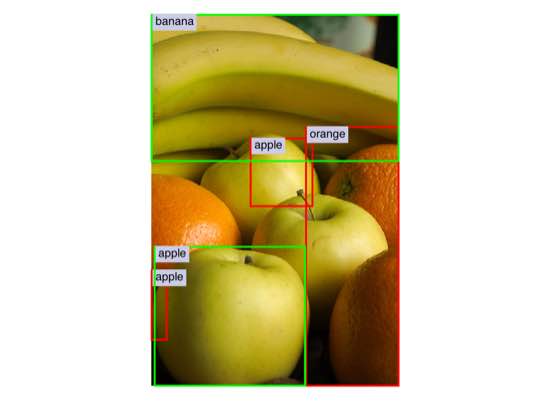} &
\includegraphics[trim = 14mm 20mm 14mm 5mm , clip=true,width=0.25\textwidth,height=\textheight,keepaspectratio]{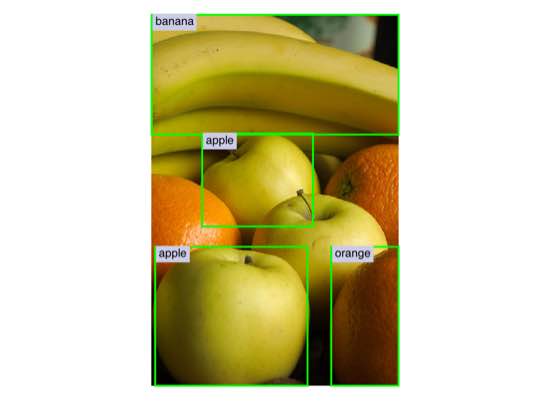} & 
\includegraphics[trim = 14mm 20mm 14mm 5mm , clip=true,width=0.25\textwidth,height=\textheight,keepaspectratio]{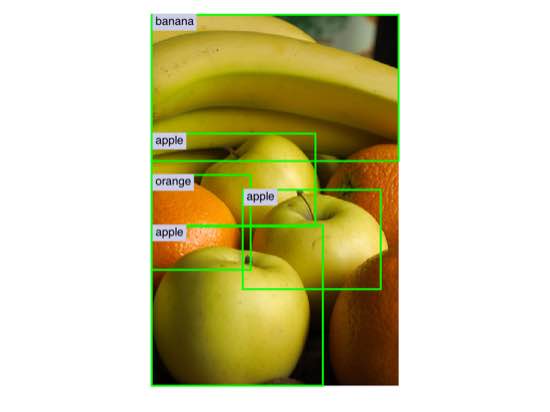} \\

\includegraphics[trim = 14mm 20mm 14mm 10mm , clip=true,width=0.25\textwidth,height=\textheight,keepaspectratio]{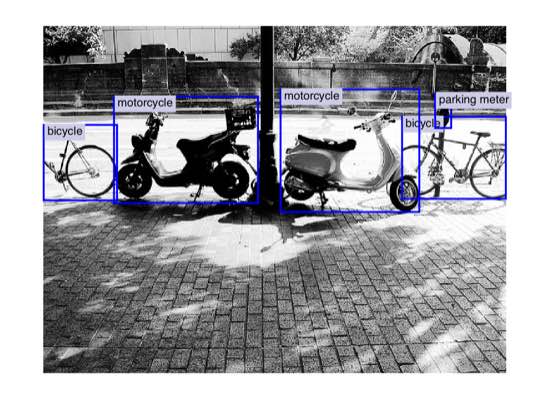} & 
\includegraphics[trim = 14mm 20mm 14mm 10mm , clip=true,width=0.25\textwidth,height=\textheight,keepaspectratio]{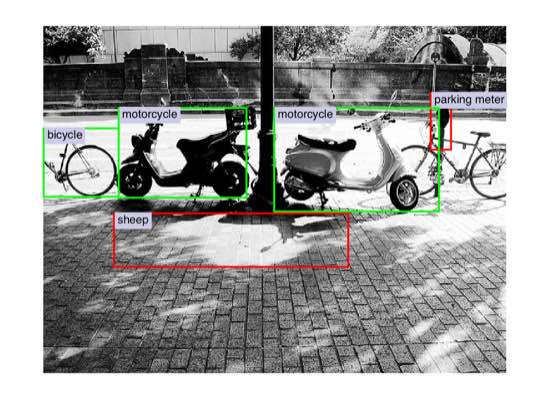} &
\includegraphics[trim = 14mm 20mm 14mm 10mm , clip=true,width=0.25\textwidth,height=\textheight,keepaspectratio]{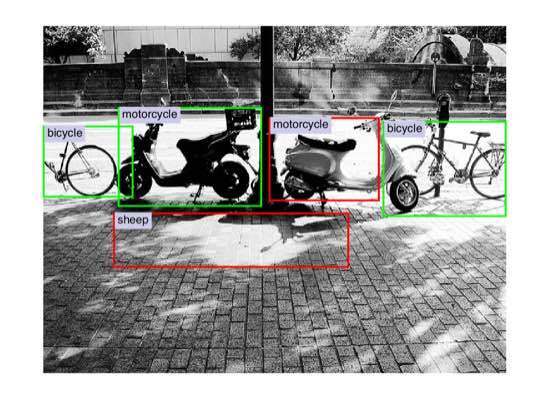} & 
\includegraphics[trim = 14mm 20mm 14mm 10mm , clip=true,width=0.25\textwidth,height=\textheight,keepaspectratio]{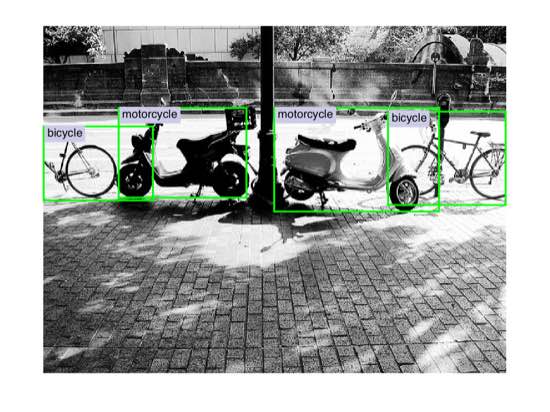} \\

\includegraphics[trim = 14mm 20mm 14mm 10mm , clip=true,width=0.25\textwidth,height=\textheight,keepaspectratio]{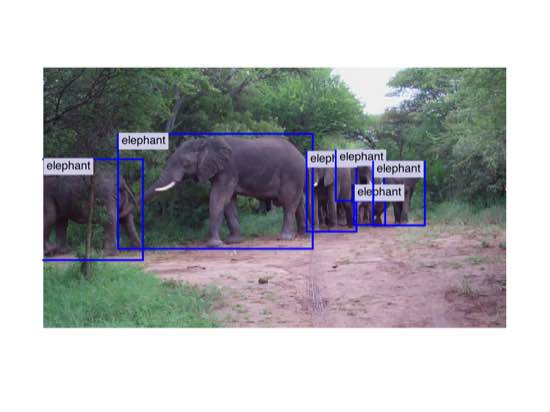} & 
\includegraphics[trim = 14mm 20mm 14mm 10mm , clip=true,width=0.25\textwidth,height=\textheight,keepaspectratio]{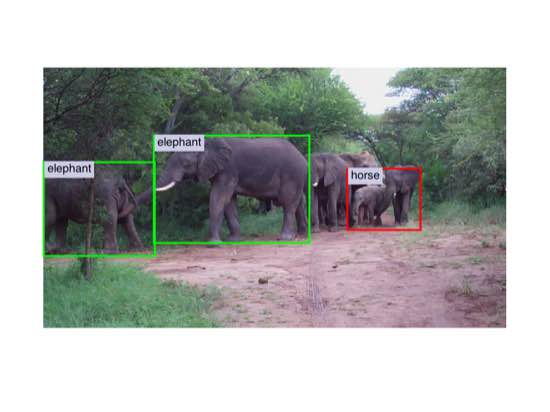} &
\includegraphics[trim = 14mm 20mm 14mm 10mm , clip=true,width=0.25\textwidth,height=\textheight,keepaspectratio]{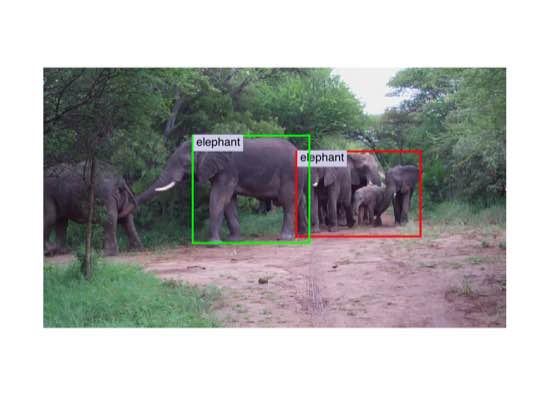} & 
\includegraphics[trim = 14mm 20mm 14mm 10mm , clip=true,width=0.25\textwidth,height=\textheight,keepaspectratio]{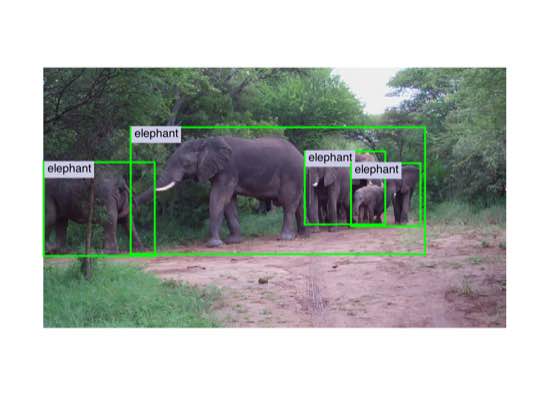} \\

\includegraphics[trim = 14mm 20mm 14mm 10mm , clip=true,width=0.25\textwidth,height=\textheight,keepaspectratio]{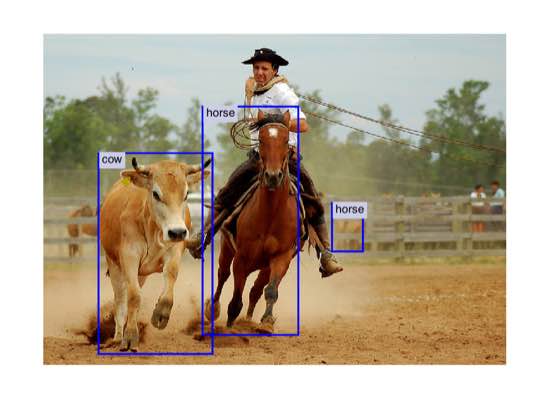} & 
\includegraphics[trim = 14mm 20mm 14mm 10mm , clip=true,width=0.25\textwidth,height=\textheight,keepaspectratio]{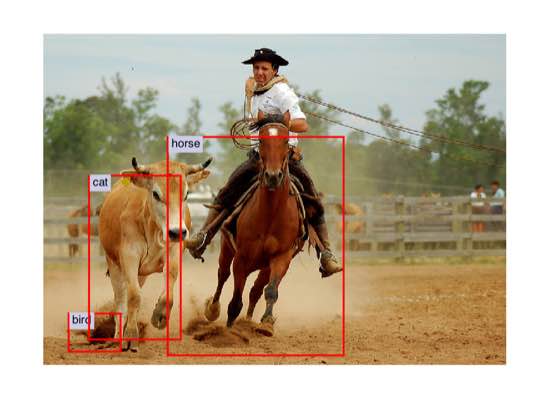} &
\includegraphics[trim = 14mm 20mm 14mm 10mm , clip=true,width=0.25\textwidth,height=\textheight,keepaspectratio]{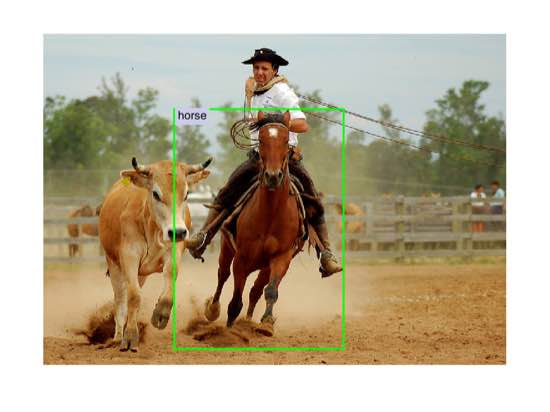} & 
\includegraphics[trim = 14mm 20mm 14mm 10mm , clip=true,width=0.25\textwidth,height=\textheight,keepaspectratio]{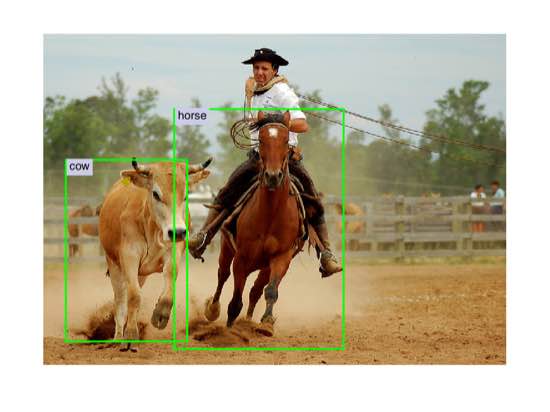} \\

\end{tabular}
\end{center}
\caption{Examples where MAPC outperforms WC+AC-NMS and SAPC + AC-NMS on Microsoft COCO. Ground truth: blue, true positives: green, false positives: red.}
\label{table:COCOBetterBest}
\end{figure*}

Exemplary, Figures \ref{table:COCOBetterBest}, \ref{table:ImageNetBetterBest} and \ref{table:LargeBetterBest} depict images, where MAPC both localised and classified objects better, resulting in an overall significantly better detection result. For both, a large number of object instances per image as in COCO, as well as a large number of detection classes as in our ImageNet 1,825 categories dataset or our qualitative dataset with 7,404 categories, MAPC performs better than the baselines. Hence, based on LSDA detections, our method
enables a very fine-grained and precise detection, be it between different types of bags, flowers or trees.

The following sections analyse more images qualitatively and are organised as follows. Section \ref{betterloc} discusses images, where MAPC significantly better localised objects. Section \ref{betterclass} analyses images, where MAPC significantly better classified objects. Finally, we show some failure cases of MAPC in Section \ref{failuremapc}.

\subsection{Improved localisation}
\label{betterloc}
\begin{figure*}
\begin{center}
\begin{tabular}{c@{}c@{}c@{}c}
Ground Truth & WC+AC-NMS & SAPC + AC-NMS & MAPC (ours) \\
\hline\hline

\includegraphics[trim = 14mm 20mm 14mm 20mm , clip=true,width=0.25\textwidth,height=\textheight,keepaspectratio]{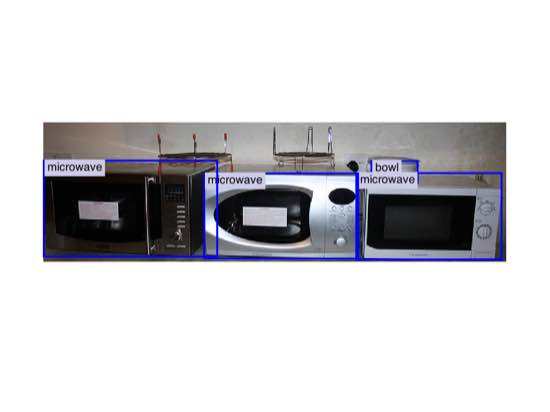} & 
\includegraphics[trim = 14mm 20mm 14mm 20mm , clip=true,width=0.25\textwidth,height=\textheight,keepaspectratio]{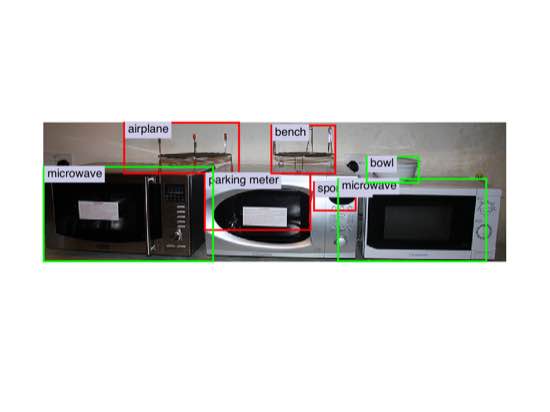} & 
\includegraphics[trim = 14mm 20mm 14mm 20mm , clip=true,width=0.25\textwidth,height=\textheight,keepaspectratio]{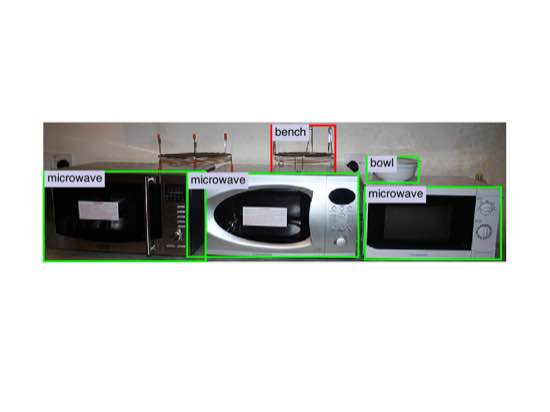} & 
\includegraphics[trim = 14mm 20mm 14mm 20mm , clip=true,width=0.25\textwidth,height=\textheight,keepaspectratio]{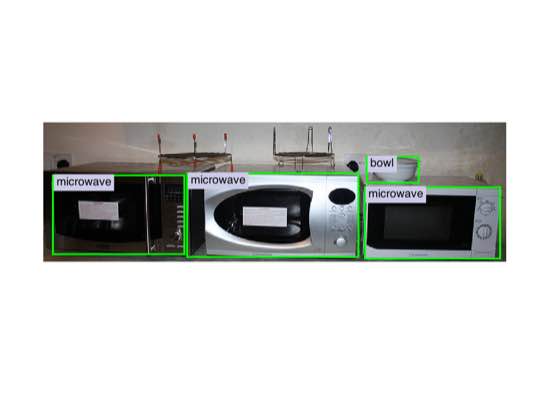} \\

\includegraphics[trim = 14mm 20mm 14mm 20mm , clip=true,width=0.25\textwidth,height=\textheight,keepaspectratio]{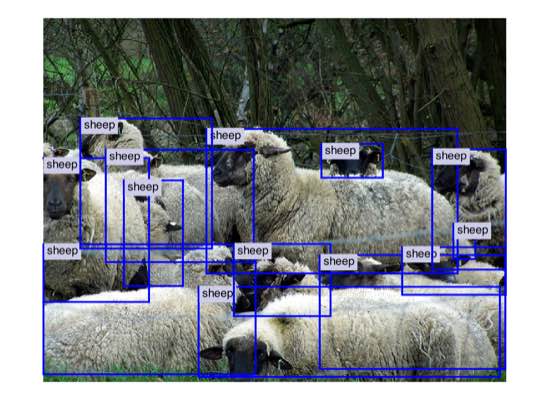} & 
\includegraphics[trim = 14mm 20mm 14mm 20mm , clip=true,width=0.25\textwidth,height=\textheight,keepaspectratio]{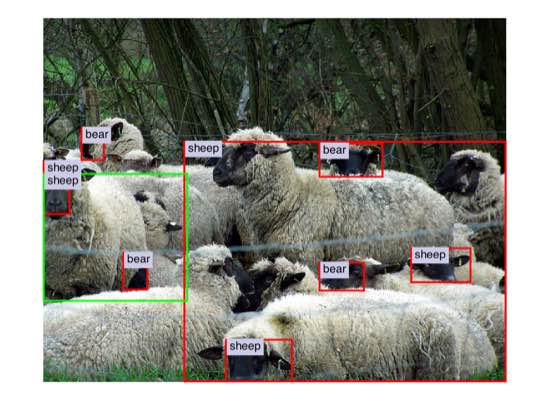} & 
\includegraphics[trim = 14mm 20mm 14mm 20mm , clip=true,width=0.25\textwidth,height=\textheight,keepaspectratio]{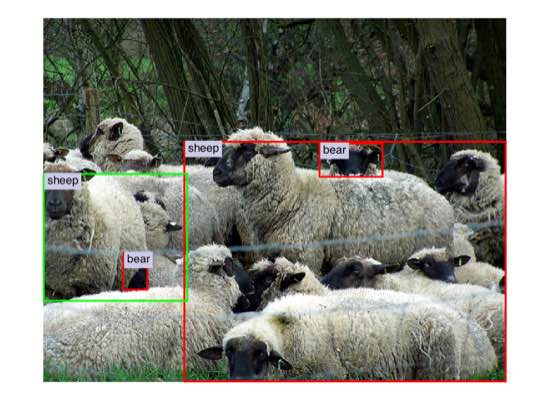} & 
\includegraphics[trim = 14mm 20mm 14mm 20mm , clip=true,width=0.25\textwidth,height=\textheight,keepaspectratio]{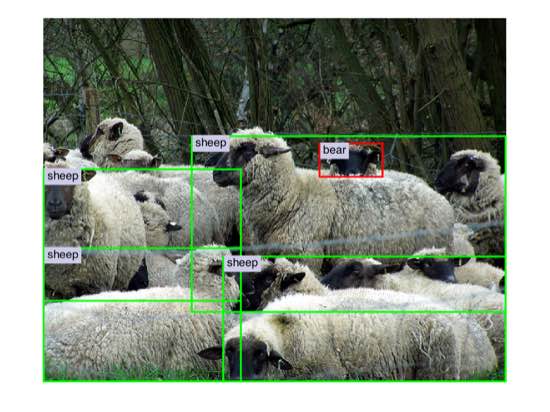} \\

\includegraphics[trim = 14mm 15mm 14mm 5mm , clip=true,width=0.25\textwidth,height=\textheight,keepaspectratio]{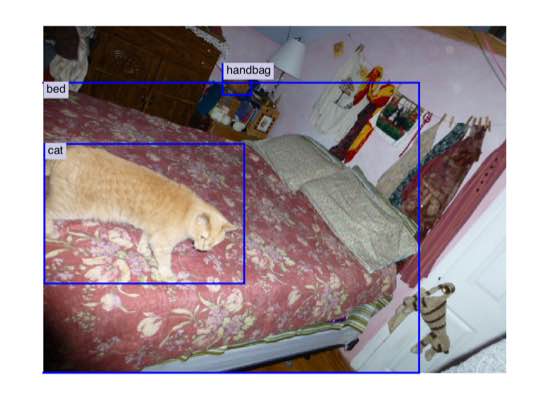} & 
\includegraphics[trim = 14mm 10mm 14mm 5mm , clip=true,width=0.25\textwidth,height=\textheight,keepaspectratio]{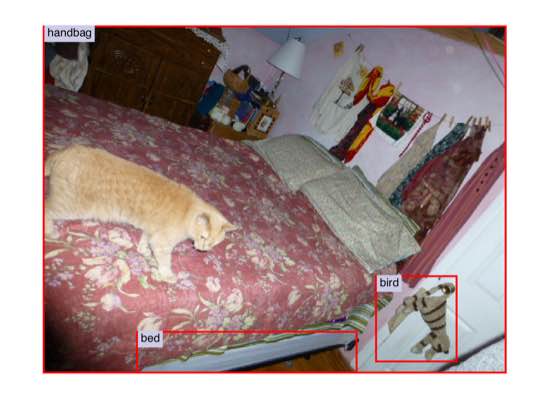} &
\includegraphics[trim = 14mm 10mm 14mm 5mm , clip=true,width=0.25\textwidth,height=\textheight,keepaspectratio]{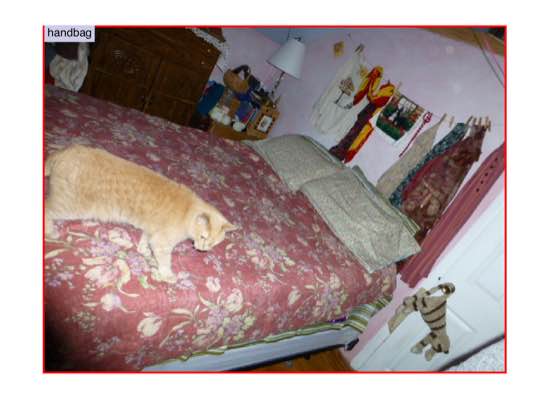} & 
\includegraphics[trim = 14mm 10mm 14mm 5mm , clip=true,width=0.25\textwidth,height=\textheight,keepaspectratio]{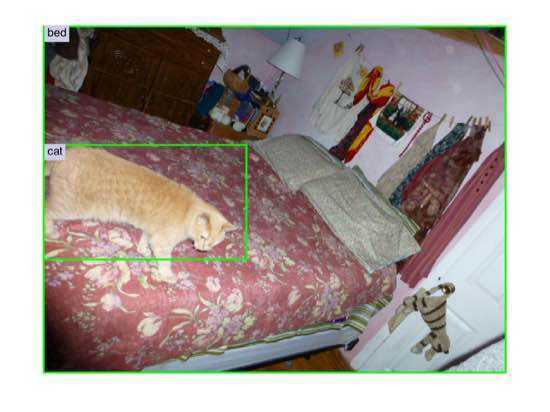} \\

\includegraphics[trim = 14mm 20mm 14mm 10mm , clip=true,width=0.25\textwidth,height=\textheight,keepaspectratio]{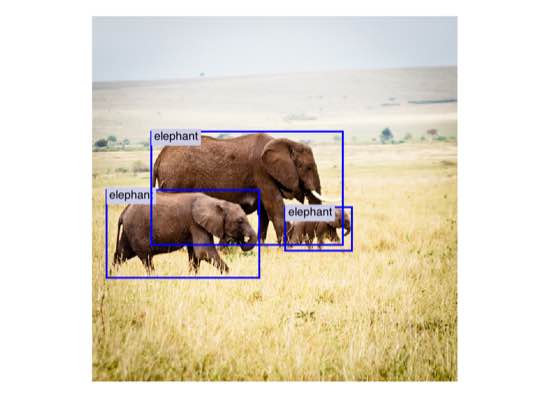} & 
\includegraphics[trim = 14mm 20mm 14mm 10mm , clip=true,width=0.25\textwidth,height=\textheight,keepaspectratio]{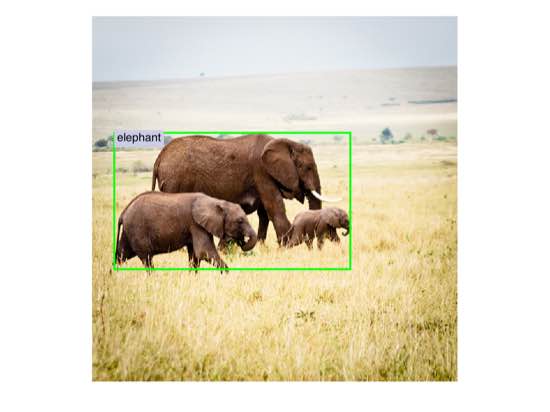} &
\includegraphics[trim = 14mm 20mm 14mm 10mm , clip=true,width=0.25\textwidth,height=\textheight,keepaspectratio]{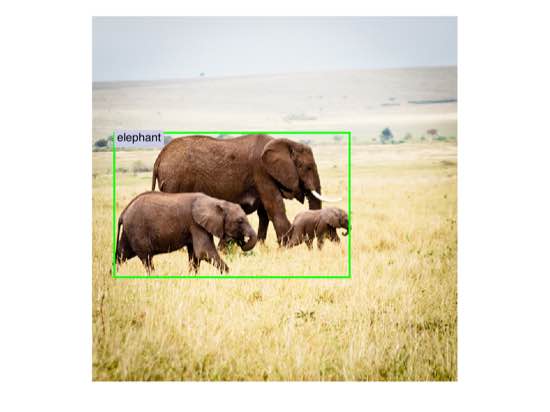} & 
\includegraphics[trim = 14mm 20mm 14mm 10mm , clip=true,width=0.25\textwidth,height=\textheight,keepaspectratio]{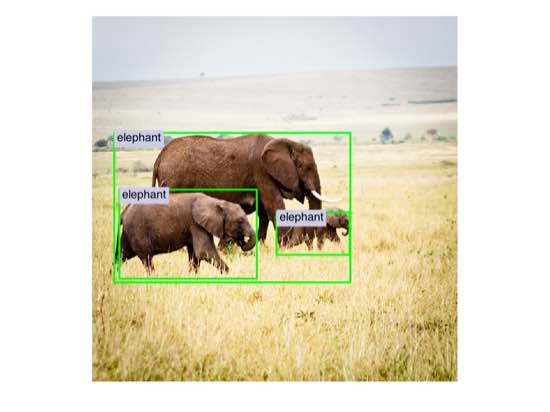} \\

\includegraphics[trim = 14mm 10mm 14mm 5mm , clip=true,width=0.25\textwidth,height=\textheight,keepaspectratio]{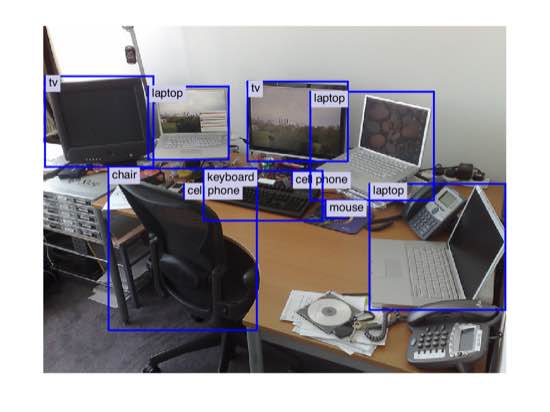} & 
\includegraphics[trim = 14mm 10mm 14mm 5mm , clip=true,width=0.25\textwidth,height=\textheight,keepaspectratio]{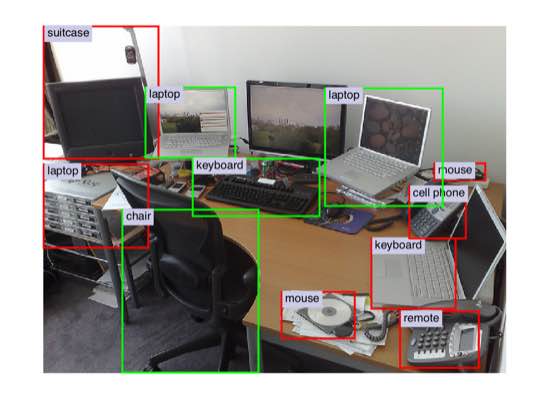} &
\includegraphics[trim = 14mm 10mm 14mm 5mm , clip=true,width=0.25\textwidth,height=\textheight,keepaspectratio]{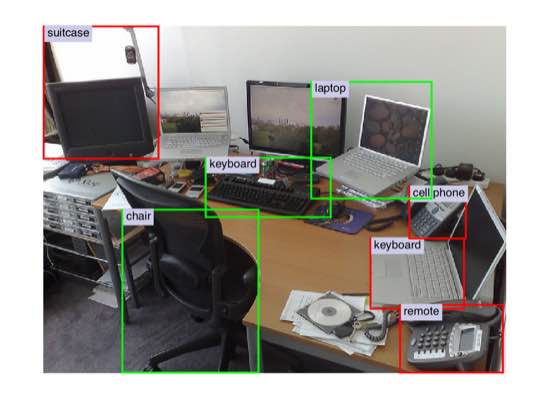} & 
\includegraphics[trim = 14mm 10mm 14mm 5mm , clip=true,width=0.25\textwidth,height=\textheight,keepaspectratio]{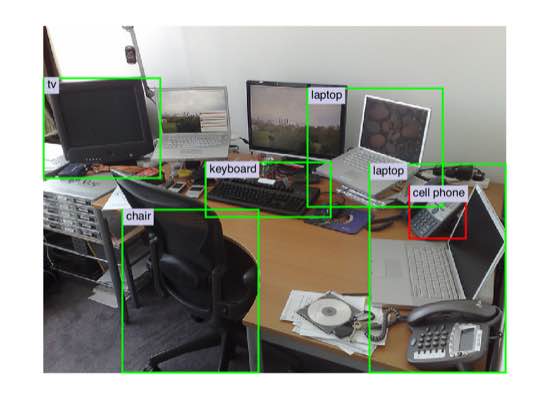} \\

\includegraphics[trim = 14mm 10mm 14mm 10mm , clip=true,width=0.25\textwidth,height=\textheight,keepaspectratio]{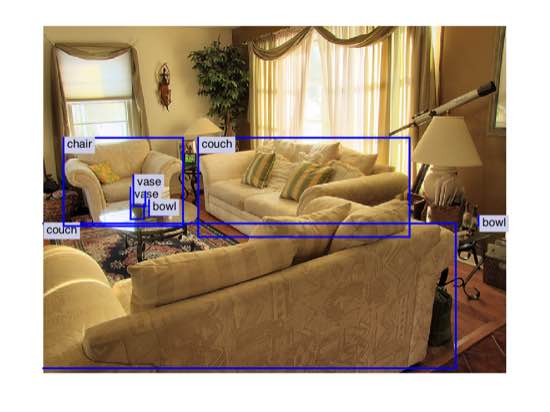} & 
\includegraphics[trim = 14mm 10mm 14mm 10mm , clip=true,width=0.25\textwidth,height=\textheight,keepaspectratio]{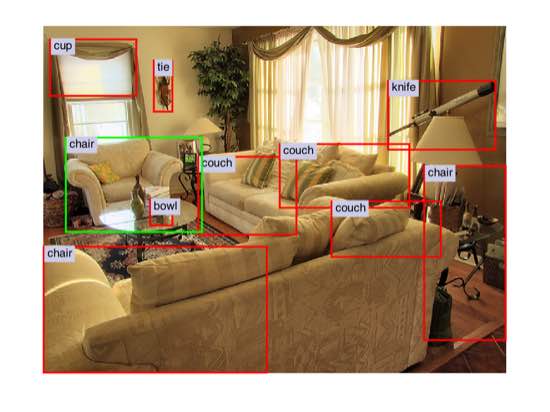} &
\includegraphics[trim = 14mm 10mm 14mm 10mm , clip=true,width=0.25\textwidth,height=\textheight,keepaspectratio]{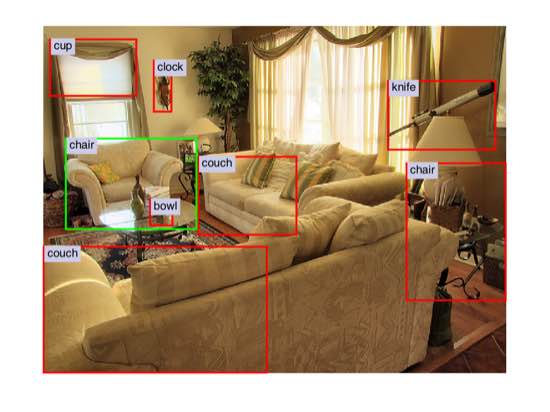} & 
\includegraphics[trim = 14mm 10mm 14mm 10mm , clip=true,width=0.25\textwidth,height=\textheight,keepaspectratio]{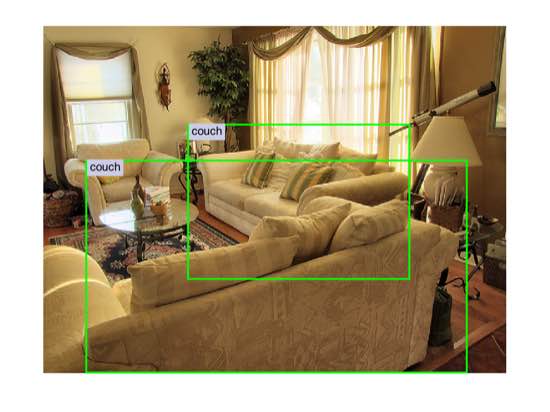} \\

\end{tabular}
\end{center}
\caption{Examples where MAPC better localises objects than WC+AC-NMS and SAPC + AC-NMS on Microsoft COCO. Ground truth: blue, true positives: green, false positives: red.}
\label{table:COCOBetterLoc}
\end{figure*}

In this section, we show and generally discuss images, where MAPC significantly better localises objects than the baselines. These images can be seen in Figures \ref{table:COCOBetterLoc}, \ref{table:ImageNetBetterLoc} and \ref{table:LargeBetterLoc}. 
As can be seen, WC+AC-NMS often selects bigger detection proposals that group multiple objects together. Thereby, NMS simply takes the maximum scoring detection proposal and locally suppresses all other detections which overlap with this detection. Hence, if large proposals score high in a detection setup, those detection proposals suppress all other detection proposals, which makes a more detailed detection on the object level impossible. But also in other scenarios, where such large detections do not appear, WC+AC-NMS performs worse in localising objects as can be seen in the images. 
In contrast, SAPC \cite{eth_biwi_01126} optimizes over the whole set of detection proposals, which improves the localisation compared to WC+AC-NMS. However, SAPC does not regularise across classes. Thus, MAPC, which groups spatially and semantically similar detections together, does much better in localising objects. MAPC benefits from the fact that usually multiple similarly sized and classified detection proposals lie on the same object due to the visual similarity of such proposals. Hence, local semantically and spatially similar clusters of detection proposals are formed over objects all over the image, which are grouped together by MAPC. This results in a precise object localisation, as can be seen in the images.

\subsection{Improved Classification}
\label{betterclass}
\begin{figure*}
\begin{center}
\begin{tabular}{c@{}c@{}c@{}c}
Ground Truth & WC+AC-NMS & SAPC + AC-NMS & MAPC (ours) \\
\hline\hline

\includegraphics[trim = 14mm 20mm 14mm 10mm , clip=true,width=0.25\textwidth,height=\textheight,keepaspectratio]{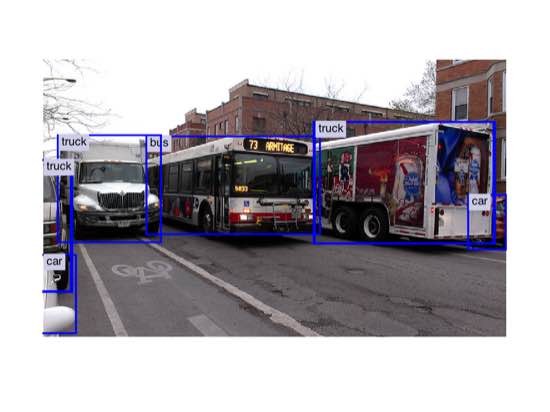} & 
\includegraphics[trim = 14mm 20mm 14mm 10mm , clip=true,width=0.25\textwidth,height=\textheight,keepaspectratio]{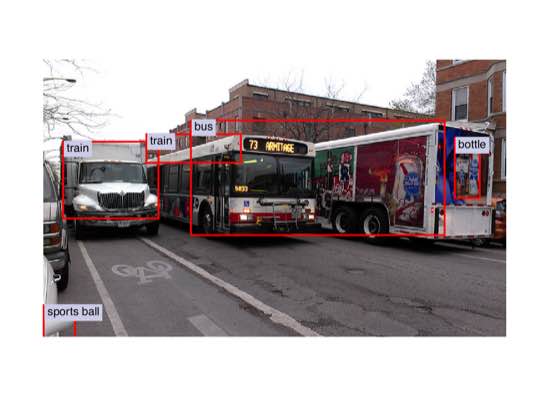} &
\includegraphics[trim = 14mm 20mm 14mm 10mm , clip=true,width=0.25\textwidth,height=\textheight,keepaspectratio]{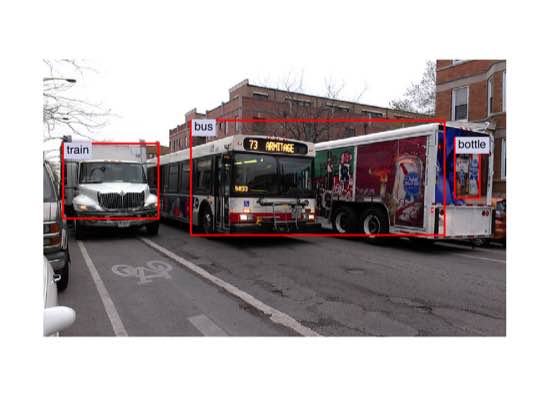} & 
\includegraphics[trim = 14mm 20mm 14mm 10mm , clip=true,width=0.25\textwidth,height=\textheight,keepaspectratio]{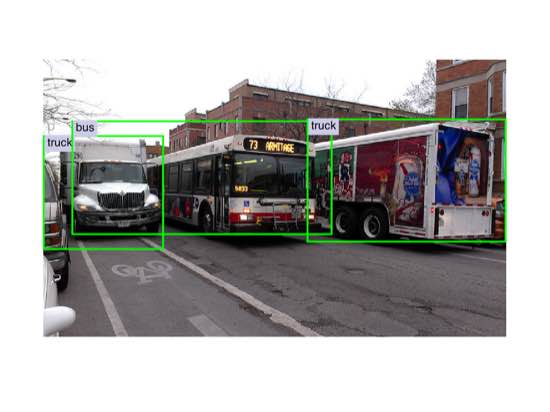} \\

\includegraphics[trim = 14mm 10mm 14mm 0mm , clip=true,width=0.25\textwidth,height=\textheight,keepaspectratio]{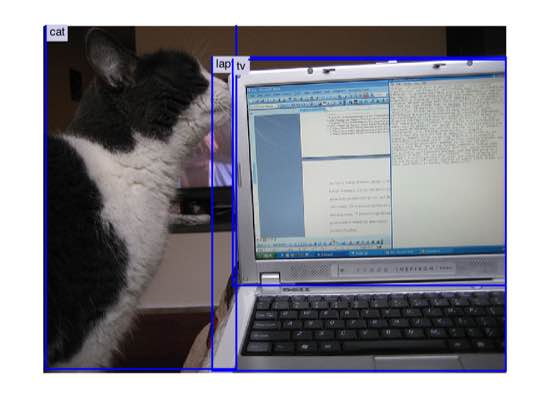} & 
\includegraphics[trim = 14mm 10mm 14mm 0mm , clip=true,width=0.25\textwidth,height=\textheight,keepaspectratio]{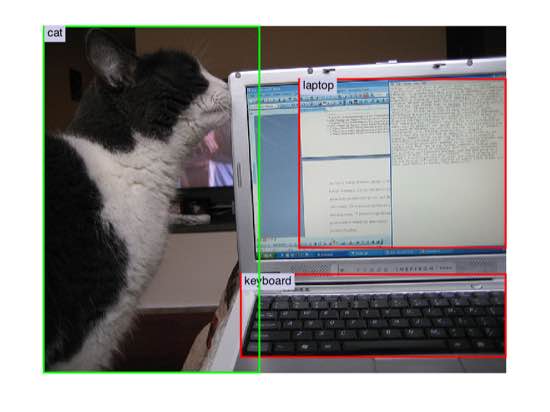} &
\includegraphics[trim = 14mm 10mm 14mm 0mm , clip=true,width=0.25\textwidth,height=\textheight,keepaspectratio]{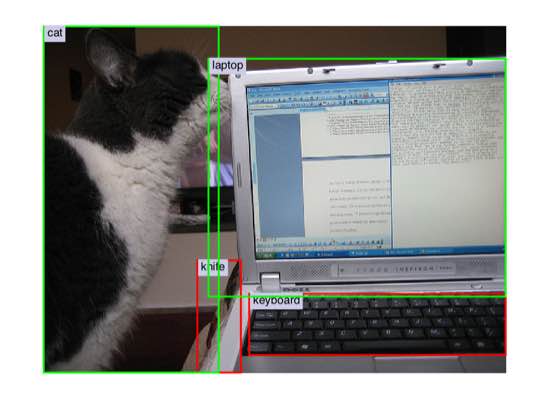} & 
\includegraphics[trim = 14mm 10mm 14mm 0mm , clip=true,width=0.25\textwidth,height=\textheight,keepaspectratio]{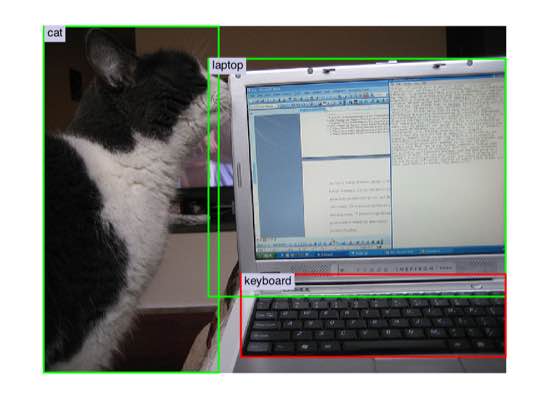} \\

\includegraphics[trim = 14mm 20mm 14mm 10mm , clip=true,width=0.25\textwidth,height=\textheight,keepaspectratio]{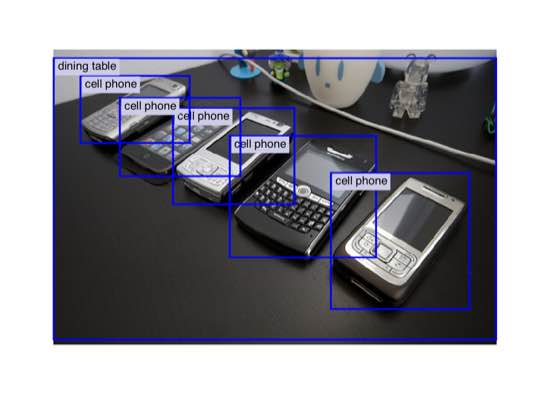} & 
\includegraphics[trim = 14mm 20mm 14mm 10mm , clip=true,width=0.25\textwidth,height=\textheight,keepaspectratio]{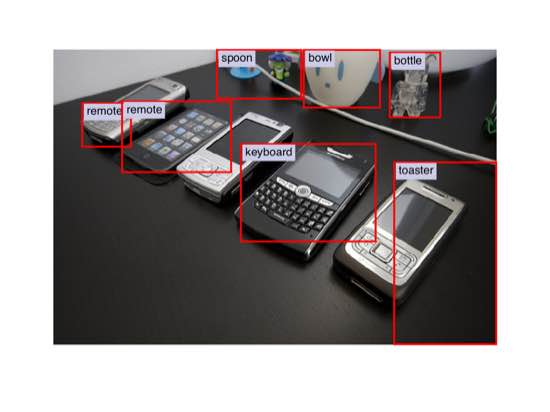} &
\includegraphics[trim = 14mm 20mm 14mm 10mm , clip=true,width=0.25\textwidth,height=\textheight,keepaspectratio]{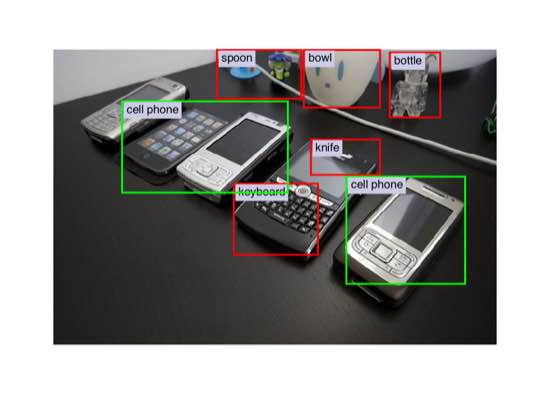} & 
\includegraphics[trim = 14mm 20mm 14mm 10mm , clip=true,width=0.25\textwidth,height=\textheight,keepaspectratio]{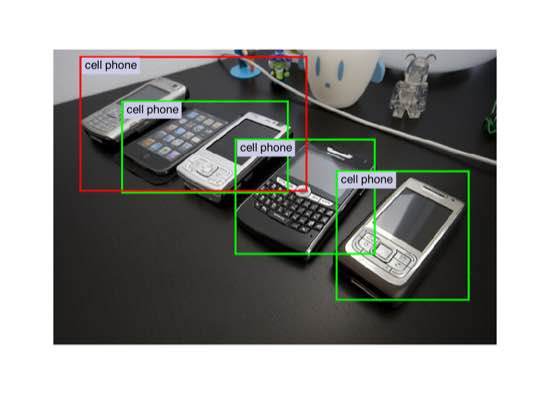} \\

\includegraphics[trim = 0mm 0mm 0mm 0mm , clip=true,width=0.25\textwidth,height=\textheight,keepaspectratio]{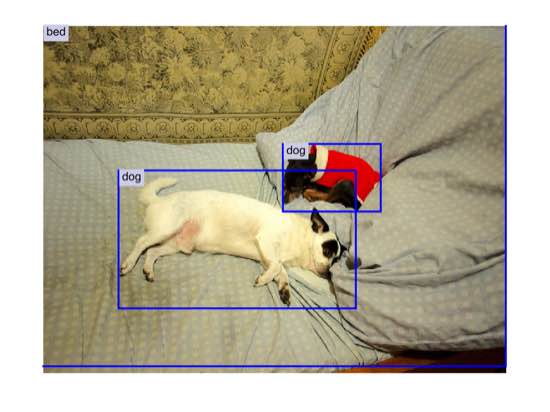} & 
\includegraphics[trim = 0mm 0mm 0mm 0mm , clip=true,width=0.25\textwidth,height=\textheight,keepaspectratio]{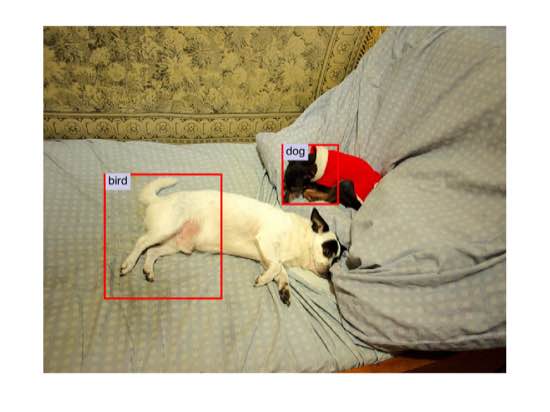} &
\includegraphics[trim = 0mm 0mm 0mm 0mm , clip=true,width=0.25\textwidth,height=\textheight,keepaspectratio]{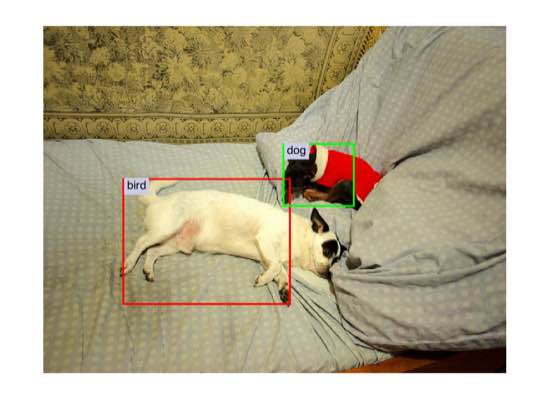} & 
\includegraphics[trim = 0mm 0mm 0mm 0mm , clip=true,width=0.25\textwidth,height=\textheight,keepaspectratio]{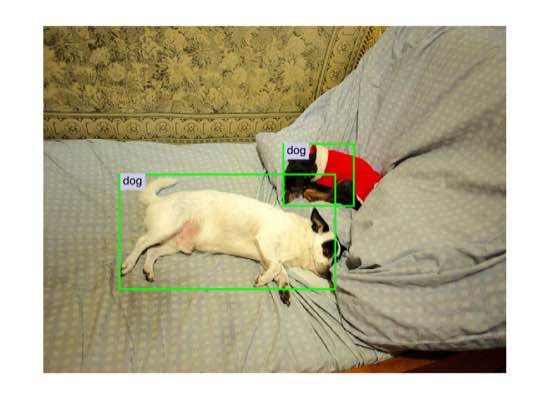} \\

\includegraphics[trim = 14mm 20mm 14mm 10mm , clip=true,width=0.25\textwidth,height=\textheight,keepaspectratio]{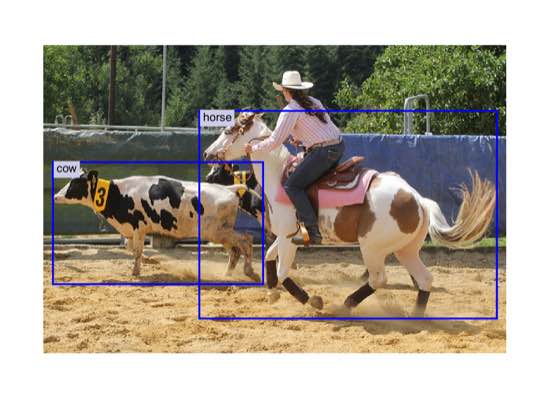} & 
\includegraphics[trim = 14mm 20mm 14mm 10mm , clip=true,width=0.25\textwidth,height=\textheight,keepaspectratio]{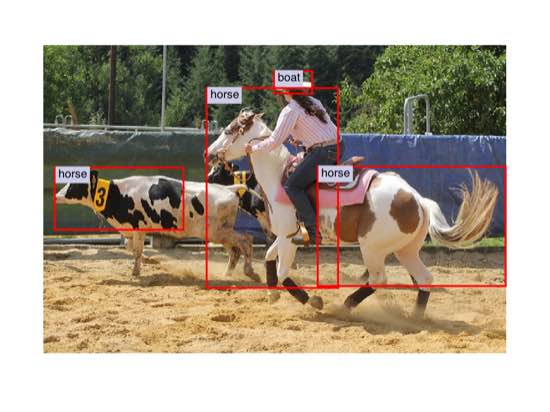} &
\includegraphics[trim = 14mm 20mm 14mm 10mm , clip=true,width=0.25\textwidth,height=\textheight,keepaspectratio]{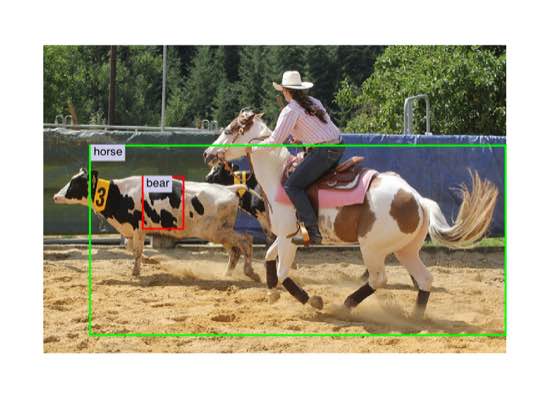} & 
\includegraphics[trim = 14mm 20mm 14mm 10mm , clip=true,width=0.25\textwidth,height=\textheight,keepaspectratio]{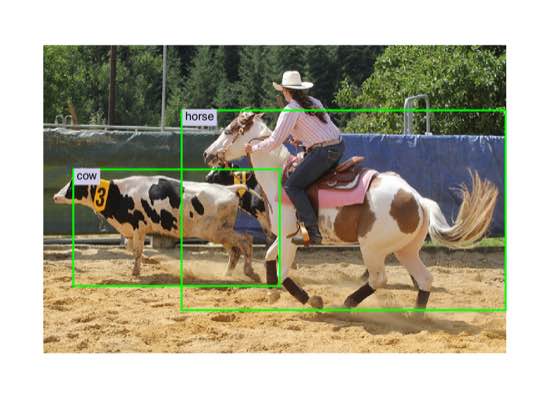} \\

\includegraphics[trim = 14mm 5mm 14mm 5mm , clip=true,width=0.25\textwidth,height=\textheight,keepaspectratio]{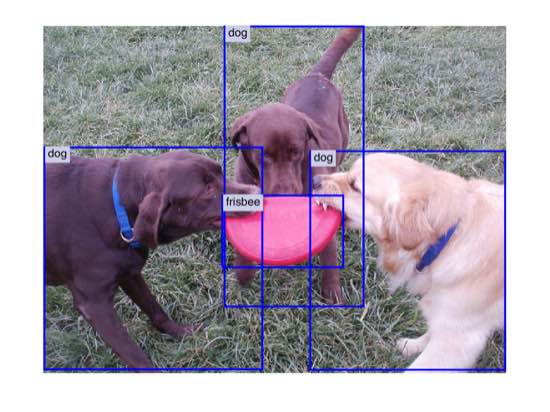} & 
\includegraphics[trim = 14mm 5mm 14mm 5mm , clip=true,width=0.25\textwidth,height=\textheight,keepaspectratio]{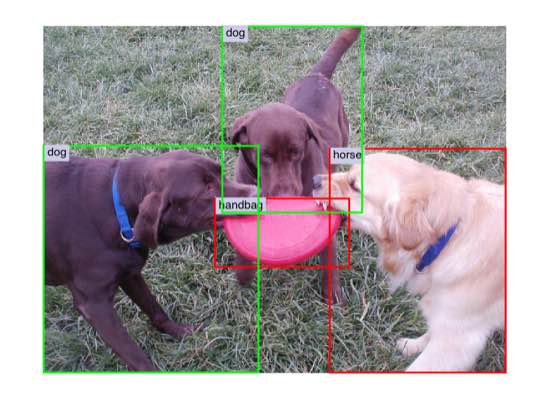} &
\includegraphics[trim = 14mm 5mm 14mm 5mm , clip=true,width=0.25\textwidth,height=\textheight,keepaspectratio]{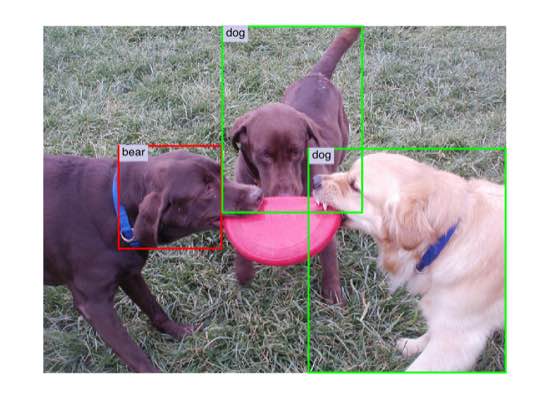} & 
\includegraphics[trim = 14mm 5mm 14mm 5mm , clip=true,width=0.25\textwidth,height=\textheight,keepaspectratio]{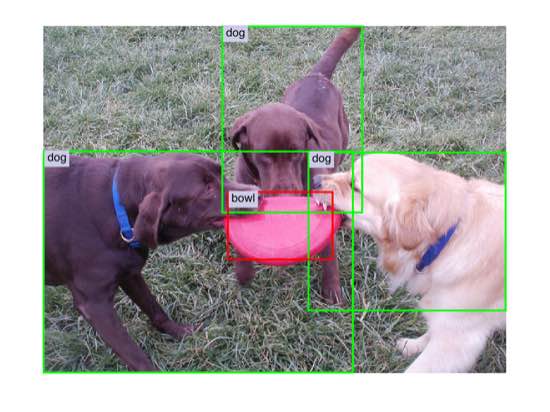} \\

\end{tabular}
\end{center}
\caption{Examples where MAPC better classifies objects than WC+AC-NMS and SAPC + AC-NMS on Microsoft COCO. Ground truth: blue, true positives: green, false positives: red.}
\label{table:COCOBetterClass}
\end{figure*}

\begin{figure*}
\begin{center}
\begin{tabular}{c@{}c@{}c@{}c}
Ground Truth & WC+AC-NMS & SAPC + AC-NMS & MAPC (ours) \\
\hline\hline
\includegraphics[trim = 14mm 5mm 14mm 5mm , clip=true,width=0.25\textwidth,height=\textheight,keepaspectratio]{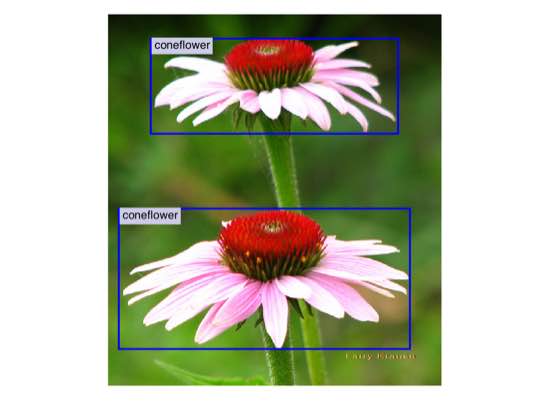} & 
\includegraphics[trim = 14mm 5mm 14mm 5mm , clip=true,width=0.25\textwidth,height=\textheight,keepaspectratio]{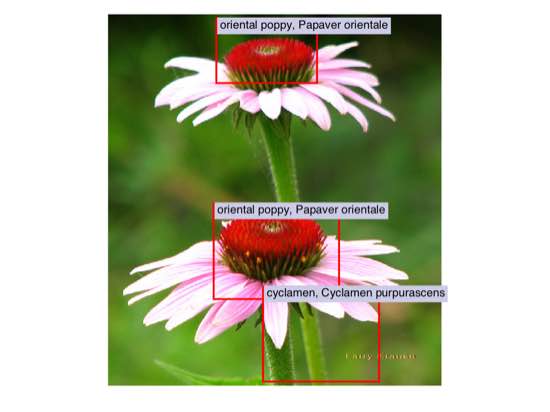} &
\includegraphics[trim = 14mm 5mm 14mm 5mm , clip=true,width=0.25\textwidth,height=\textheight,keepaspectratio]{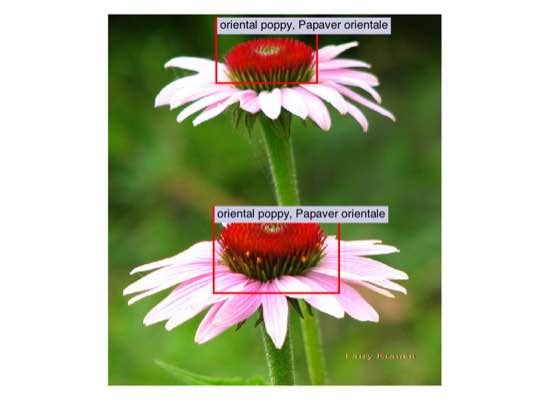} & 
\includegraphics[trim = 14mm 5mm 14mm 5mm , clip=true,width=0.25\textwidth,height=\textheight,keepaspectratio]{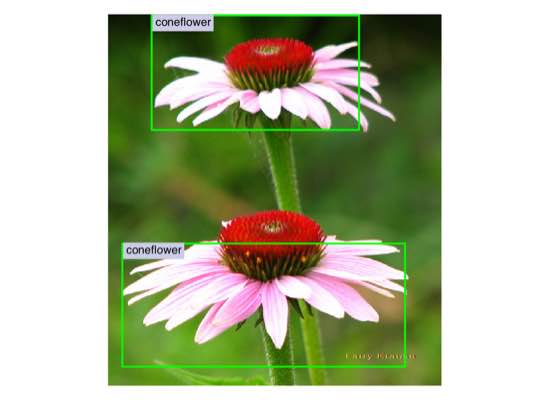} \\

\includegraphics[trim = 5mm 0mm 5mm 15mm , clip=true,width=0.25\textwidth,height=\textheight,keepaspectratio]{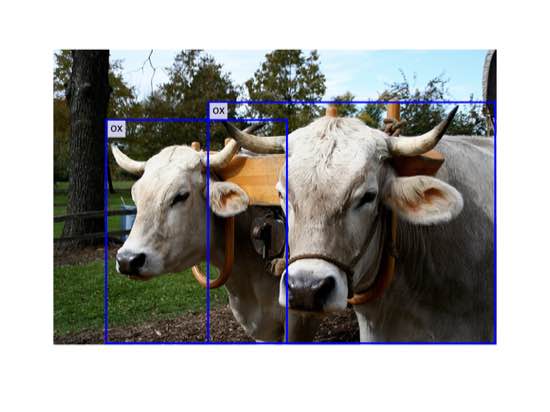} & 
\includegraphics[trim = 5mm 0mm 5mm 15mm , clip=true,width=0.25\textwidth,height=\textheight,keepaspectratio]{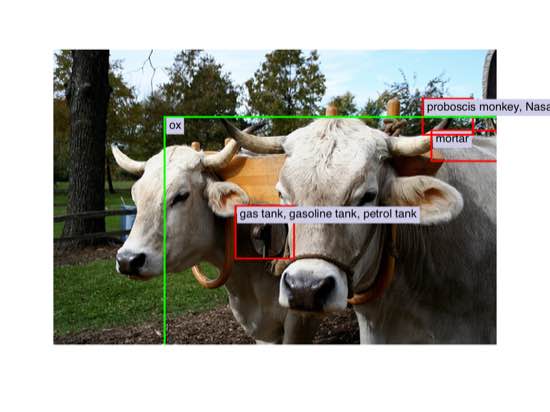} &
\includegraphics[trim = 5mm 0mm 5mm 15mm , clip=true,width=0.25\textwidth,height=\textheight,keepaspectratio]{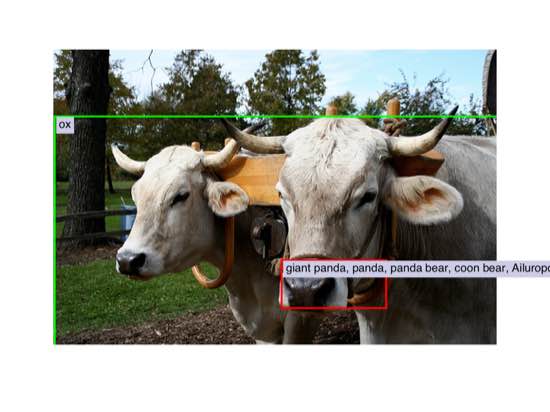} & 
\includegraphics[trim = 5mm 0mm 5mm 15mm , clip=true,width=0.25\textwidth,height=\textheight,keepaspectratio]{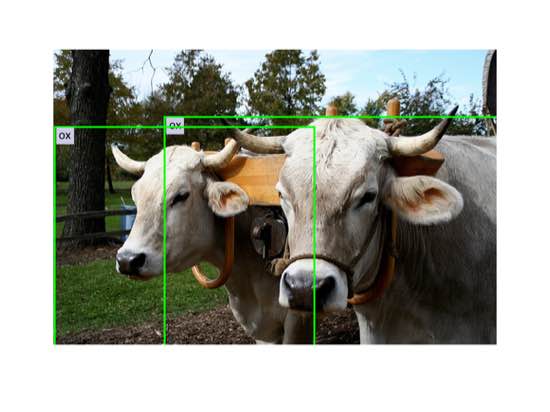} \\

\includegraphics[trim = 14mm 5mm 14mm 5mm , clip=true,width=0.25\textwidth,height=\textheight,keepaspectratio]{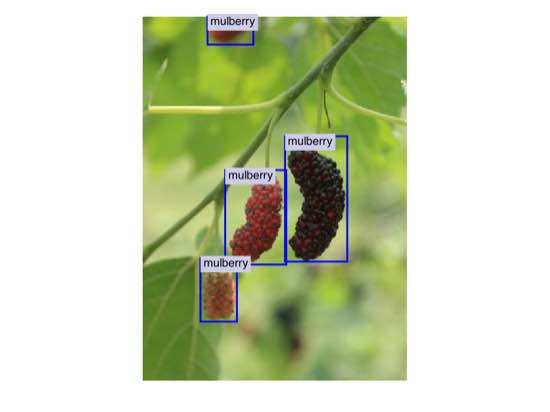} & 
\includegraphics[trim = 14mm 5mm 14mm 5mm , clip=true,width=0.25\textwidth,height=\textheight,keepaspectratio]{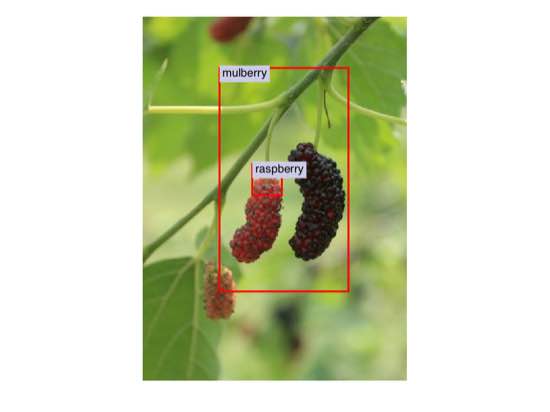} &
\includegraphics[trim = 14mm 5mm 14mm 5mm , clip=true,width=0.25\textwidth,height=\textheight,keepaspectratio]{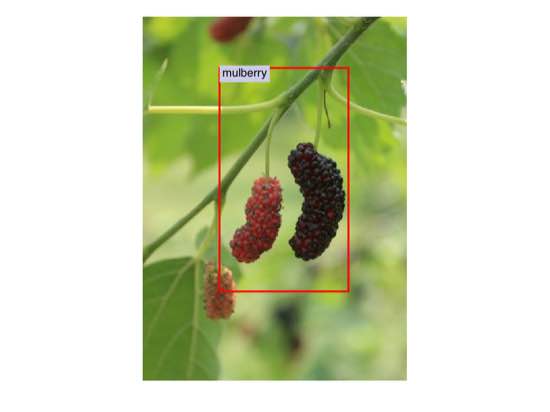} & 
\includegraphics[trim = 14mm 5mm 14mm 5mm , clip=true,width=0.25\textwidth,height=\textheight,keepaspectratio]{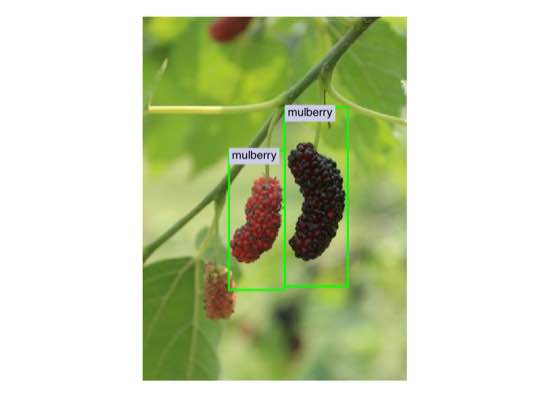} \\

\includegraphics[trim = 14mm 5mm 14mm 5mm , clip=true,width=0.25\textwidth,height=\textheight,keepaspectratio]{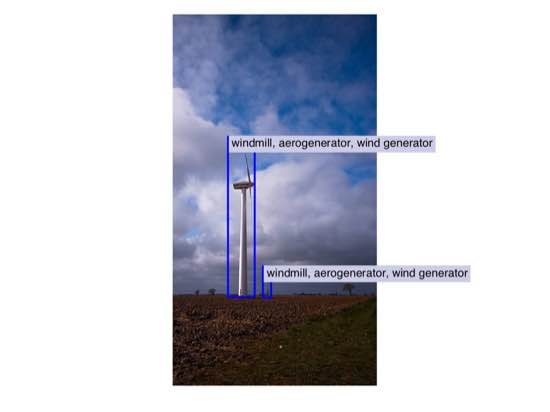} & 
\includegraphics[trim = 14mm 5mm 14mm 5mm , clip=true,width=0.25\textwidth,height=\textheight,keepaspectratio]{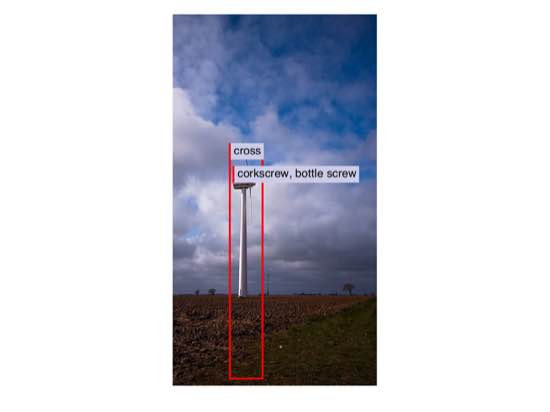} &
\includegraphics[trim = 14mm 5mm 14mm 5mm , clip=true,width=0.25\textwidth,height=\textheight,keepaspectratio]{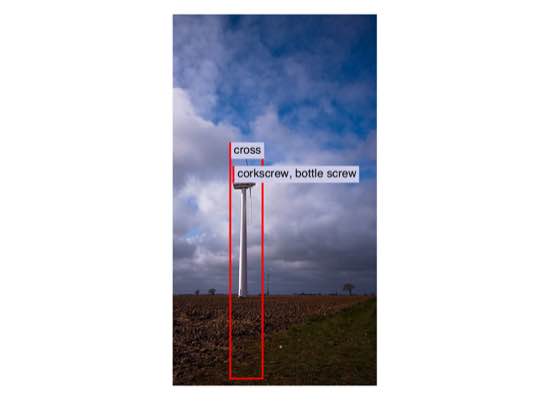} & 
\includegraphics[trim = 14mm 5mm 14mm 5mm , clip=true,width=0.25\textwidth,height=\textheight,keepaspectratio]{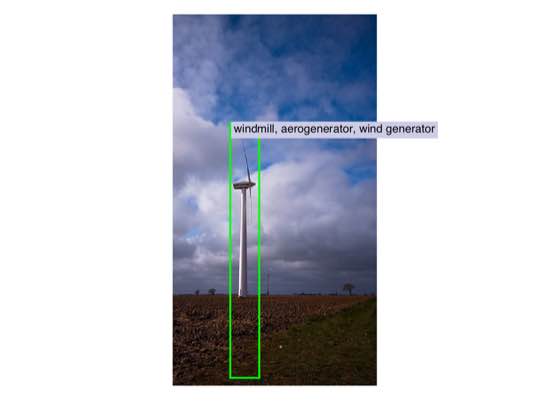} \\

\includegraphics[trim = 14mm 5mm 14mm 5mm , clip=true,width=0.25\textwidth,height=\textheight,keepaspectratio]{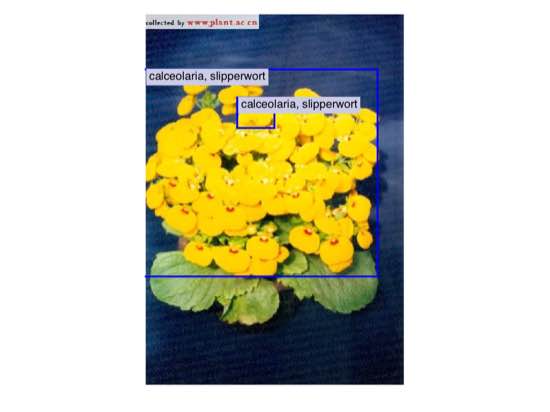} & 
\includegraphics[trim = 14mm 5mm 14mm 5mm , clip=true,width=0.25\textwidth,height=\textheight,keepaspectratio]{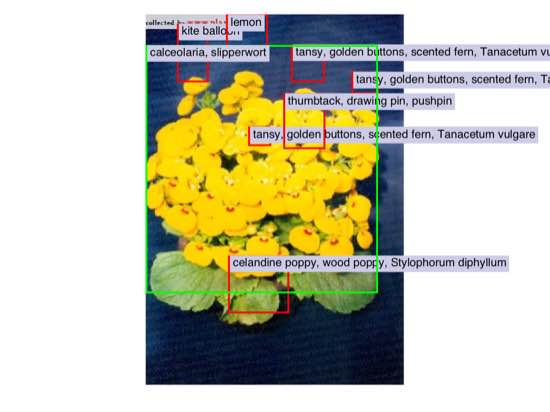} &
\includegraphics[trim = 14mm 5mm 14mm 5mm , clip=true,width=0.25\textwidth,height=\textheight,keepaspectratio]{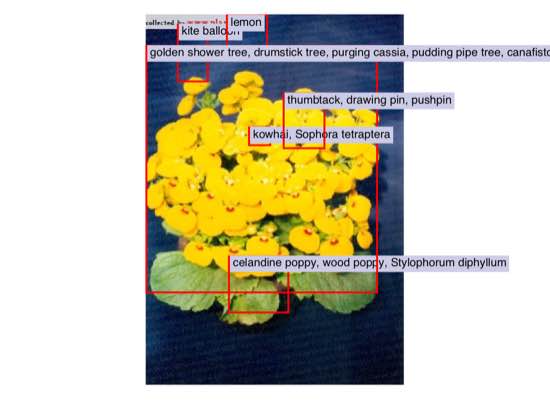} & 
\includegraphics[trim = 14mm 5mm 14mm 5mm , clip=true,width=0.25\textwidth,height=\textheight,keepaspectratio]{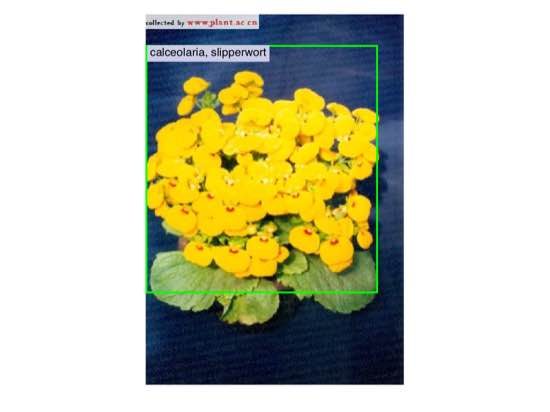} \\

\includegraphics[trim = 0mm 0mm 0mm 20mm , clip=true,width=0.25\textwidth,height=\textheight,keepaspectratio]{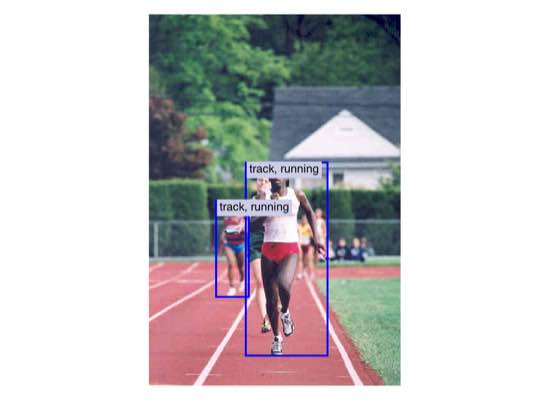} & 
\includegraphics[trim = 0mm 0mm 0mm 20mm , clip=true,width=0.25\textwidth,height=\textheight,keepaspectratio]{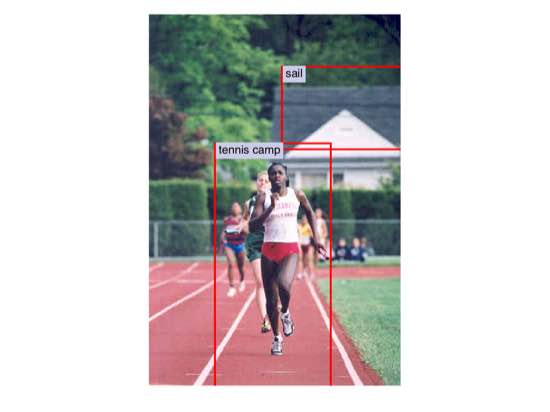} &
\includegraphics[trim = 0mm 0mm 0mm 20mm , clip=true,width=0.25\textwidth,height=\textheight,keepaspectratio]{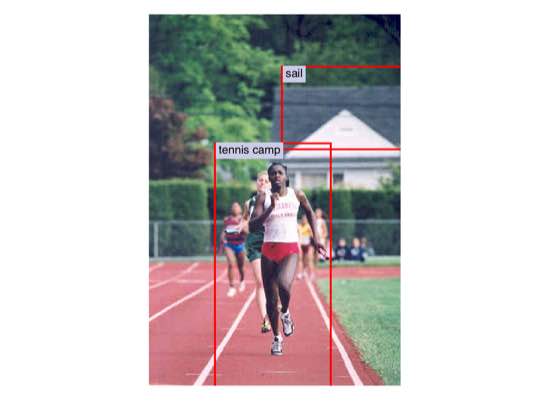} & 
\includegraphics[trim = 0mm 0mm 0mm 20mm , clip=true,width=0.25\textwidth,height=\textheight,keepaspectratio]{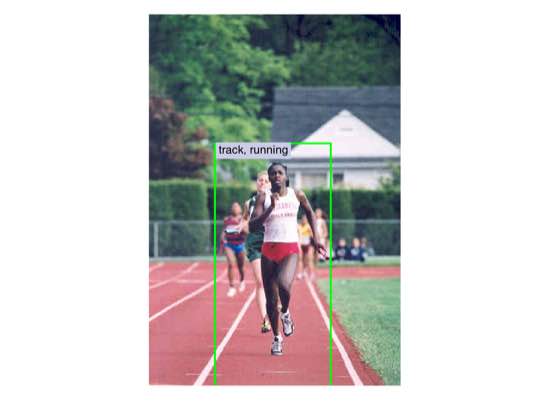} \\

\end{tabular}
\end{center}
\caption{Examples where MAPC outperforms WC+AC-NMS and SAPC + AC-NMS on 1,825 ImageNet categories. Ground truth: blue, true positives: green, false positives: red.}
\label{table:ImageNetBetterBest}
\end{figure*}

\begin{figure*}
\begin{center}
\begin{tabular}{c@{}c@{}c@{}c}
Ground Truth & WC+AC-NMS & SAPC + AC-NMS & MAPC (ours) \\
\hline\hline
\includegraphics[trim = 14mm 10mm 14mm 10mm , clip=true,width=0.25\textwidth,height=\textheight,keepaspectratio]{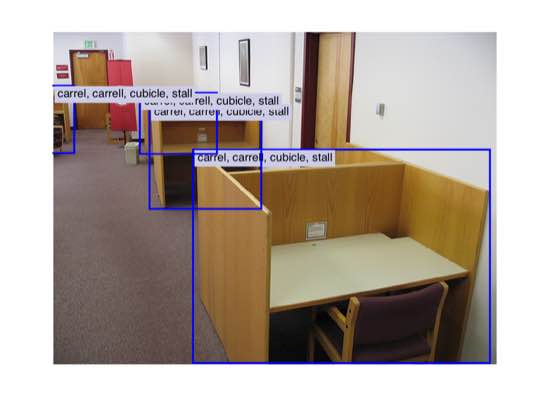} & 
\includegraphics[trim = 14mm 10mm 14mm 10mm , clip=true,width=0.25\textwidth,height=\textheight,keepaspectratio]{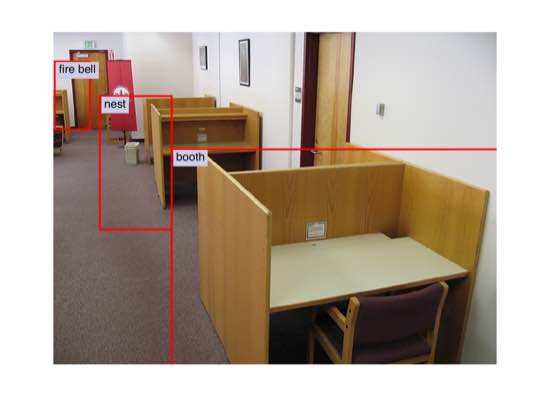} &
\includegraphics[trim = 14mm 10mm 14mm 10mm , clip=true,width=0.25\textwidth,height=\textheight,keepaspectratio]{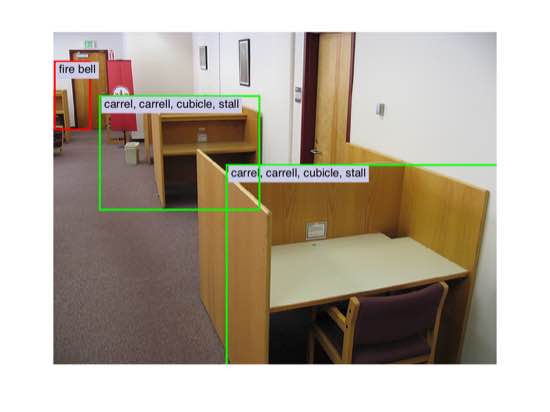} & 
\includegraphics[trim = 14mm 10mm 14mm 10mm , clip=true,width=0.25\textwidth,height=\textheight,keepaspectratio]{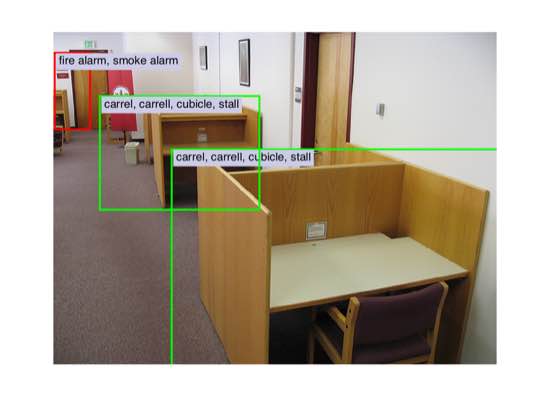} \\

\includegraphics[trim = 0mm 0mm 0mm 5mm , clip=true,width=0.25\textwidth,height=\textheight,keepaspectratio]{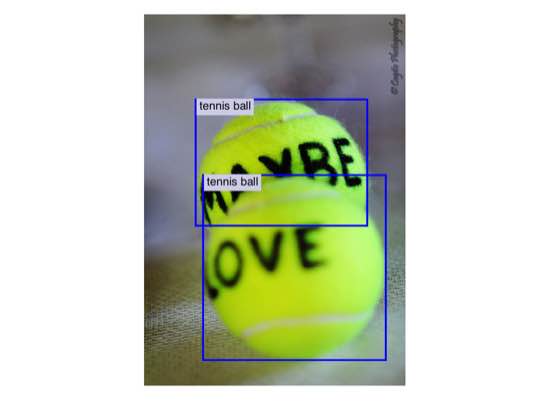} & 
\includegraphics[trim = 0mm 0mm 0mm 5mm , clip=true,width=0.25\textwidth,height=\textheight,keepaspectratio]{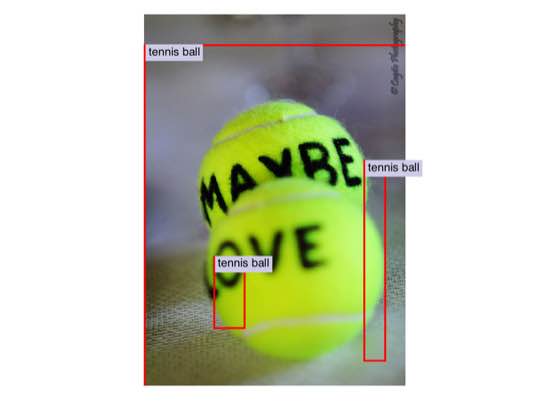} &
\includegraphics[trim = 0mm 0mm 0mm 5mm , clip=true,width=0.25\textwidth,height=\textheight,keepaspectratio]{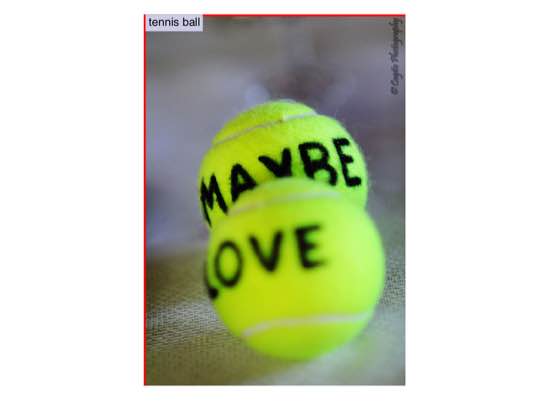} & 
\includegraphics[trim = 0mm 0mm 0mm 5mm , clip=true,width=0.25\textwidth,height=\textheight,keepaspectratio]{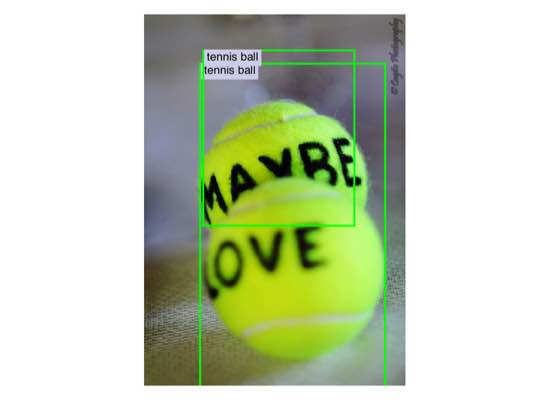} \\

\includegraphics[trim = 0mm 0mm 0mm 5mm , clip=true,width=0.25\textwidth,height=\textheight,keepaspectratio]{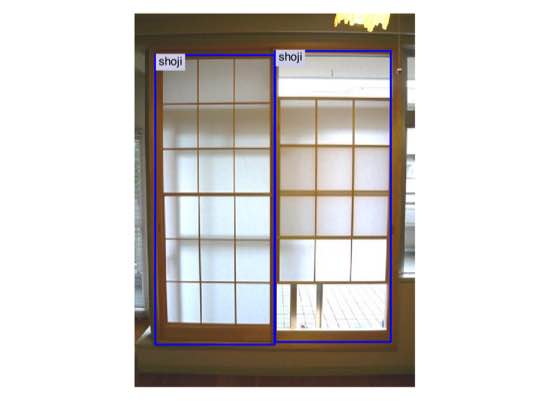} & 
\includegraphics[trim = 0mm 0mm 0mm 5mm , clip=true,width=0.25\textwidth,height=\textheight,keepaspectratio]{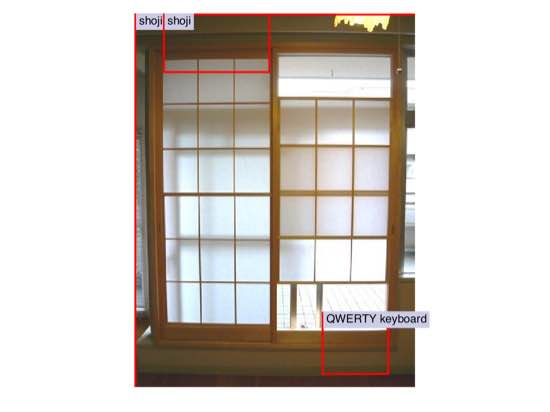} &
\includegraphics[trim = 0mm 0mm 0mm 5mm , clip=true,width=0.25\textwidth,height=\textheight,keepaspectratio]{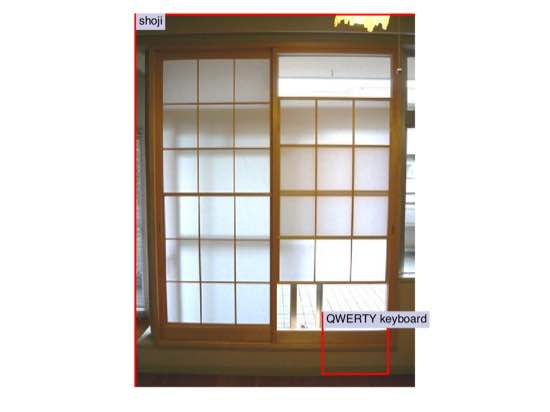} & 
\includegraphics[trim = 0mm 0mm 0mm 5mm , clip=true,width=0.25\textwidth,height=\textheight,keepaspectratio]{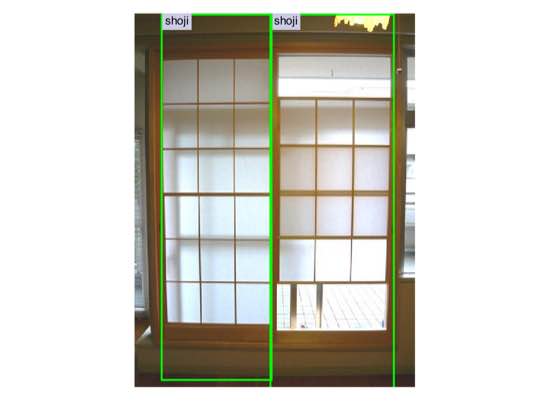} \\

\includegraphics[trim = 0mm 0mm 0mm 5mm , clip=true,width=0.25\textwidth,height=\textheight,keepaspectratio]{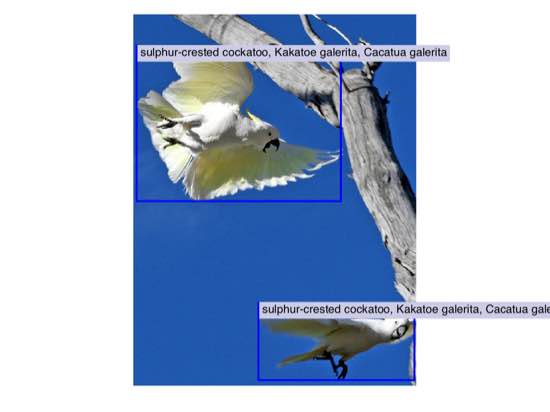} & 
\includegraphics[trim = 0mm 0mm 0mm 5mm , clip=true,width=0.25\textwidth,height=\textheight,keepaspectratio]{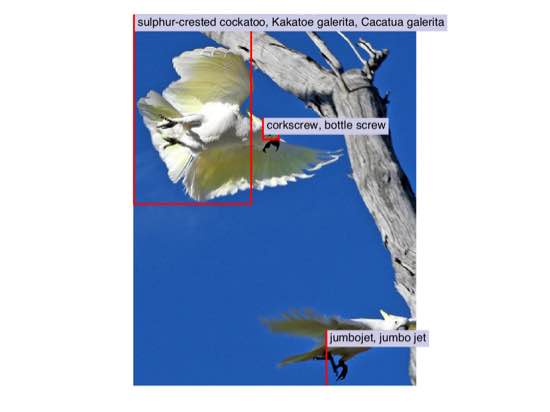} &
\includegraphics[trim = 0mm 0mm 0mm 5mm , clip=true,width=0.25\textwidth,height=\textheight,keepaspectratio]{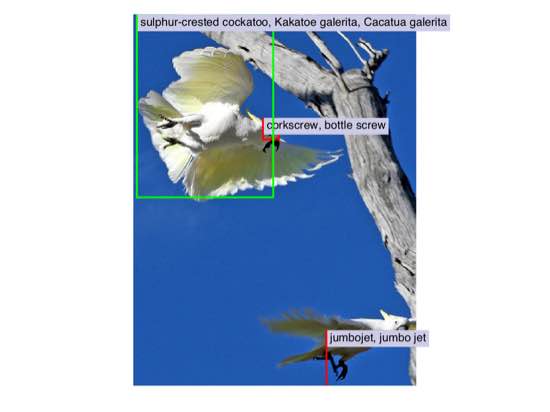} & 
\includegraphics[trim = 0mm 0mm 0mm 5mm , clip=true,width=0.25\textwidth,height=\textheight,keepaspectratio]{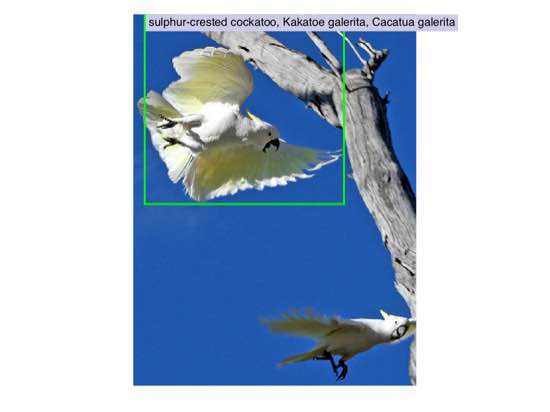} \\

\includegraphics[trim = 5mm 5mm 5mm 5mm , clip=true,width=0.25\textwidth,height=\textheight,keepaspectratio]{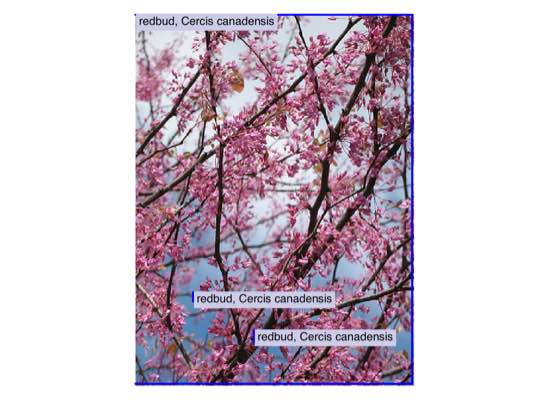} & 
\includegraphics[trim = 5mm 5mm 5mm 5mm , clip=true,width=0.25\textwidth,height=\textheight,keepaspectratio]{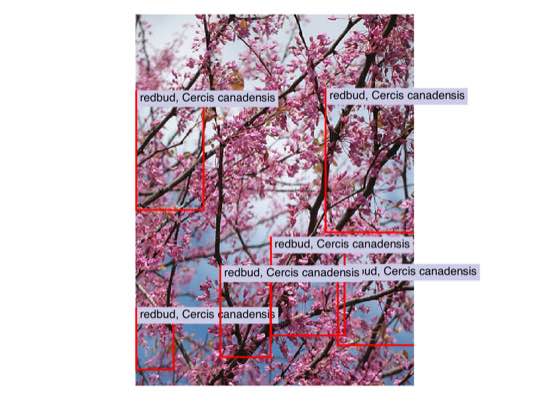} &
\includegraphics[trim = 5mm 5mm 5mm 5mm , clip=true,width=0.25\textwidth,height=\textheight,keepaspectratio]{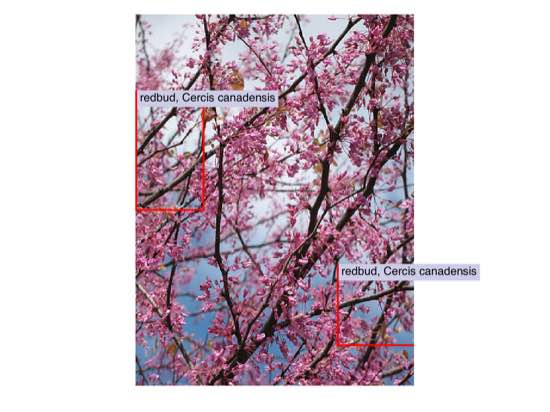} & 
\includegraphics[trim = 5mm 5mm 5mm 5mm , clip=true,width=0.25\textwidth,height=\textheight,keepaspectratio]{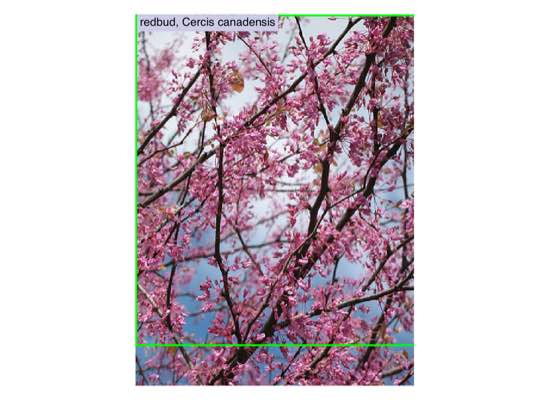} \\

\includegraphics[trim = 14mm 30mm 14mm 20mm , clip=true,width=0.25\textwidth,height=\textheight,keepaspectratio]{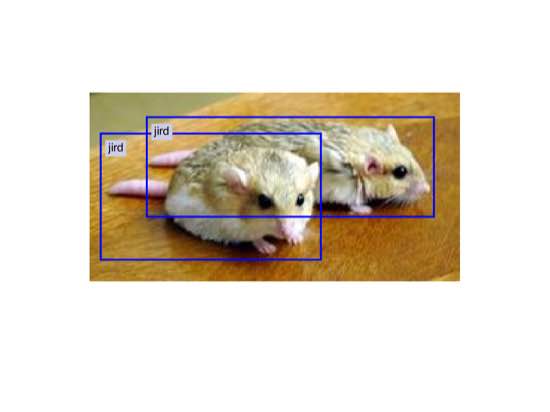} & 
\includegraphics[trim = 14mm 30mm 14mm 20mm , clip=true,width=0.25\textwidth,height=\textheight,keepaspectratio]{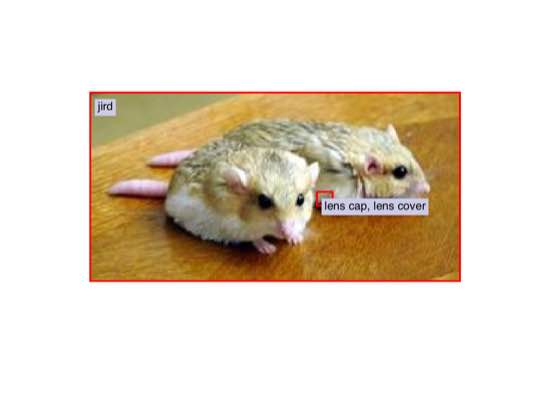} &
\includegraphics[trim = 14mm 30mm 14mm 20mm , clip=true,width=0.25\textwidth,height=\textheight,keepaspectratio]{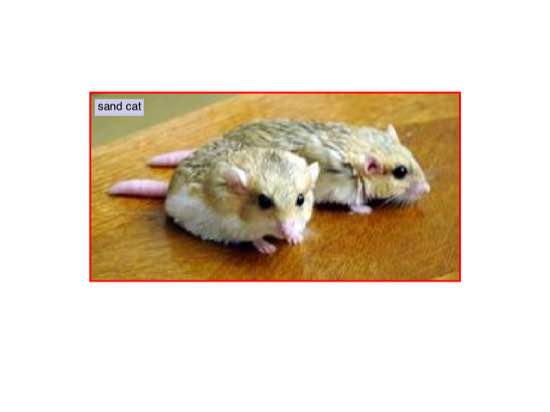} & 
\includegraphics[trim = 14mm 30mm 14mm 20mm , clip=true,width=0.25\textwidth,height=\textheight,keepaspectratio]{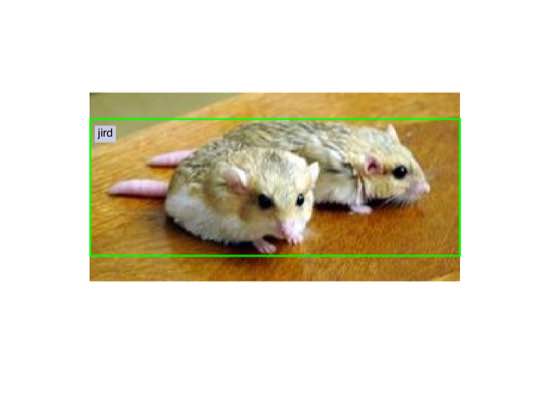} \\

\end{tabular}
\end{center}
\caption{Examples where MAPC better localises objects than WC+AC-NMS and SAPC + AC-NMS on 1,825 ImageNet classes. Ground truth: blue, true positives: green, false positives: red.}
\label{table:ImageNetBetterLoc}
\end{figure*}

\begin{figure*}
\begin{center}
\begin{tabular}{c@{}c@{}c@{}c}
Ground Truth & WC+AC-NMS & SAPC + AC-NMS & MAPC (ours) \\
\hline\hline
\includegraphics[trim = 5mm 20mm 5mm 10mm , clip=true,width=0.25\textwidth,height=\textheight,keepaspectratio]{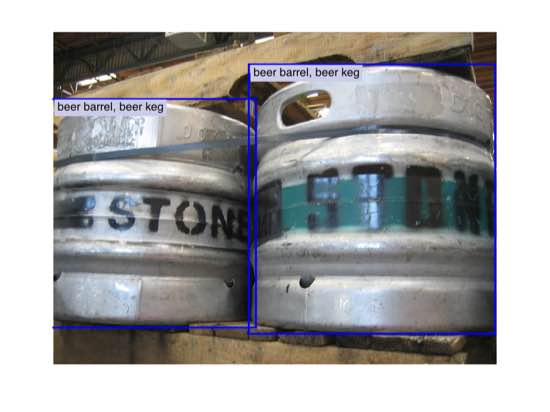} & 
\includegraphics[trim = 5mm 20mm 5mm 10mm , clip=true,width=0.25\textwidth,height=\textheight,keepaspectratio]{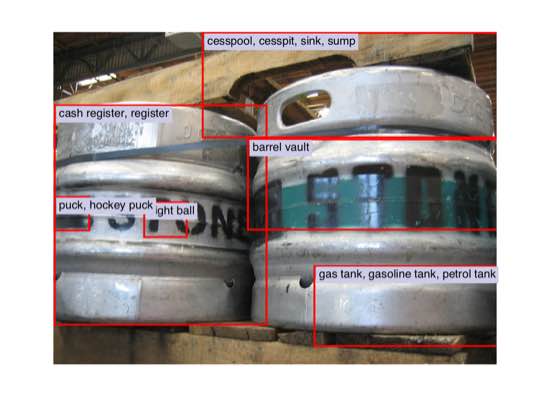} &
\includegraphics[trim = 5mm 20mm 5mm 10mm , clip=true,width=0.25\textwidth,height=\textheight,keepaspectratio]{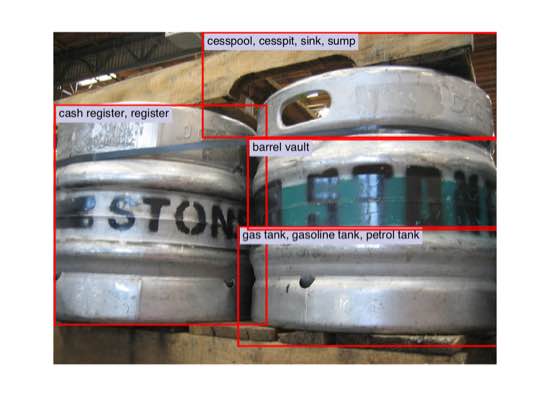} & 
\includegraphics[trim = 5mm 20mm 5mm 10mm , clip=true,width=0.25\textwidth,height=\textheight,keepaspectratio]{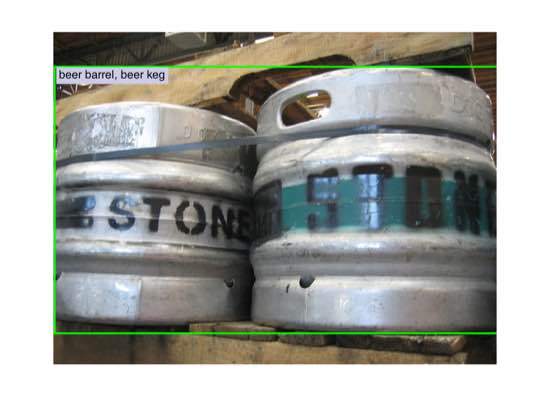} \\

\includegraphics[trim = 14mm 20mm 14mm 10mm , clip=true,width=0.25\textwidth,height=\textheight,keepaspectratio]{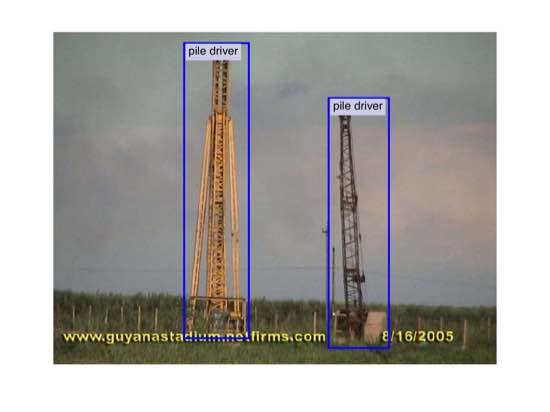} & 
\includegraphics[trim = 14mm 20mm 14mm 10mm , clip=true,width=0.25\textwidth,height=\textheight,keepaspectratio]{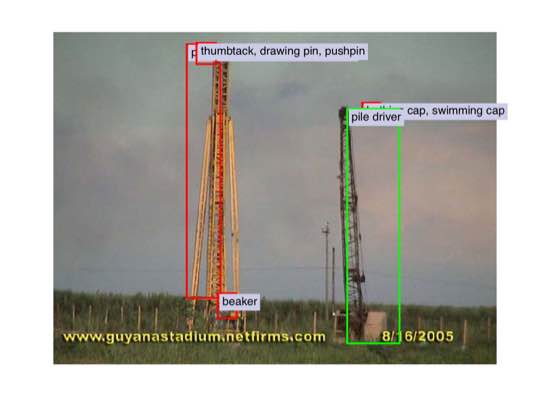} &
\includegraphics[trim = 14mm 20mm 14mm 10mm , clip=true,width=0.25\textwidth,height=\textheight,keepaspectratio]{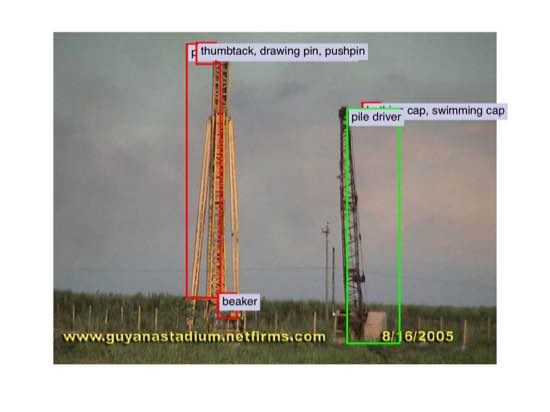} & 
\includegraphics[trim = 14mm 20mm 14mm 10mm , clip=true,width=0.25\textwidth,height=\textheight,keepaspectratio]{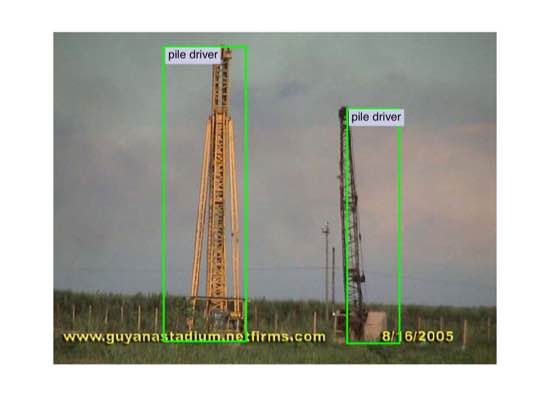} \\

\includegraphics[trim = 5mm 20mm 5mm 10mm , clip=true,width=0.25\textwidth,height=\textheight,keepaspectratio]{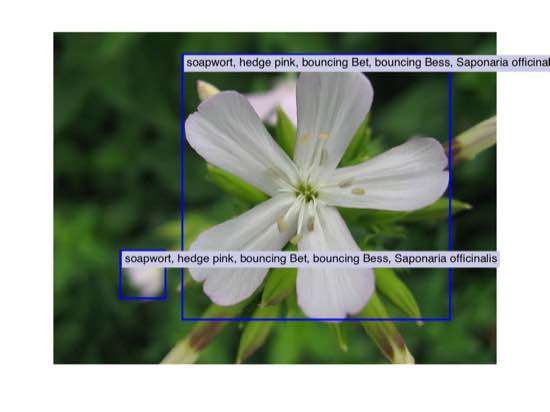} & 
\includegraphics[trim = 5mm 20mm 5mm 10mm , clip=true,width=0.25\textwidth,height=\textheight,keepaspectratio]{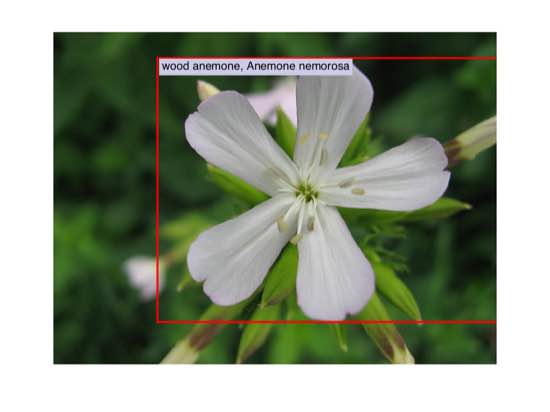} &
\includegraphics[trim = 5mm 20mm 5mm 10mm , clip=true,width=0.25\textwidth,height=\textheight,keepaspectratio]{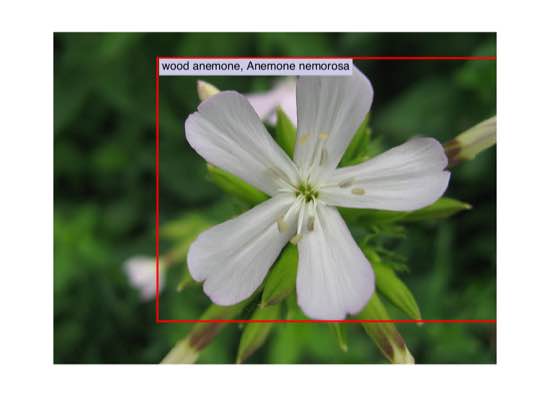} & 
\includegraphics[trim = 5mm 20mm 5mm 10mm , clip=true,width=0.25\textwidth,height=\textheight,keepaspectratio]{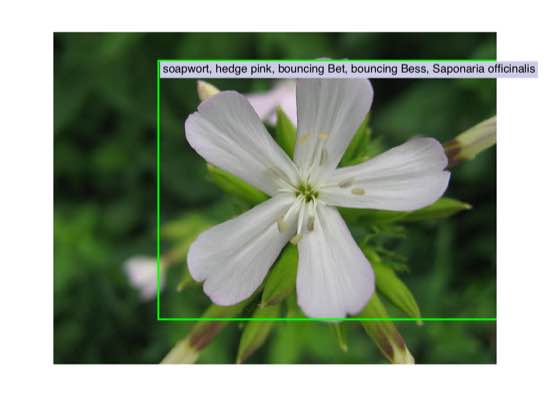} \\

\includegraphics[trim = 14mm 20mm 14mm 0mm , clip=true,width=0.25\textwidth,height=\textheight,keepaspectratio]{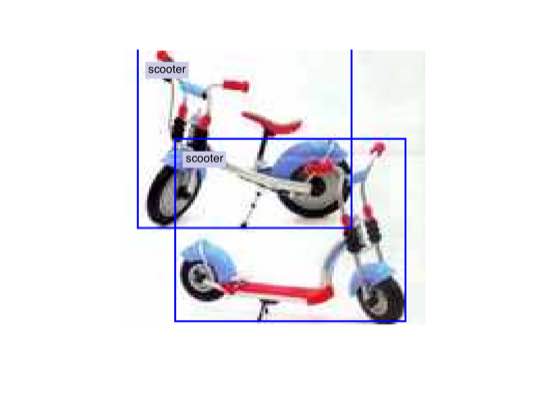} & 
\includegraphics[trim = 14mm 20mm 14mm 0mm , clip=true,width=0.25\textwidth,height=\textheight,keepaspectratio]{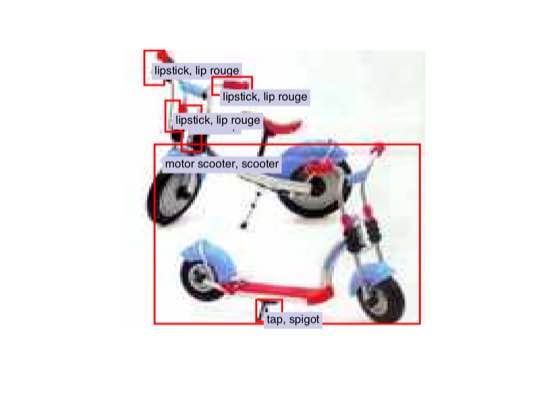} &
\includegraphics[trim = 14mm 20mm 14mm 0mm , clip=true,width=0.25\textwidth,height=\textheight,keepaspectratio]{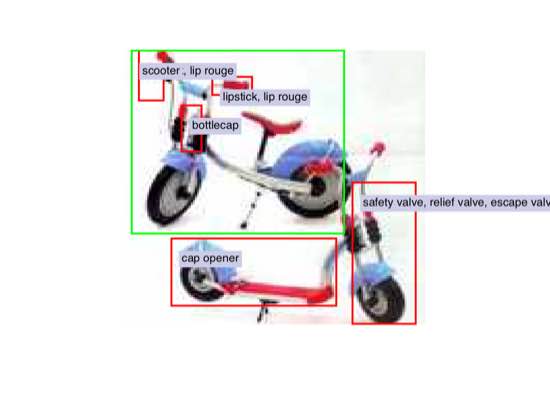} & 
\includegraphics[trim = 14mm 20mm 14mm 0mm , clip=true,width=0.25\textwidth,height=\textheight,keepaspectratio]{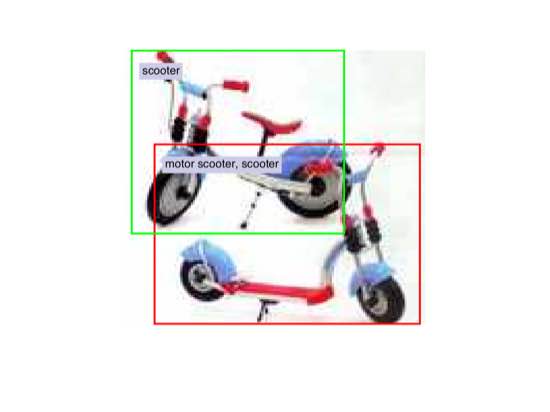} \\

\includegraphics[trim = 14mm 20mm 14mm 10mm , clip=true,width=0.25\textwidth,height=\textheight,keepaspectratio]{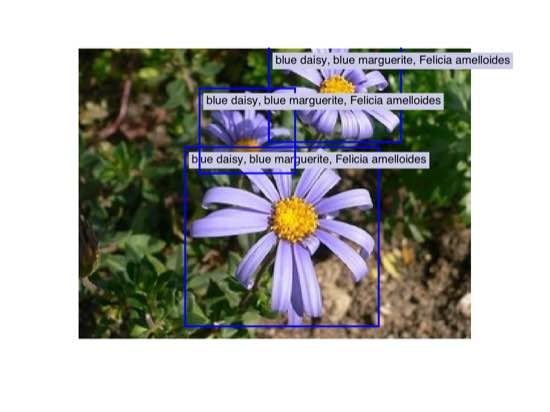} & 
\includegraphics[trim = 14mm 20mm 14mm 10mm , clip=true,width=0.25\textwidth,height=\textheight,keepaspectratio]{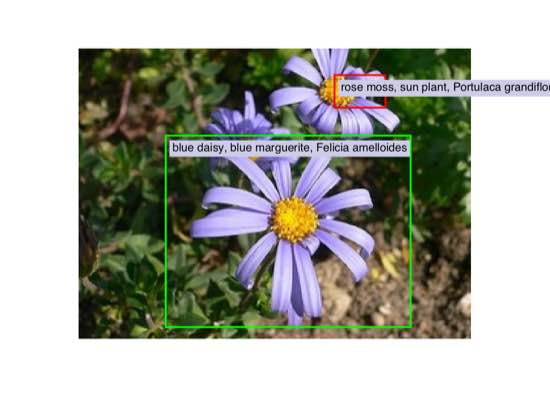} &
\includegraphics[trim = 14mm 20mm 14mm 10mm , clip=true,width=0.25\textwidth,height=\textheight,keepaspectratio]{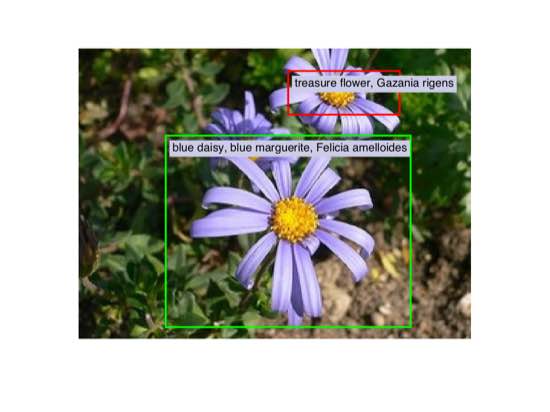} & 
\includegraphics[trim = 14mm 20mm 14mm 10mm , clip=true,width=0.25\textwidth,height=\textheight,keepaspectratio]{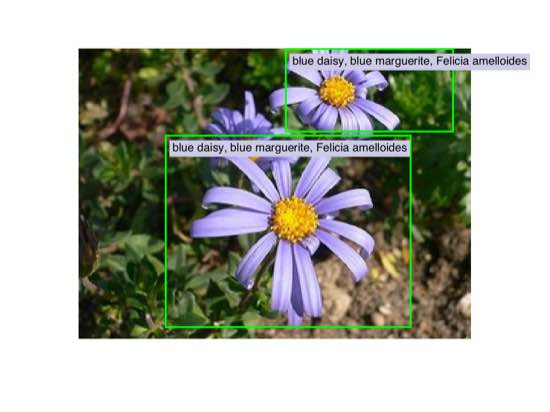} \\

\includegraphics[trim = 14mm 0mm 14mm 10mm , clip=true,width=0.25\textwidth,height=\textheight,keepaspectratio]{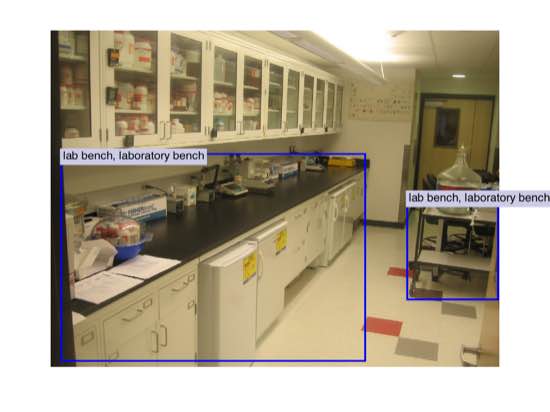} & 
\includegraphics[trim = 14mm 0mm 14mm 10mm , clip=true,width=0.25\textwidth,height=\textheight,keepaspectratio]{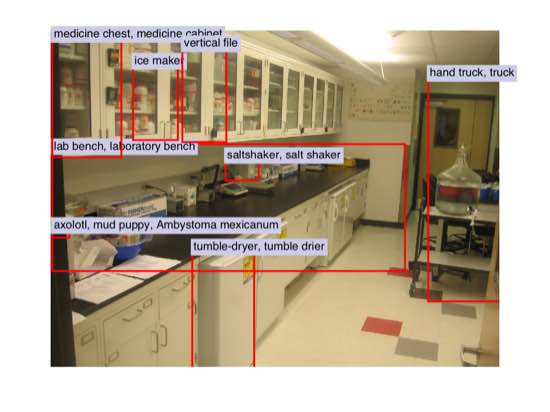} &
\includegraphics[trim = 14mm 0mm 14mm 10mm , clip=true,width=0.25\textwidth,height=\textheight,keepaspectratio]{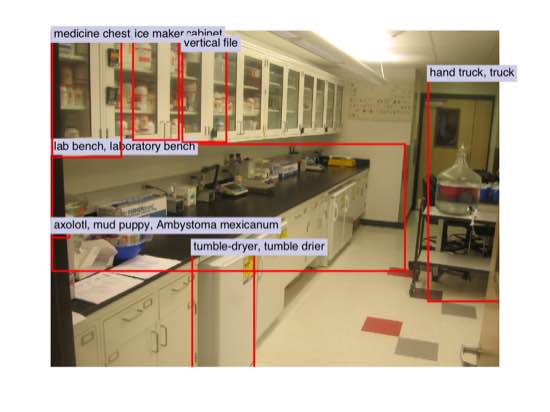} & 
\includegraphics[trim = 14mm 0mm 14mm 10mm , clip=true,width=0.25\textwidth,height=\textheight,keepaspectratio]{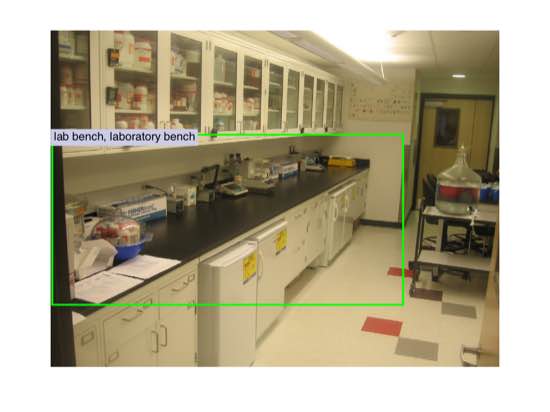} \\

\end{tabular}
\end{center}
\caption{Examples where MAPC better classifies objects than WC+AC-NMS and SAPC + AC-NMS on 1,825 ImageNet classes. Ground truth: blue, true positives: green, false positives: red.}
\label{table:ImageNetBetterClass}
\end{figure*}

\begin{figure*}
\begin{center}
\begin{tabular}{c@{}c@{}c}
WC+AC-NMS & SAPC + AC-NMS & MAPC (ours) \\
\hline\hline

\includegraphics[trim = 14mm 20mm 14mm 20mm , clip=true,width=0.25\textwidth,height=\textheight,keepaspectratio]{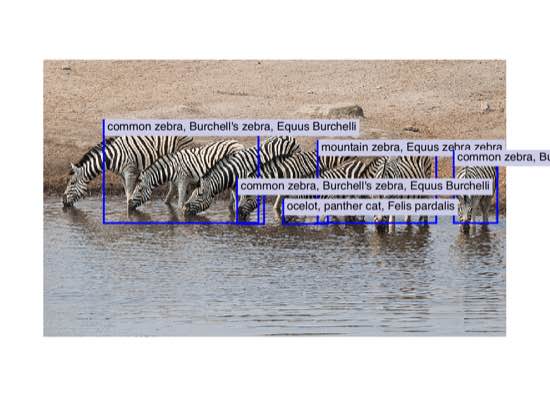} & 
\includegraphics[trim = 14mm 20mm 14mm 20mm , clip=true,width=0.25\textwidth,height=\textheight,keepaspectratio]{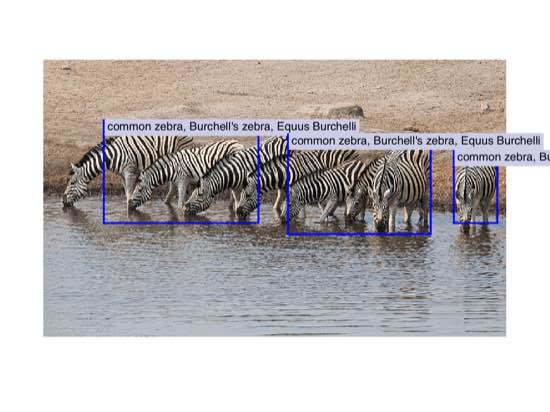} & 
\includegraphics[trim = 14mm 20mm 14mm 20mm , clip=true,width=0.25\textwidth,height=\textheight,keepaspectratio]{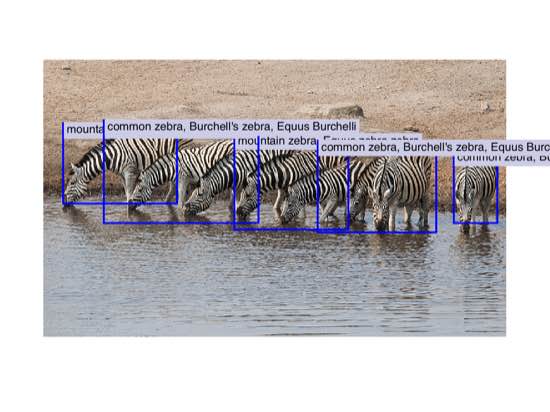} \\

\includegraphics[trim = 0mm 0mm 0mm 5mm , clip=true,width=0.25\textwidth,height=\textheight,keepaspectratio]{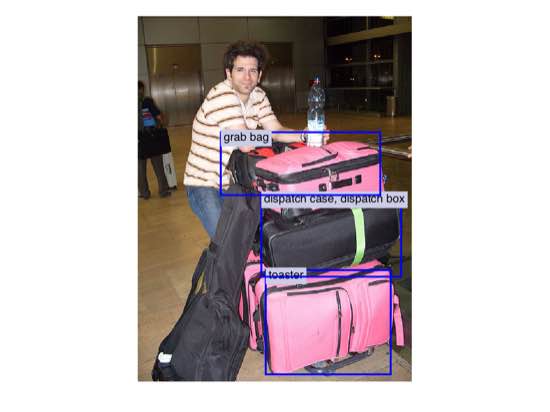} & 
\includegraphics[trim = 0mm 0mm 0mm 5mm , clip=true,width=0.25\textwidth,height=\textheight,keepaspectratio]{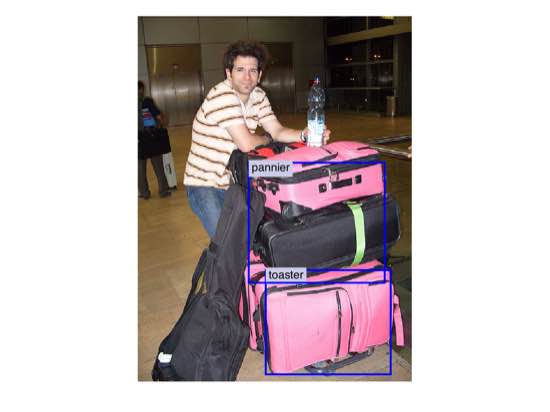} & 
\includegraphics[trim = 0mm 0mm 0mm 5mm , clip=true,width=0.25\textwidth,height=\textheight,keepaspectratio]{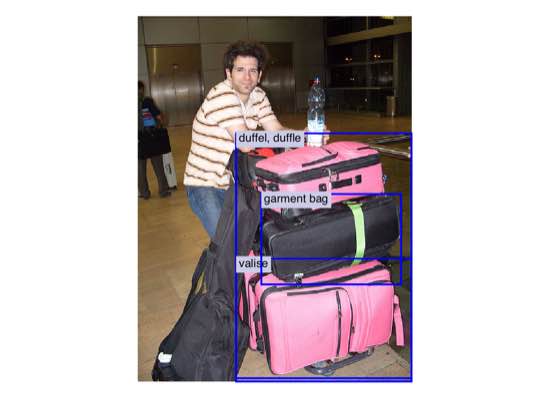} \\

\includegraphics[trim = 0mm 0mm 0mm 5mm , clip=true,width=0.25\textwidth,height=\textheight,keepaspectratio]{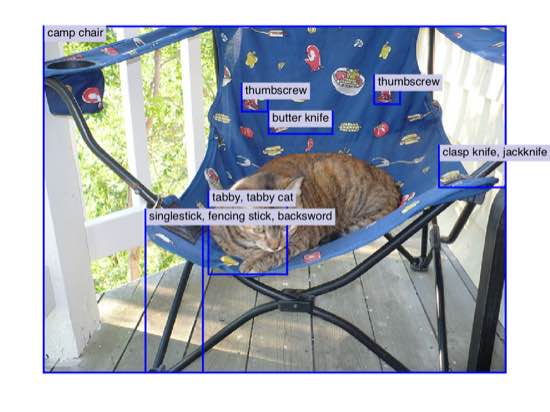} & 
\includegraphics[trim = 0mm 0mm 0mm 5mm , clip=true,width=0.25\textwidth,height=\textheight,keepaspectratio]{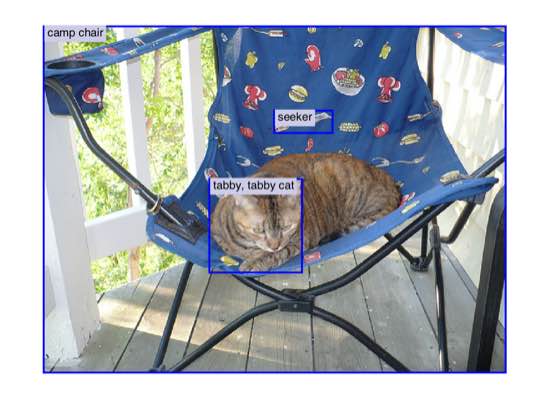} & 
\includegraphics[trim = 0mm 0mm 0mm 5mm , clip=true,width=0.25\textwidth,height=\textheight,keepaspectratio]{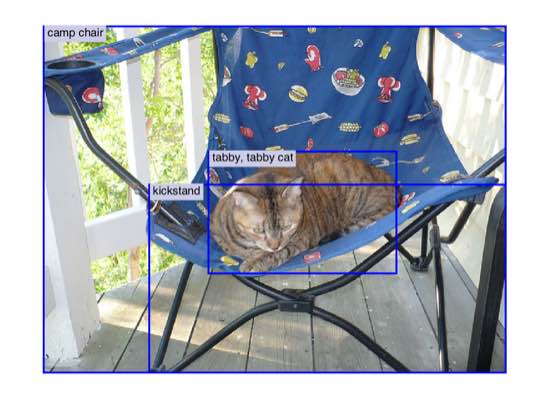} \\

\includegraphics[trim = 0mm 0mm 0mm 0mm , clip=true,width=0.25\textwidth,height=\textheight,keepaspectratio]{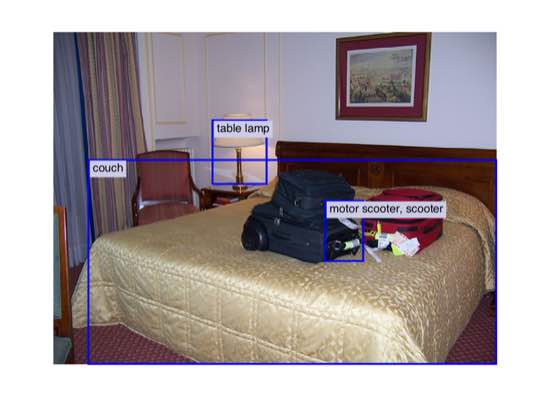} & 
\includegraphics[trim = 0mm 0mm 0mm 0mm, clip=true,width=0.25\textwidth,height=\textheight,keepaspectratio]{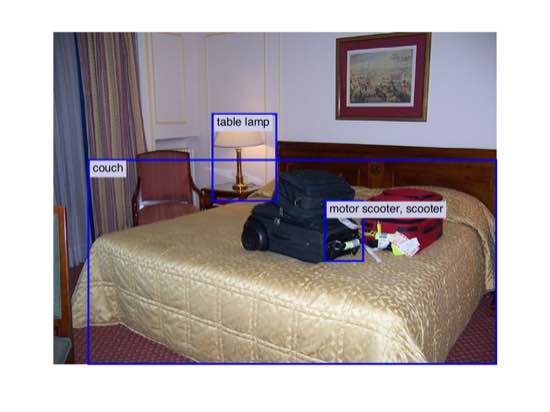} & 
\includegraphics[trim = 0mm 0mm 0mm 0mm , clip=true,width=0.25\textwidth,height=\textheight,keepaspectratio]{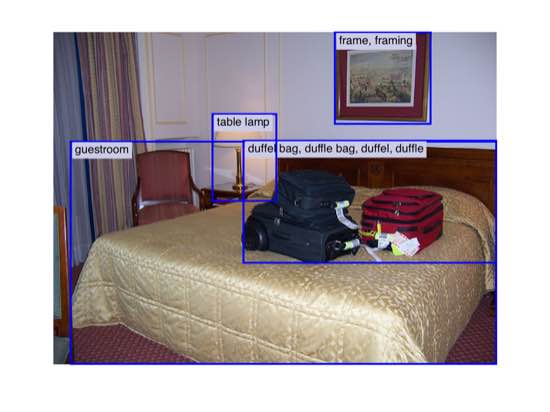} \\

\includegraphics[trim = 14mm 20mm 14mm 20mm , clip=true,width=0.25\textwidth,height=\textheight,keepaspectratio]{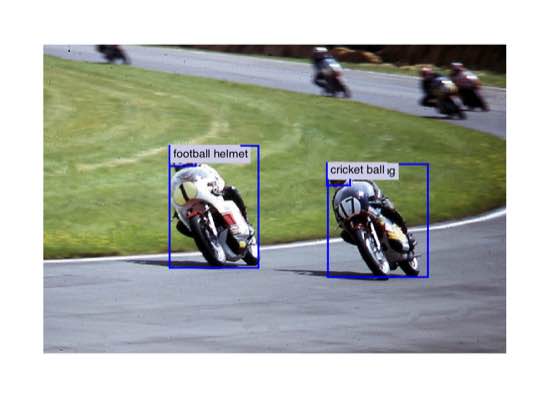} & 
\includegraphics[trim = 14mm 20mm 14mm 20mm , clip=true,width=0.25\textwidth,height=\textheight,keepaspectratio]{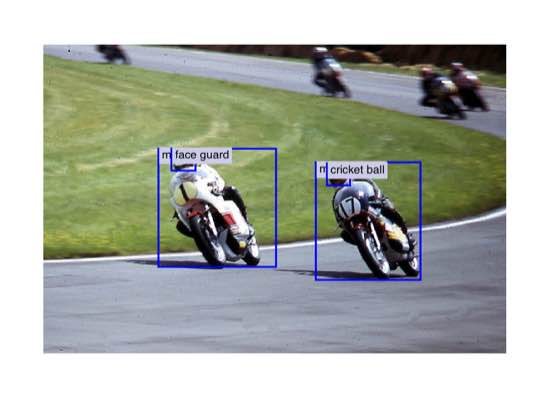} & 
\includegraphics[trim = 14mm 20mm 14mm 20mm , clip=true,width=0.25\textwidth,height=\textheight,keepaspectratio]{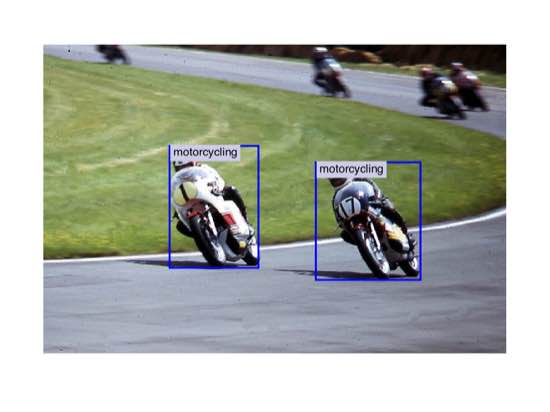} \\

\includegraphics[trim = 14mm 20mm 14mm 15mm , clip=true,width=0.25\textwidth,height=\textheight,keepaspectratio]{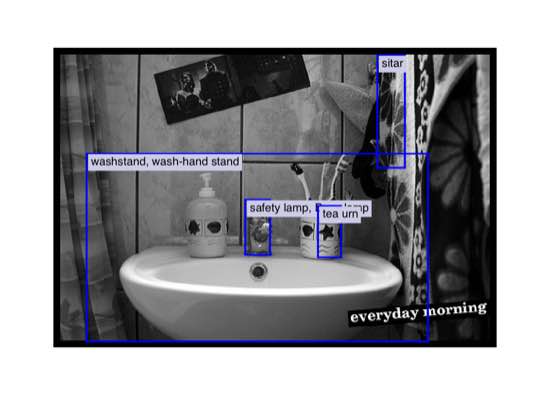} & 
\includegraphics[trim = 14mm 20mm 14mm 15mm , clip=true,width=0.25\textwidth,height=\textheight,keepaspectratio]{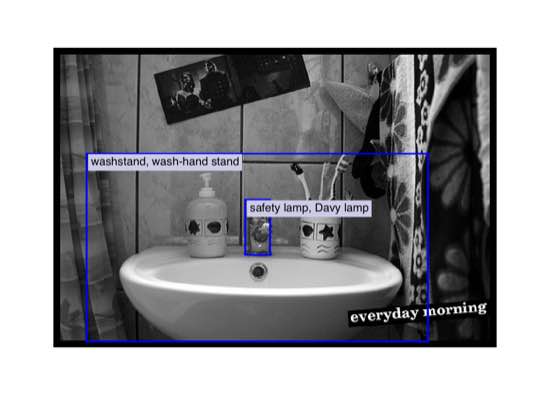} & 
\includegraphics[trim = 14mm 20mm 14mm 15mm , clip=true,width=0.25\textwidth,height=\textheight,keepaspectratio]{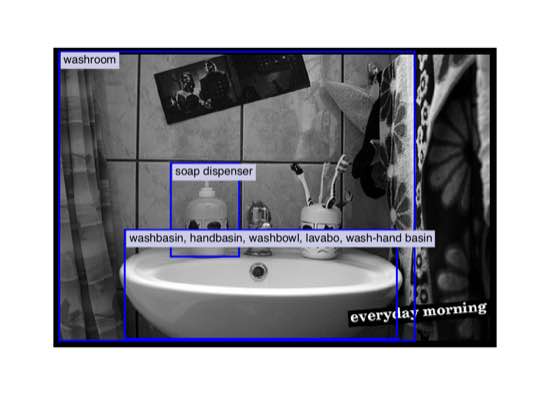} \\

\end{tabular}
\end{center}
\caption{Examples where MAPC outperforms WC+AC-NMS and SAPC + AC-NMS on 7,404 ImageNet categories. No ground truth is available. Detections: blue.}
\label{table:LargeBetterBest}
\end{figure*}

\begin{figure*}
\begin{center}
\begin{tabular}{c@{}c@{}c}
WC+AC-NMS & SAPC + AC-NMS & MAPC (ours) \\
\hline\hline

\includegraphics[trim = 0mm 0mm 0mm 0mm , clip=true,width=0.25\textwidth,height=\textheight,keepaspectratio]{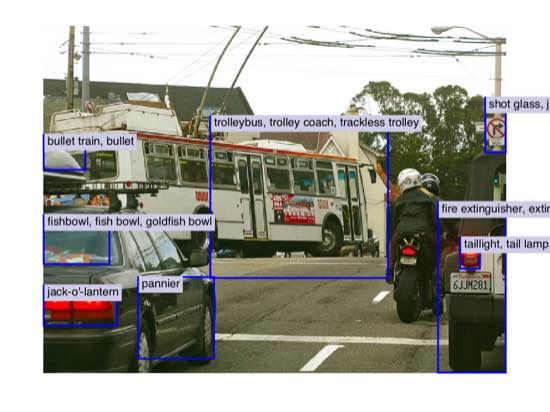} & 
\includegraphics[trim = 0mm 0mm 0mm 0mm , clip=true,width=0.25\textwidth,height=\textheight,keepaspectratio]{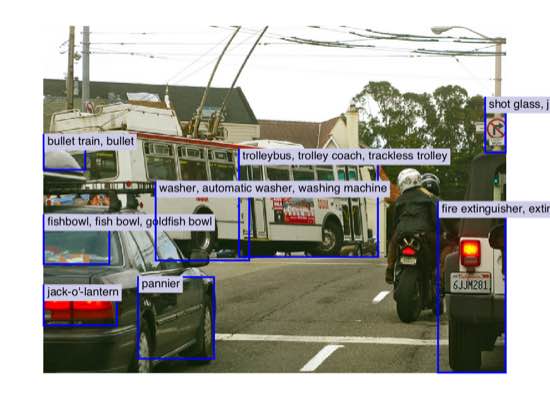} & 
\includegraphics[trim = 0mm 0mm 0mm 0mm , clip=true,width=0.25\textwidth,height=\textheight,keepaspectratio]{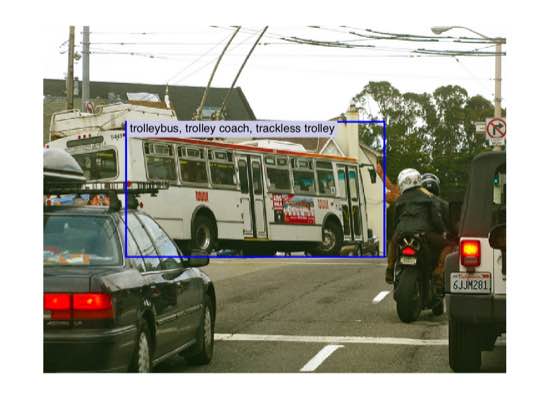} \\

\includegraphics[trim = 14mm 20mm 14mm 20mm , clip=true,width=0.25\textwidth,height=\textheight,keepaspectratio]{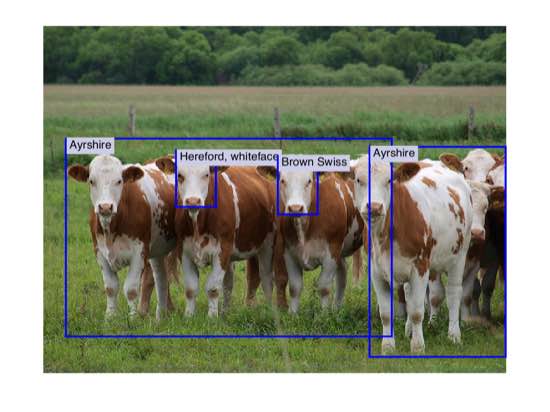} & 
\includegraphics[trim = 14mm 20mm 14mm 20mm , clip=true,width=0.25\textwidth,height=\textheight,keepaspectratio]{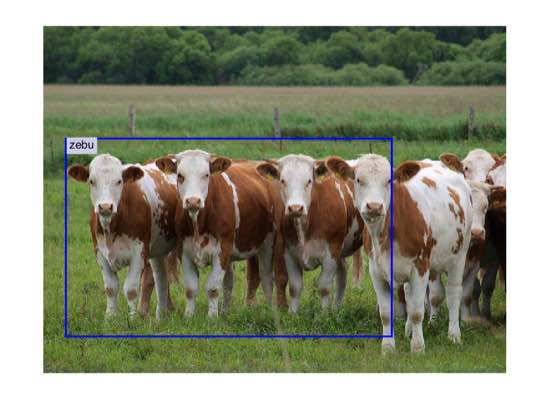} & 
\includegraphics[trim = 14mm 20mm 14mm 20mm , clip=true,width=0.25\textwidth,height=\textheight,keepaspectratio]{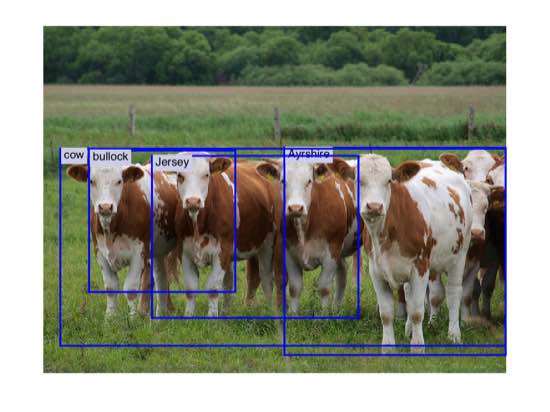} \\

\includegraphics[trim = 14mm 20mm 14mm 20mm , clip=true,width=0.25\textwidth,height=\textheight,keepaspectratio]{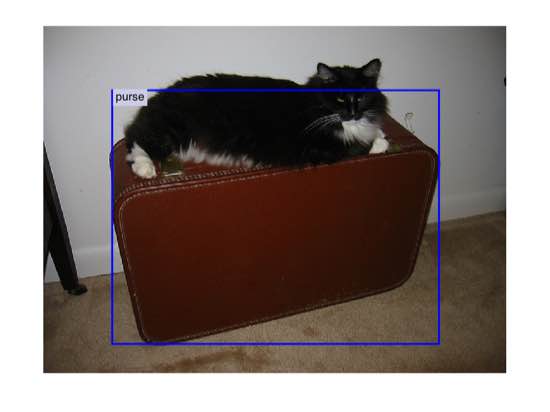} & 
\includegraphics[trim = 14mm 20mm 14mm 20mm , clip=true,width=0.25\textwidth,height=\textheight,keepaspectratio]{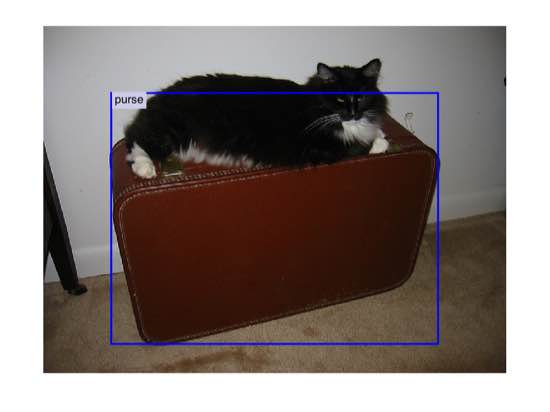} & 
\includegraphics[trim = 14mm 20mm 14mm 20mm , clip=true,width=0.25\textwidth,height=\textheight,keepaspectratio]{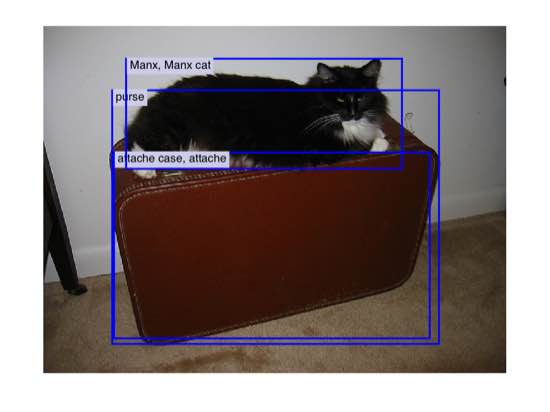} \\

\includegraphics[trim = 14mm 5mm 14mm 20mm , clip=true,width=0.25\textwidth,height=\textheight,keepaspectratio]{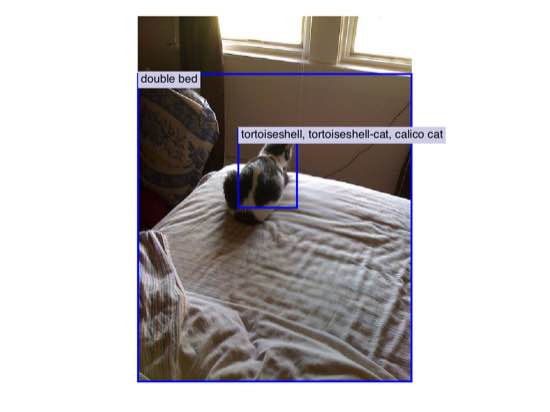} & 
\includegraphics[trim = 14mm 5mm 14mm 20mm , clip=true,width=0.25\textwidth,height=\textheight,keepaspectratio]{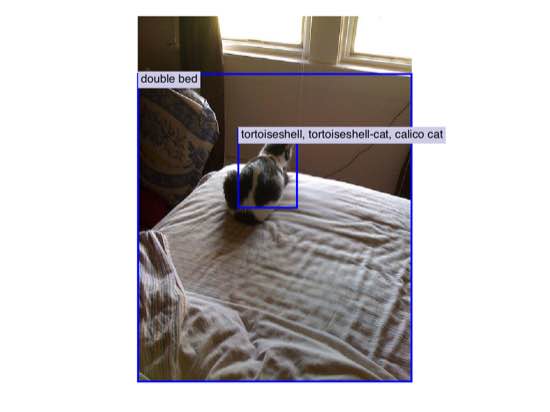} & 
\includegraphics[trim = 14mm 5mm 14mm 20mm , clip=true,width=0.25\textwidth,height=\textheight,keepaspectratio]{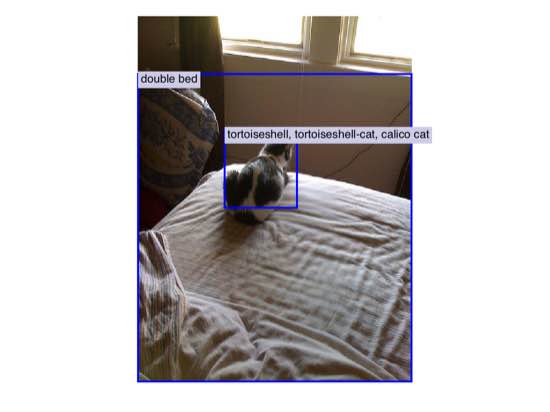} \\

\includegraphics[trim = 14mm 20mm 14mm 5mm , clip=true,width=0.25\textwidth,height=\textheight,keepaspectratio]{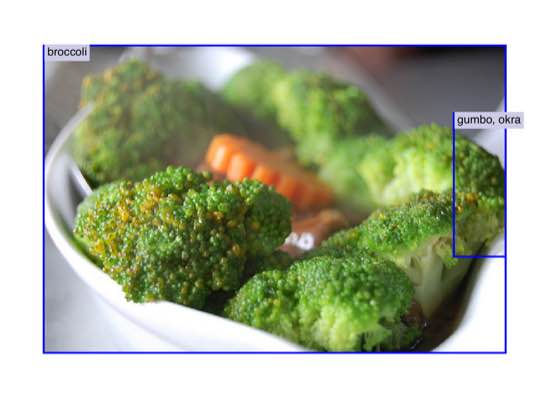} & 
\includegraphics[trim = 14mm 20mm 14mm 5mm , clip=true,width=0.25\textwidth,height=\textheight,keepaspectratio]{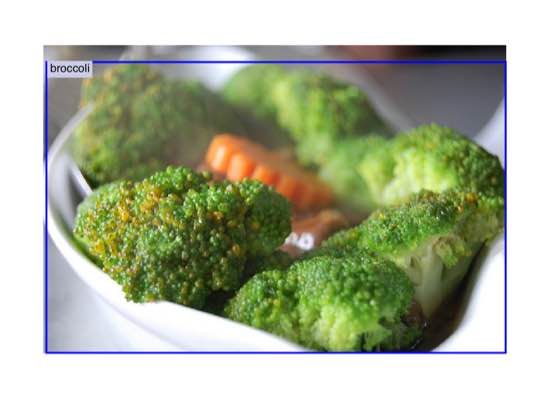} & 
\includegraphics[trim = 14mm 20mm 14mm 5mm , clip=true,width=0.25\textwidth,height=\textheight,keepaspectratio]{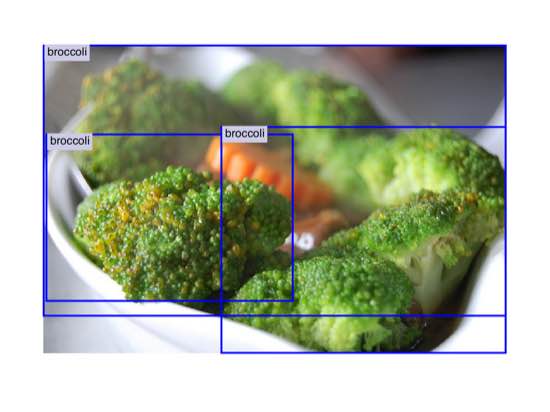} \\

\includegraphics[trim = 9mm 0mm 9mm 0mm , clip=true,width=0.25\textwidth,height=\textheight,keepaspectratio]{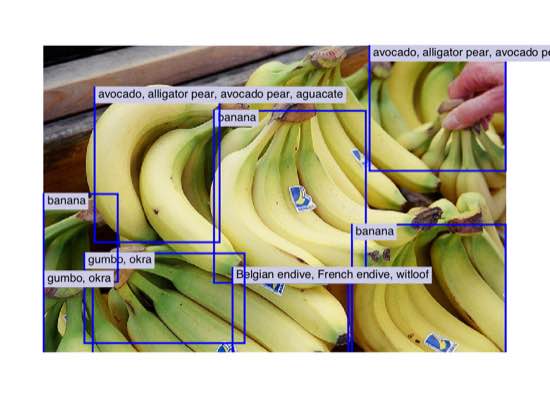} & 
\includegraphics[trim = 9mm 0mm 9mm 0mm , clip=true,width=0.25\textwidth,height=\textheight,keepaspectratio]{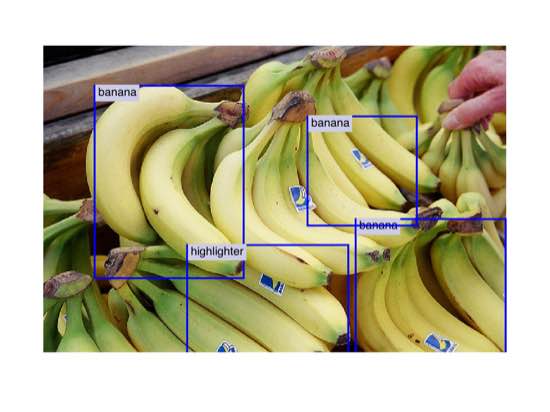} & 
\includegraphics[trim = 9mm 0mm 9mm 0mm , clip=true,width=0.25\textwidth,height=\textheight,keepaspectratio]{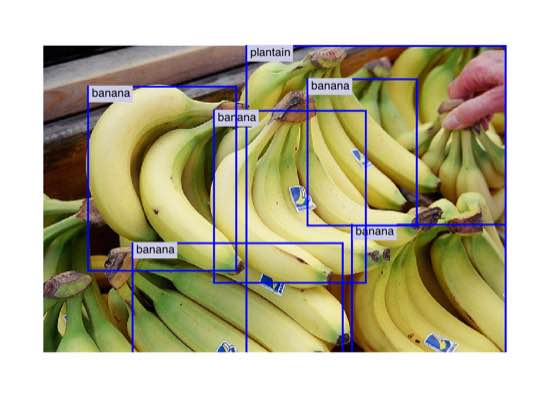} \\

\end{tabular}
\end{center}
\caption{Examples where MAPC better localises objects than WC+AC-NMS and SAPC + AC-NMS on 7,404 ImageNet classes. No ground truth is available. Detections: blue.}
\label{table:LargeBetterLoc}
\end{figure*}

\begin{figure*}
\begin{center}
\begin{tabular}{c@{}c@{}c}
WC+AC-NMS & SAPC + AC-NMS & MAPC (ours) \\
\hline\hline

\includegraphics[trim = 14mm 5mm 14mm 5mm , clip=true,width=0.25\textwidth,height=\textheight,keepaspectratio]{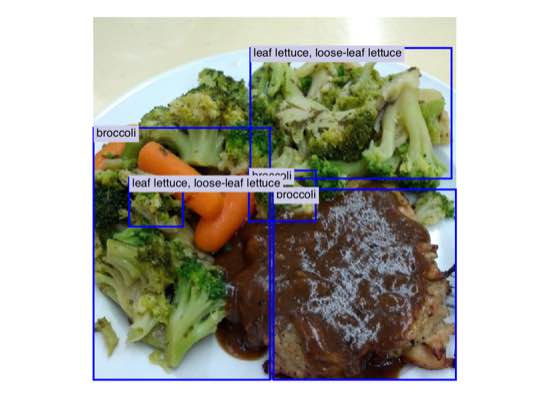} & 
\includegraphics[trim = 14mm 5mm 14mm 5mm , clip=true,width=0.25\textwidth,height=\textheight,keepaspectratio]{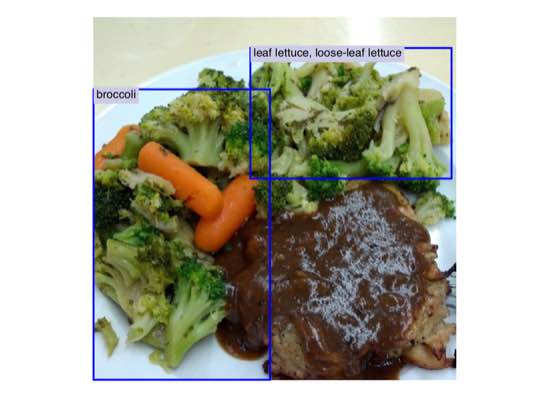} & 
\includegraphics[trim = 14mm 5mm 14mm 5mm , clip=true,width=0.25\textwidth,height=\textheight,keepaspectratio]{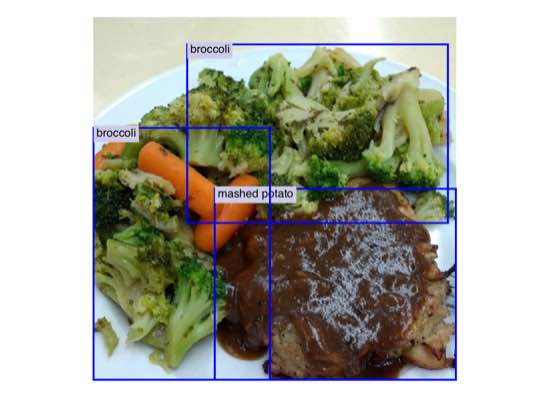} \\

\includegraphics[trim = 14mm 5mm 14mm 20mm , clip=true,width=0.25\textwidth,height=\textheight,keepaspectratio]{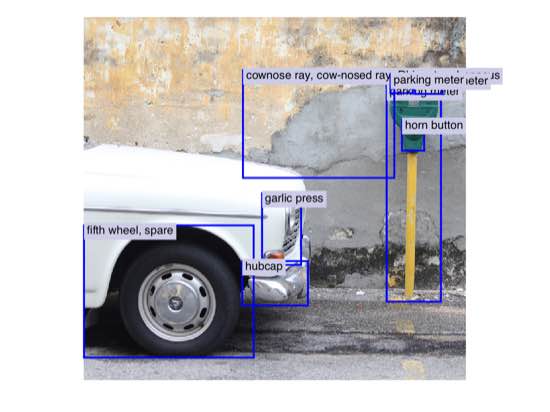} & 
\includegraphics[trim = 14mm 5mm 14mm 20mm , clip=true,width=0.25\textwidth,height=\textheight,keepaspectratio]{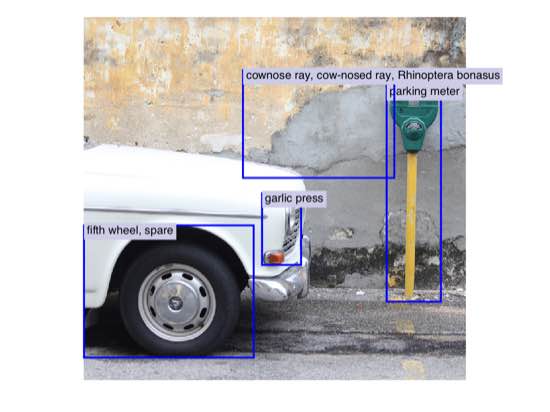} & 
\includegraphics[trim = 14mm 5mm 14mm 20mm , clip=true,width=0.25\textwidth,height=\textheight,keepaspectratio]{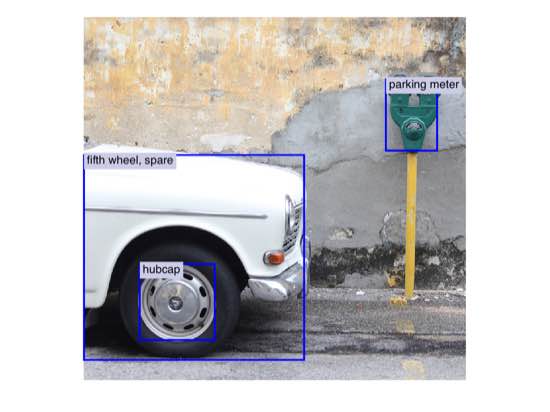} \\

\includegraphics[trim = 14mm 20mm 14mm 20mm , clip=true,width=0.25\textwidth,height=\textheight,keepaspectratio]{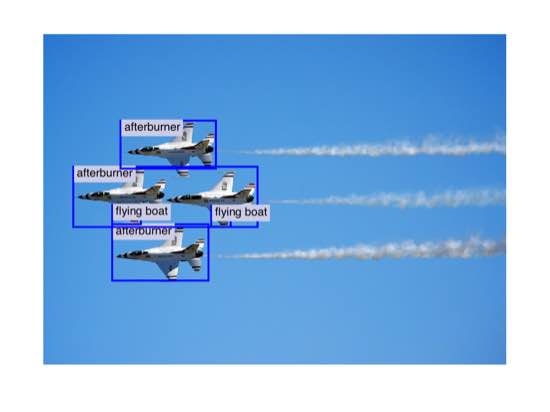} & 
\includegraphics[trim = 14mm 20mm 14mm 20mm , clip=true,width=0.25\textwidth,height=\textheight,keepaspectratio]{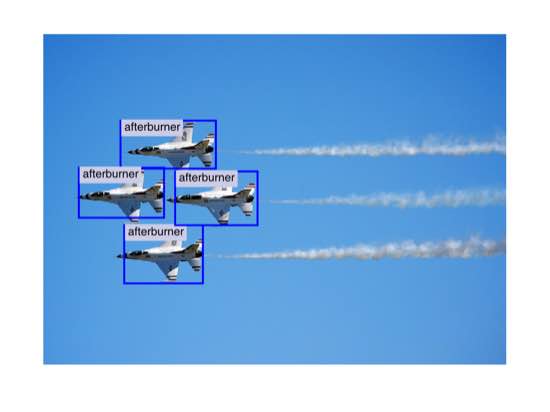} & 
\includegraphics[trim = 14mm 20mm 14mm 20mm , clip=true,width=0.25\textwidth,height=\textheight,keepaspectratio]{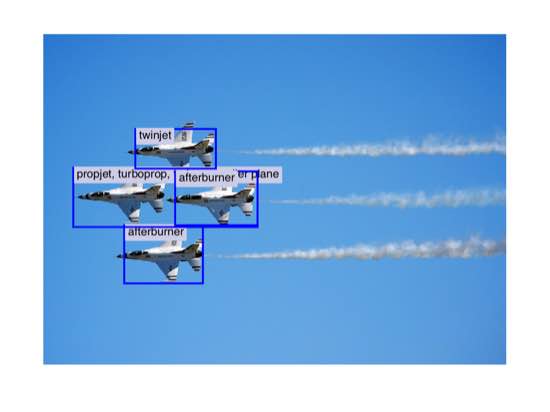} \\

\includegraphics[trim = 14mm 20mm 14mm 5mm , clip=true,width=0.25\textwidth,height=\textheight,keepaspectratio]{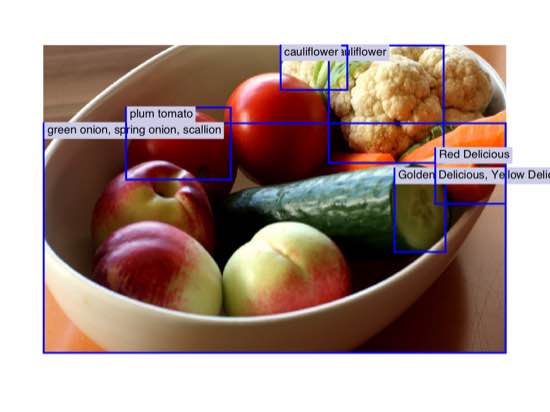} & 
\includegraphics[trim = 14mm 20mm 14mm 5mm , clip=true,width=0.25\textwidth,height=\textheight,keepaspectratio]{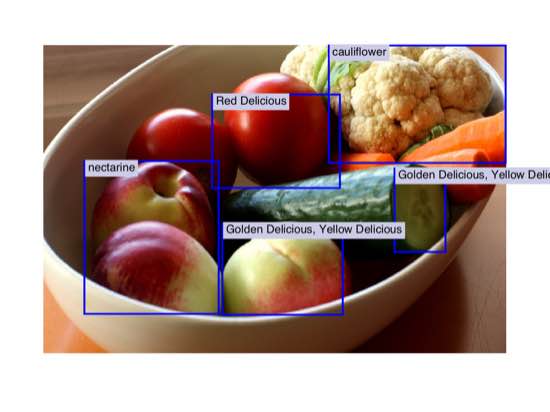} & 
\includegraphics[trim = 14mm 20mm 14mm 5mm , clip=true,width=0.25\textwidth,height=\textheight,keepaspectratio]{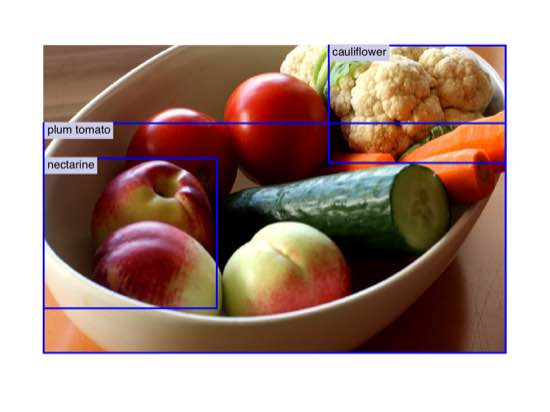} \\

\includegraphics[trim = 14mm 20mm 14mm 20mm , clip=true,width=0.25\textwidth,height=\textheight,keepaspectratio]{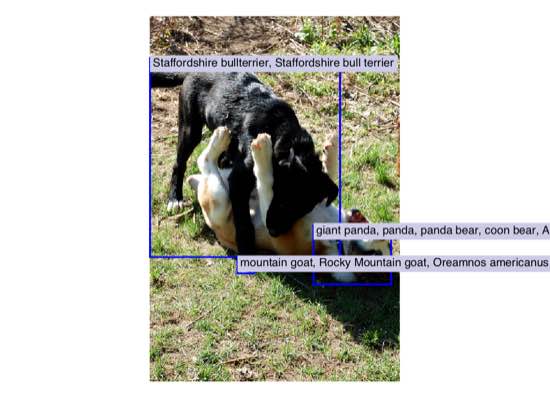} & 
\includegraphics[trim = 14mm 20mm 14mm 20mm , clip=true,width=0.25\textwidth,height=\textheight,keepaspectratio]{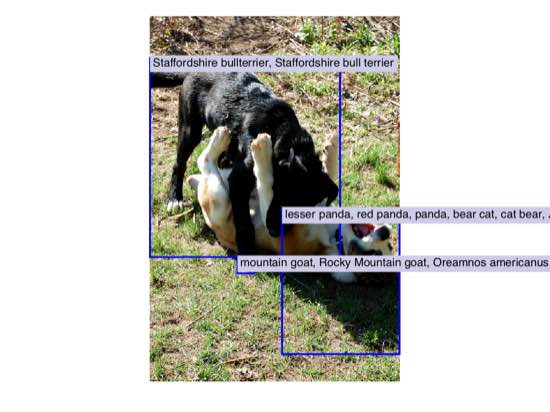} & 
\includegraphics[trim = 14mm 20mm 14mm 20mm , clip=true,width=0.25\textwidth,height=\textheight,keepaspectratio]{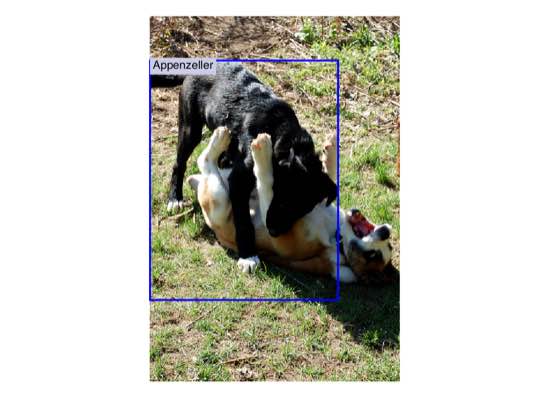} \\

\includegraphics[trim = 14mm 20mm 14mm 20mm , clip=true,width=0.25\textwidth,height=\textheight,keepaspectratio]{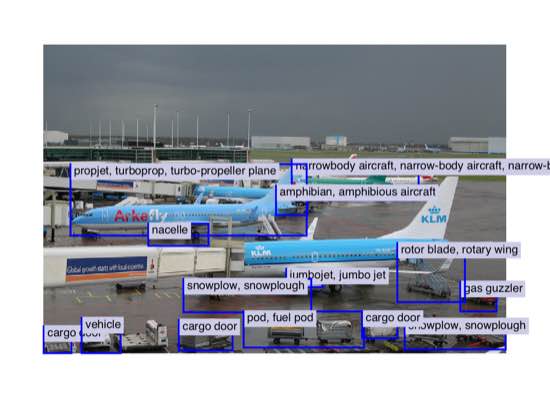} & 
\includegraphics[trim = 14mm 20mm 14mm 20mm , clip=true,width=0.25\textwidth,height=\textheight,keepaspectratio]{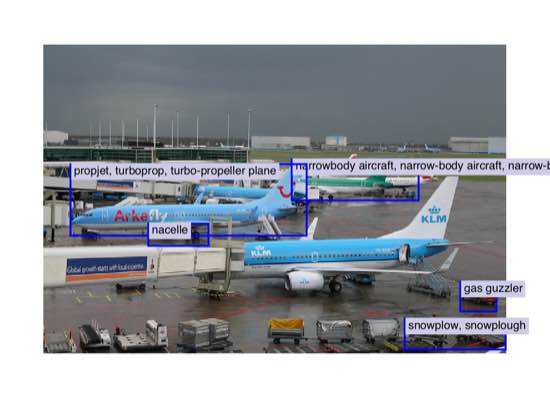} & 
\includegraphics[trim = 14mm 20mm 14mm 20mm , clip=true,width=0.25\textwidth,height=\textheight,keepaspectratio]{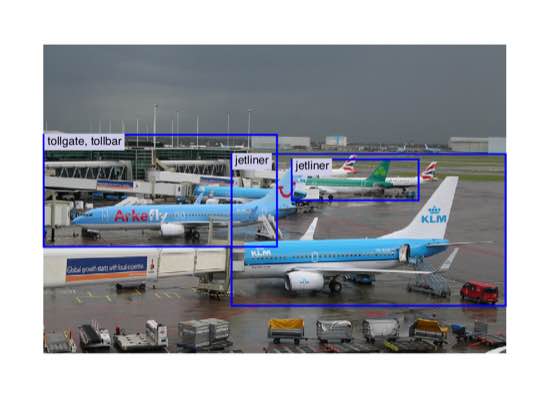} \\

\end{tabular}
\end{center}
\caption{Examples where MAPC better classifies objects than WC+AC-NMS and SAPC + AC-NMS on 7,404 ImageNet classes. No ground truth is available. Detections:
blue.}
\label{table:LargeBetterClass}
\end{figure*}

In this section, we show and generally discuss images, where MAPC significantly better classified objects than the baselines. These images can be seen in Figures \ref{table:COCOBetterClass}, \ref{table:ImageNetBetterClass} and \ref{table:LargeBetterClass}. When we look at these images, we can see that MAPC clearly benefits from taking semantic similarities into account during the labeling of detection proposals. As in multi-class detection each object detection is not only classified by one class, but by a confidence score distribution across all classes, the problem of labelling each detection with the correct class arises. Hence, the MAPC similarity was formulated such that not only the spatial similarity, but also the semantic class similarity between all detection proposals in one cluster is maximized. In contrast, SAPC and NMS do not take class labels into account at all. NMS simply suppresses overlapping detections by the top scoring detection, whereas SAPC only relies on spatial relations for its inference. Further, when SAPC and NMS is applied to object detection, each detection proposal is simply labeled by the top scoring class. The input for MAPC however can also contain multiple classes per detection proposal. MAPC will then select the class for each detection, which maximizes the semantic and spatial similarity between all detection proposals in one cluster at the same time. All in all, this results in significantly better
classified detections.

\subsection{Erroneous Examples}
\label{failuremapc}
\begin{figure*}
\begin{center}
\begin{tabular}{c@{}c@{}c@{}c}
Ground Truth & WC+AC-NMS & SAPC + AC-NMS & MAPC (ours) \\
\hline\hline

\includegraphics[trim = 14mm 5mm 14mm 5mm , clip=true,width=0.25\textwidth,height=\textheight,keepaspectratio]{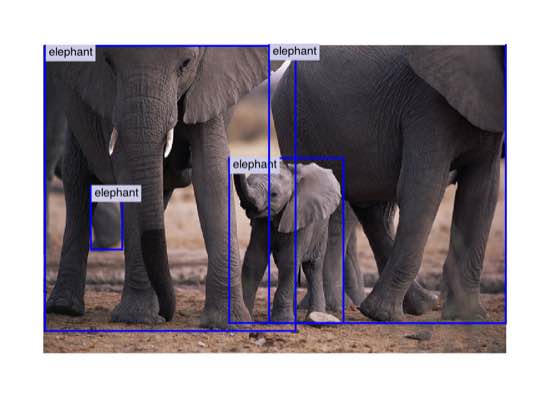} & 
\includegraphics[trim = 14mm 5mm 14mm 5mm , clip=true,width=0.25\textwidth,height=\textheight,keepaspectratio]{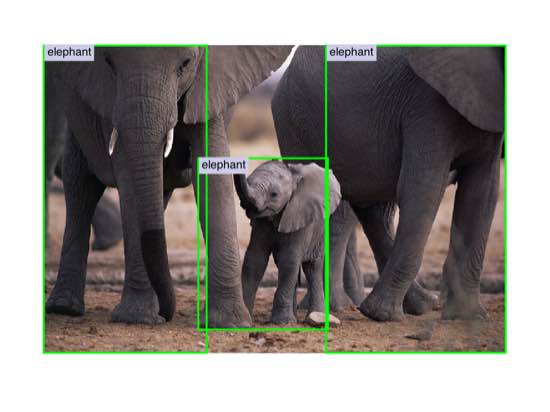} &
\includegraphics[trim = 14mm 5mm 14mm 5mm , clip=true,width=0.25\textwidth,height=\textheight,keepaspectratio]{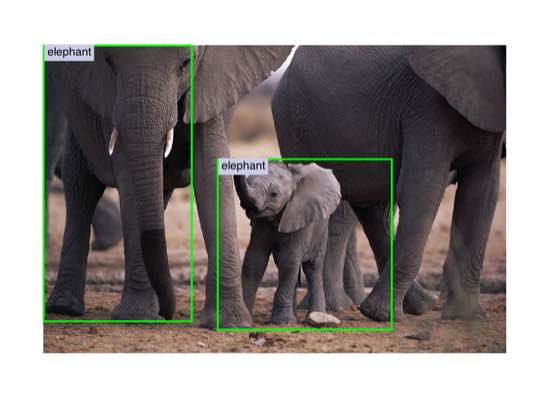} & 
\includegraphics[trim = 14mm 5mm 14mm 5mm , clip=true,width=0.25\textwidth,height=\textheight,keepaspectratio]{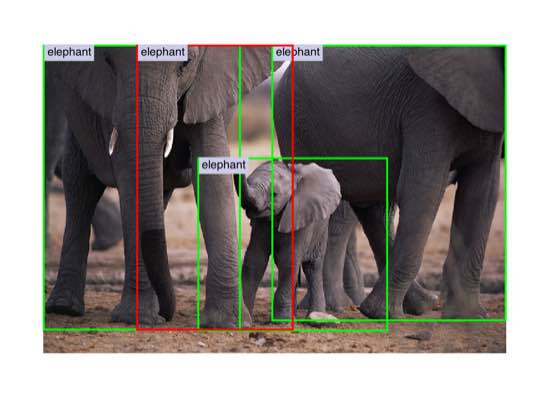} \\

\includegraphics[trim = 14mm 5mm 14mm 20mm , clip=true,width=0.25\textwidth,height=\textheight,keepaspectratio]{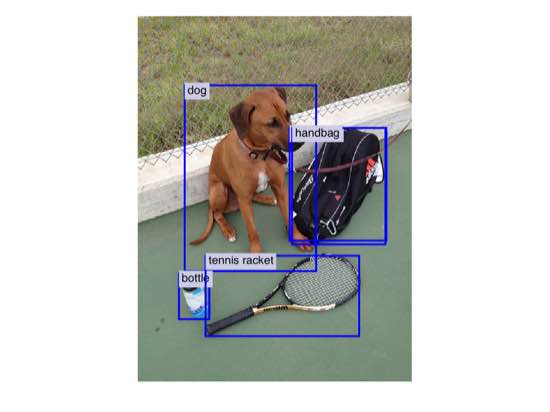} & 
\includegraphics[trim = 14mm 5mm 14mm 20mm , clip=true,width=0.25\textwidth,height=\textheight,keepaspectratio]{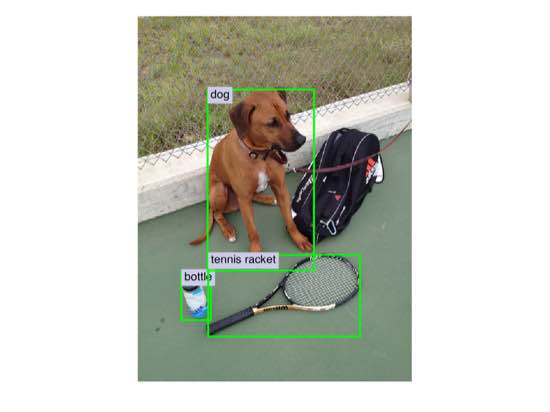} &
\includegraphics[trim = 14mm 5mm 14mm 20mm , clip=true,width=0.25\textwidth,height=\textheight,keepaspectratio]{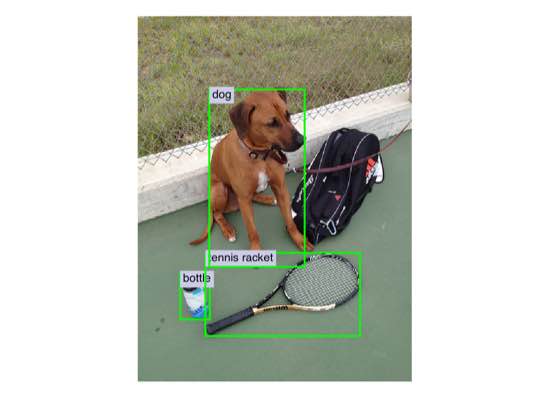} & 
\includegraphics[trim = 14mm 5mm 14mm 20mm , clip=true,width=0.25\textwidth,height=\textheight,keepaspectratio]{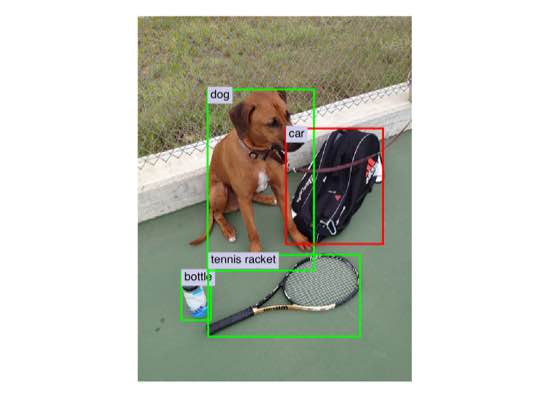} \\

\includegraphics[trim = 14mm 5mm 14mm 5mm , clip=true,width=0.25\textwidth,height=\textheight,keepaspectratio]{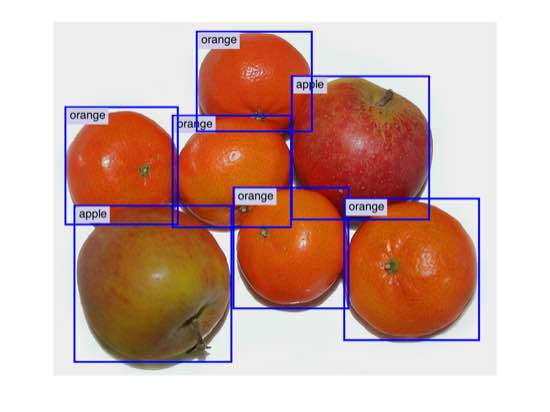} & 
\includegraphics[trim = 14mm 5mm 14mm 5mm , clip=true,width=0.25\textwidth,height=\textheight,keepaspectratio]{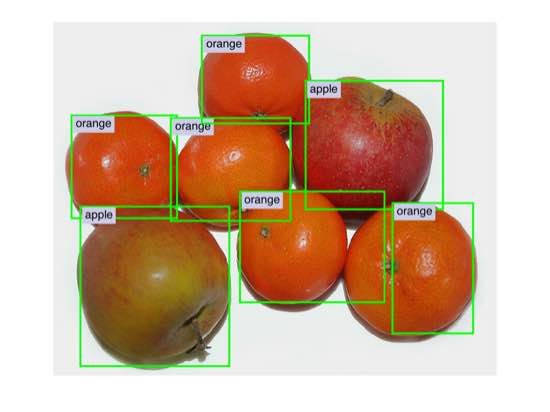} &
\includegraphics[trim = 14mm 5mm 14mm 5mm , clip=true,width=0.25\textwidth,height=\textheight,keepaspectratio]{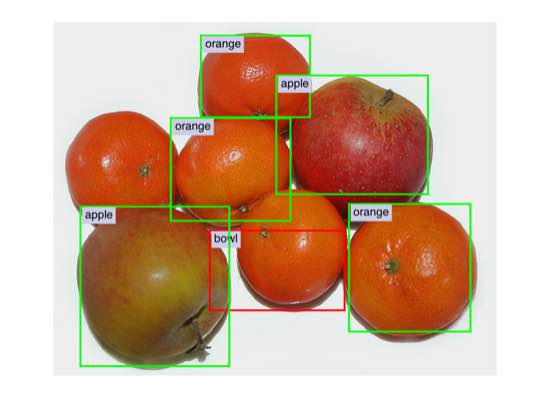} & 
\includegraphics[trim = 14mm 5mm 14mm 5mm , clip=true,width=0.25\textwidth,height=\textheight,keepaspectratio]{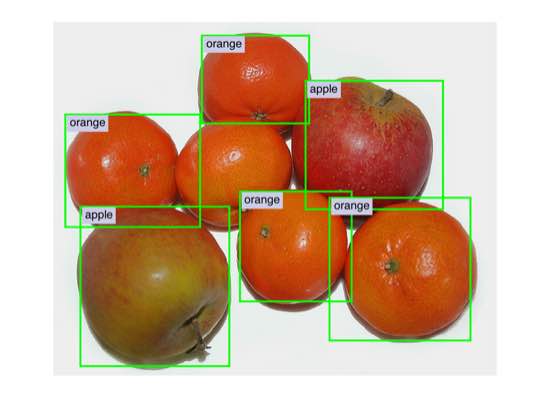} \\

\includegraphics[trim = 14mm 5mm 14mm 20mm , clip=true,width=0.25\textwidth,height=\textheight,keepaspectratio]{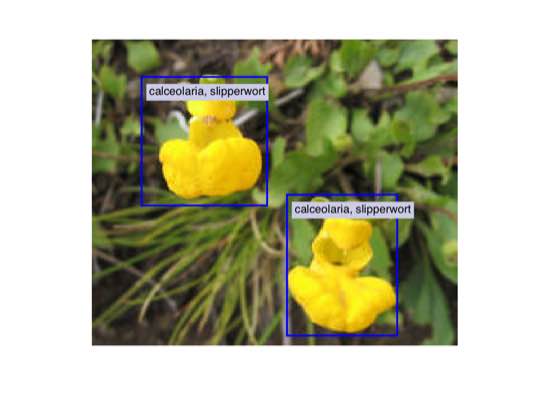} & 
\includegraphics[trim = 14mm 5mm 14mm 20mm , clip=true,width=0.25\textwidth,height=\textheight,keepaspectratio]{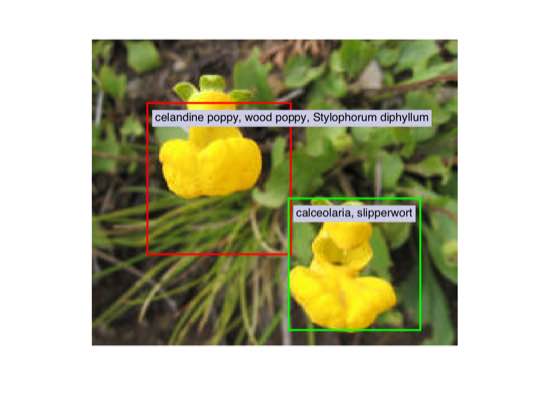} &
\includegraphics[trim = 14mm 5mm 14mm 20mm , clip=true,width=0.25\textwidth,height=\textheight,keepaspectratio]{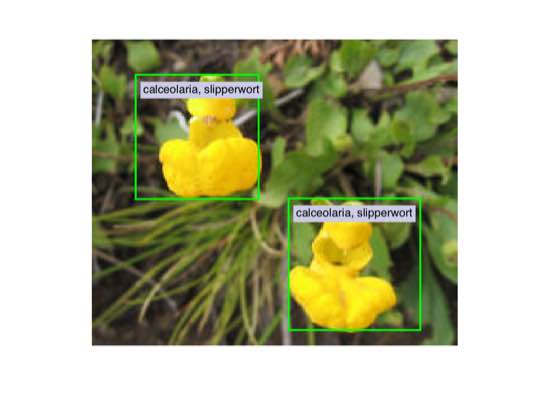} & 
\includegraphics[trim = 14mm 5mm 14mm 20mm , clip=true,width=0.25\textwidth,height=\textheight,keepaspectratio]{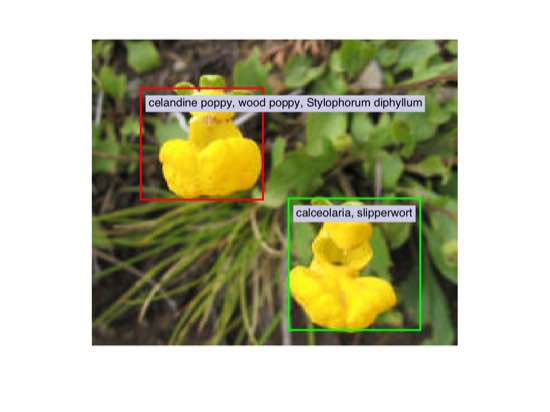} \\

\includegraphics[trim = 14mm 5mm 14mm 20mm , clip=true,width=0.25\textwidth,height=\textheight,keepaspectratio]{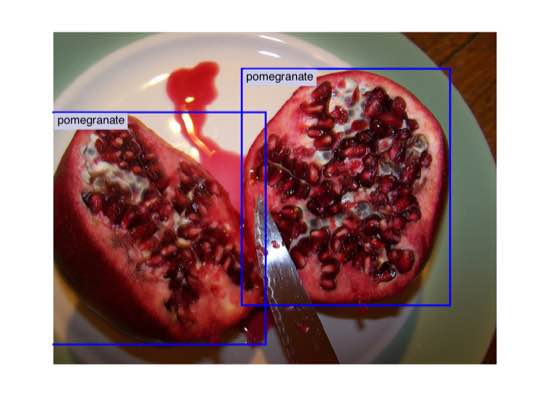} & 
\includegraphics[trim = 14mm 5mm 14mm 20mm , clip=true,width=0.25\textwidth,height=\textheight,keepaspectratio]{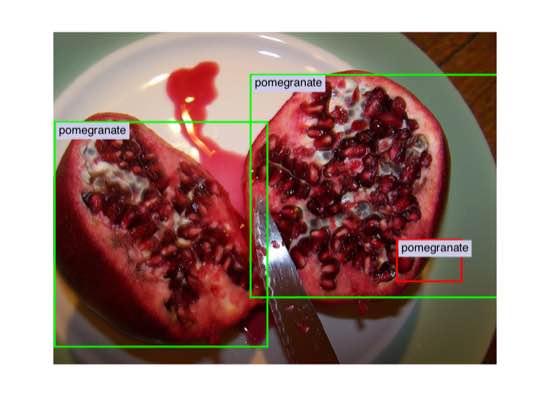} &
\includegraphics[trim = 14mm 5mm 14mm 20mm , clip=true,width=0.25\textwidth,height=\textheight,keepaspectratio]{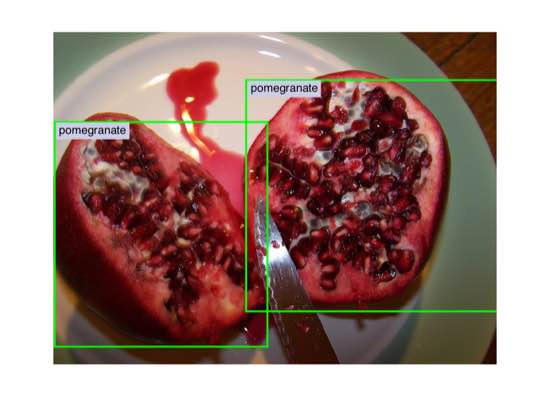} & 
\includegraphics[trim = 14mm 5mm 14mm 20mm , clip=true,width=0.25\textwidth,height=\textheight,keepaspectratio]{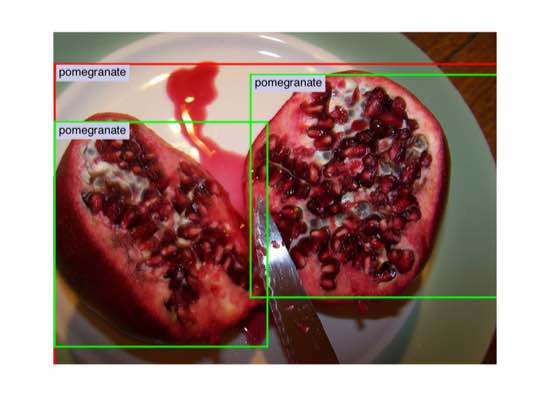} \\

\includegraphics[trim = 14mm 25mm 14mm 5mm , clip=true,width=0.25\textwidth,height=\textheight,keepaspectratio]{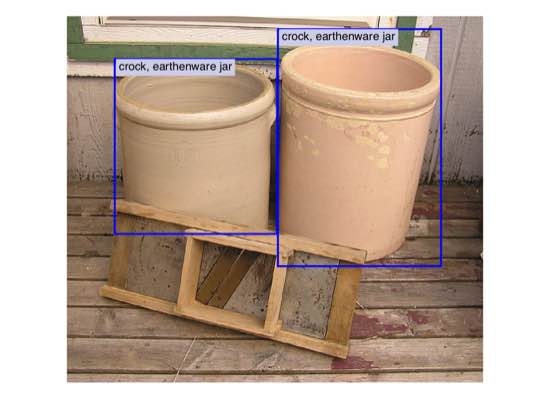} & 
\includegraphics[trim = 14mm 25mm 14mm 5mm , clip=true,width=0.25\textwidth,height=\textheight,keepaspectratio]{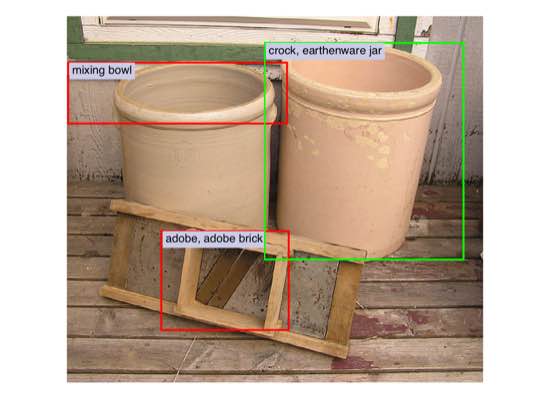} &
\includegraphics[trim = 14mm 25mm 14mm 5mm , clip=true,width=0.25\textwidth,height=\textheight,keepaspectratio]{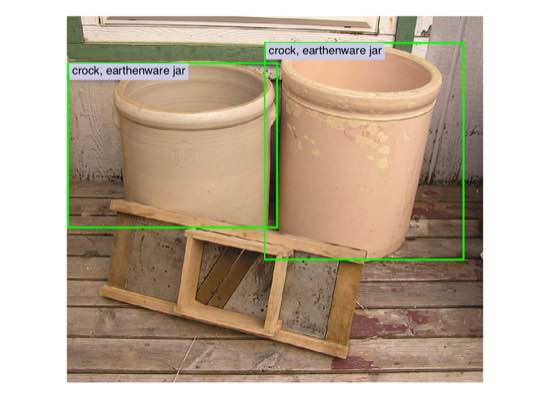} & 
\includegraphics[trim = 14mm 25mm 14mm 5mm , clip=true,width=0.25\textwidth,height=\textheight,keepaspectratio]{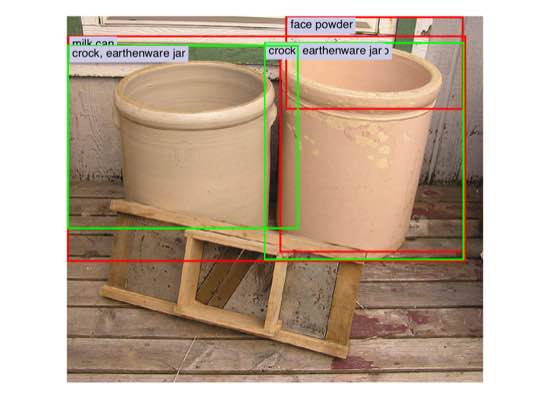} \\

\end{tabular}
\end{center}
\caption{Examples where MAPC performs worse than WC+AC-NMS or SAPC + AC-NMS on
Microsoft COCO.
Ground truth: blue, true positives: green, false positives: red.}
\label{table:Bad}
\end{figure*}

In this section we look at images, where MAPC performs worse than SAPC + AC-NMS or WC+AC-NMS. As can be seen in Figure \ref{table:Bad}, MAPC sometimes results in multiple detections for the same object. Hence, MAPC not always eliminates all false positives. These false positives can be both from the same category or a different category than the underlying true positive detection. This means that the semantic or spatial clustering respectively did not work correctly. In these cases, the parameters can be adjusted until all false positives are eliminated. Nonetheless, there are cases, where MAPC detects the wrong category or localises an object in the wrong area, whereas SAPC + AC-NMS or WC+AC-NMS
find the correct object and label it correctly.

\section{MAPC explained step-by-step}
\begin{figure*}
\begin{center}
\includegraphics[width=\textwidth,height=\textheight,keepaspectratio]{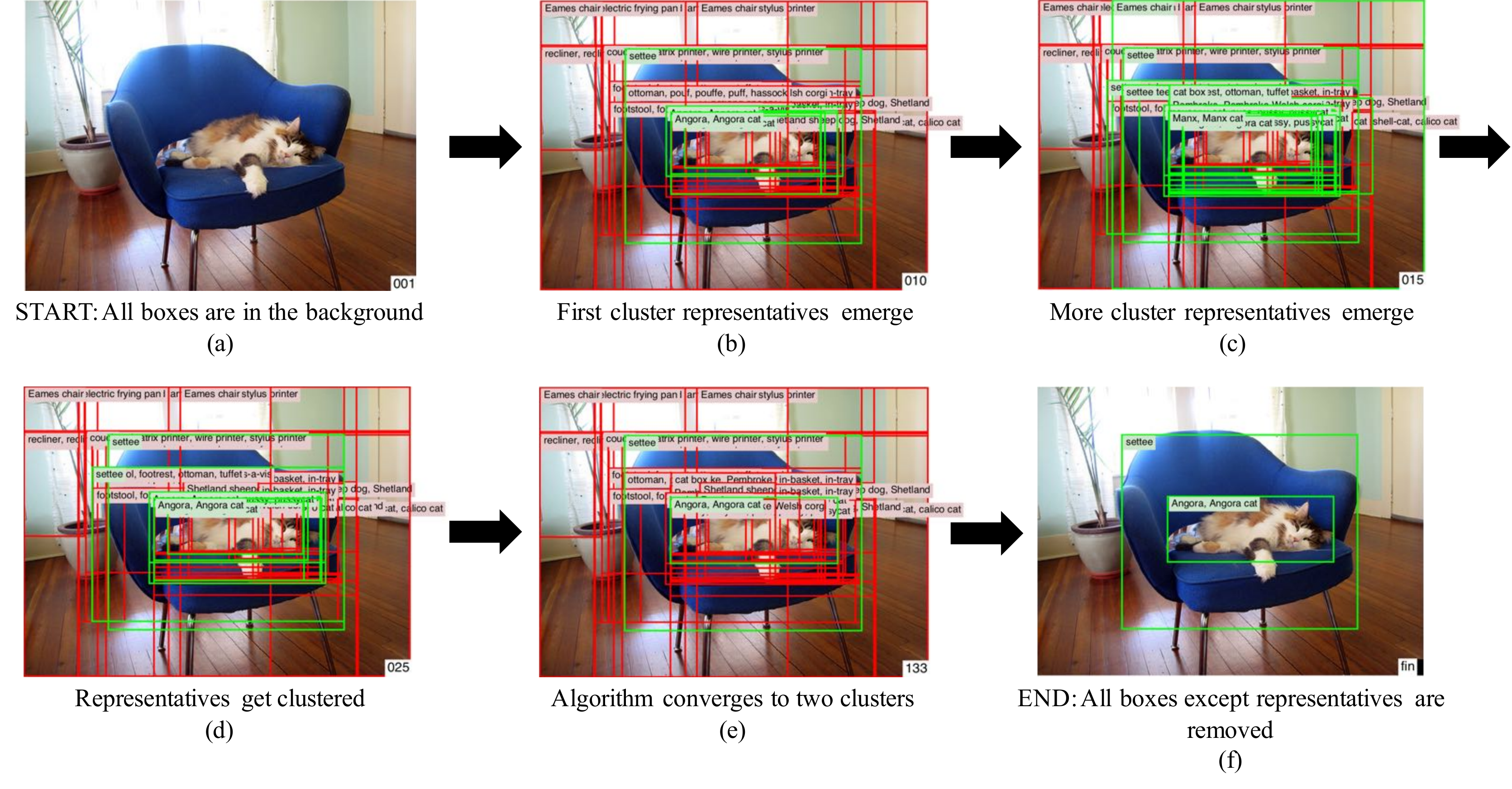}
\caption{MAPC clustering process step-by-step. Cluster representatives: green bounding boxes. Cluster members: red bounding boxes. All bounding boxes which belong to the background cluster are not depicted. The number of iterations is illustrated in the lower right corner of each picture. %
}
\label{figure:ClusteringProcess}
\end{center}
\end{figure*}

In order to better understand how MAPC works, this section will explain
step-by-step how MAPC clusters are formed throughout the iterative message passing updates on one example. This example is also used to verify that indeed spatially and semantically similar object detection proposals are clustered together, and that the detections are selected, which best represent these clusters. The most relevant iterations of MAPC are depicted in Figure \ref{figure:ClusteringProcess}. All detection proposals which are cluster representatives are coloured green. All detection proposals which are not cluster representatives but belong to one of the green representatives are depicted in red. All detection proposals which belong to the background cluster are not depicted.

MAPC is an iterative algorithm, which follows the message passing paradigm of \cite{eth_biwi_01126} until convergence. As can be seen in the first image of Figure \ref{figure:ClusteringProcess}, all detection proposals are in the background cluster at the initial iteration (Figure
\ref{figure:ClusteringProcess} (a)). In the following, messages are passed between all detection proposals. Based on the spatial semantic similarity, detection proposals with similar classes and locations get clustered together and first clusters with representatives emerge (Figure \ref{figure:ClusteringProcess} (b)). More clusters are formed in this
process (Figure \ref{figure:ClusteringProcess} (c)). These clusters are grouped together until the similarity between each representative and each detection proposal in the clusters is maximized (Figure \ref{figure:ClusteringProcess} (d)). Once the setup of clusters with their respective representatives does not change anymore, the algorithm converges (Figure \ref{figure:ClusteringProcess} (e)). All detection proposals, except the cluster representatives, get removed, which results in the depicted object detections (Figure \ref{figure:ClusteringProcess} (f)). It can be observed that throughout the whole process cluster representatives are mainly different types of \emph{chairs} or \emph{cats}. Thus, this example shows that object detections are chosen as cluster representatives that represent the spatial and semantic distribution of all detection proposals. Summarizing, MAPC clusters semantically and spatially similar detection proposals, representing them by the object detections which maximize the similarity between all detection proposals and their representatives.

\end{document}